\documentclass[sn-mathphys,Numbered]{sn-jnl}
\usepackage{amssymb}
\usepackage{multirow}
\usepackage{booktabs} %
\usepackage{float}
\usepackage{subfigure}
\usepackage{graphics,graphicx}
\graphicspath{{figs/},{../figs/},{../},{}}
\usepackage{hyperref} %
\hypersetup{
  colorlinks = true,
  urlcolor = blue, 
}
\usepackage[ruled,algo2e]{algorithm2e}%
\usepackage{xcolor}

\SetCommentSty{mycommfont}
\usepackage{amsmath,xspace}
\DeclareMathAlphabet{\mathbsf}{OT1}{cmss}{bx}{n}%
\DeclareMathAlphabet{\mathlf}{OT1}{cmss}{m}{l}%
\DeclareMathAlphabet{\mathslf}{OT1}{cmss}{m}{sl}%

\newcommand{\dosage}{\ensuremath{\mathlf{dosage}}\xspace}
\newcommand{\variation}{\ensuremath{\mathlf{variation}}\xspace}
\newcommand{\engagement}{\ensuremath{\mathlf{engagement}}\xspace}
\newcommand{\location}{\ensuremath{\mathlf{location}}\xspace}
\newcommand{\temperature}{\ensuremath{\mathlf{temperature}}\xspace}

\newcommand{\var}[1][z]{\ensuremath{\mathsf{#1}}\xspace}
\newcommand{\avail}{I}
\newcommand{\resimtag}{{\mrm{sim}}}

\newcommand{\resimtraj}{\texttt{ParaSim}\xspace}

\newcommand{\setcomment}[1]{{\footnotesize{/\!/\! #1}}}

\newcommand{\lval}[1][]{\texttt{Frac}_{#1}}
\newcommand{\intscore}[1][]{\ensuremath{\texttt{Score\_int}_{#1}}\xspace}
\newcommand{\isint}[1][\hypoint]{\ensuremath{\texttt{User\_int}_{#1}}\xspace}
\newcommand{\numint}[1][\hypoint]{\ensuremath{\texttt{\#User\_int}_{#1}}\xspace}

\newcommand{\numbootstrap}{\mfk{B}}

\newcommand{\mfk}{\mathfrak}

\newcommand{\indicator}{\boldsymbol{1}}

\newcommand{\x}{x}

\newcommand{\axi}[1][i]{\x_{#1}}

\renewcommand{\l}{\ell}

\newcommand{\vareps}{\varepsilon}

\newcommand{\wtil}[1]{\widetilde{#1}}

\newcommand{\pseqxn}[1][n]{(\axi[i])_{i\geq 1}} %
\newcommand{\pseqxnn}[1][n]{(\axi[i])_{i=1}^n} %

\SetKwInput{KwInit}{Init}

\newcommand{\brackets}[1]{\left[ #1 \right]}

\newcommand{\sparenth}[1]{( #1)}
\newcommand{\sbraces}[1]{ \{ #1\}}
\newcommand{\parenth}[1]{\left( #1 \right)}

\newcommand{\tp}{^\top}
\newcommand{\inv}{^{-1}}
\newcommand{\real}{\ensuremath{\mathbb{R}}}

\def\balign#1\ealign{\begin{align}#1\end{align}}
\def\baligns#1\ealigns{\begin{align*}#1\end{align*}}
\def\balignat#1\ealign{\begin{alignat}#1\end{alignat}}
\def\balignats#1\ealigns{\begin{alignat*}#1\end{alignat*}}
\def\bitemize#1\eitemize{\begin{itemize}#1\end{itemize}}
\def\benumerate#1\eenumerate{\begin{enumerate}#1\end{enumerate}}

\newenvironment{talign*}
 {\let\displaystyle\textstyle\csname align*\endcsname}
 {\endalign}
\newenvironment{talign}
 {\let\displaystyle\textstyle\csname align\endcsname}
 {\endalign}

\def\balignst#1\ealignst{\begin{talign*}#1\end{talign*}}
\def\balignt#1\ealignt{\begin{talign}#1\end{talign}}
\newcommand{\qtext}[1]{\quad\text{#1}\quad} 
\newcommand{\stext}[1]{\ \text{#1}\ }

\let\originalleft\left
\let\originalright\right
\renewcommand{\left}{\mathopen{}\mathclose\bgroup\originalleft}
\renewcommand{\right}{\aftergroup\egroup\originalright}

\def\tinycitep*#1{{\tiny\citep*{#1}}}
\def\tinycitealt*#1{{\tiny\citealt*{#1}}}
\def\tinycite*#1{{\tiny\cite*{#1}}}
\def\smallcitep*#1{{\scriptsize\citep*{#1}}}
\def\smallcitealt*#1{{\scriptsize\citealt*{#1}}}
\def\smallcite*#1{{\scriptsize\cite*{#1}}}

\def\mbb#1{\mathbb{#1}}
\def\mc#1{\mathcal{#1}}
\def\mrm#1{\mathrm{#1}}
\def\trm#1{\textrm{#1}}
\def\tbf#1{\textbf{#1}}

\def\til#1{\widetilde{#1}}

\def\textsum{{\textstyle\sum}} %
\def\Q{\mathbb{Q}}
\def\N{\mathbb{N}}
\def\<{\left\langle} %
\def\>{\right\rangle}

\def\defeq{\triangleq} %

\newcommand{\floor}[1]{\lfloor{#1}\rfloor}
\newcommand{\ceil}[1]{\lceil{#1}\rceil}
\def\E{\mbb{E}} %

\def\P{\mbb{P}} %

\newcommand{\iid}{\textrm{i.i.d.}\ }

\providecommand{\diag}{\mathop\mathrm{diag}}

\ifdefined\nonewproofenvironments\else
\ifdefined\ispres\else
\newenvironment{proof-sketch}{\noindent\textbf{Proof Sketch}
  \hspace*{1em}}{\qed\bigskip\\}
\newenvironment{proof-idea}{\noindent\textbf{Proof Idea}
  \hspace*{1em}}{\qed\bigskip\\}
\newenvironment{proof-of-lemma}[1][{}]{\noindent\textbf{Proof of Lemma {#1}}
  \hspace*{1em}}{\qed\\}
\newenvironment{proof-of-theorem}[1][{}]{\noindent\textbf{Proof of Theorem {#1}}
  \hspace*{1em}}{\qed\\}
\newenvironment{proof-attempt}{\noindent\textbf{Proof Attempt}
  \hspace*{1em}}{\qed\bigskip\\}

\usepackage{amsthm,thmtools,thm-restate}
\usepackage{cleveref}  
\usepackage{autonum}
\newtheorem{remark}{Remark}

\newtheorem{definition}{Definition}

\newtheorem{assumption}{Assumption}

\newtheorem{corollary}{Corollary}
\newtheorem{lemma}{Lemma}
\newtheorem{proposition}{Proposition}

\newtheorem{myproposition}{Proposition}

\newtheorem{mycorollary}{Corollary}

\newtheorem{mylemma}{Lemma}

\newtheorem{mydefinition}{Definition}

\newtheorem{myremark}{Remark}

\newtheorem{myassumption}{Assumption}

 \crefname{appendix}{App.}{App.}
\crefname{equation}{}{}
\crefname{lemma}{Lem.}{Lem.}
\crefname{theorem}{Thm.}{Thm.}
\crefname{Corollary}{Cor.}{Cors.}
\crefname{algorithm}{Alg.}{Algs.}

\crefname{section}{Sec.}{Sec.}
\crefname{table}{Tab.}{Tab.}
\crefname{remark}{Rem.}{Rem.}
\crefname{definition}{Def.}{Def.}
\crefname{Proposition}{Prop.}{Prop.}
\crefname{myremark}{Rem.}{Rem.}
\crefname{mylemma}{Lem.}{Lem.}
\crefname{mydefinition}{Def.}{Defs.}
\crefname{myproposition}{Prop.}{Prop.}
\crefname{mycorollary}{Cor.}{Cors.}
\crefname{myassumption}{Assum.}{Assum.}
\crefname{figure}{Fig.}{Fig.}
\crefname{enumi}{}{}
\crefname{name}{}{} %

\usepackage{sansmath} %
\usepackage{enumitem}

\usepackage{amsfonts}%
\usepackage{amsthm}%
\usepackage{mathrsfs}%
\usepackage[title]{appendix}%
\usepackage{textcomp}%
\usepackage{manyfoot}%
\usepackage{listings}%

\usepackage[normalem]{ulem}
\usepackage{pifont}%
\newcommand{\xmark}{\ding{55}}%
\usepackage{makecell}

\begin{document}

\title{Did we personalize? Assessing personalization by an online reinforcement learning algorithm using resampling}

\author[1]{\fnm{Susobhan} \sur{Ghosh}}\email{susobhan\_ghosh@g.harvard.edu}
\equalcont{These authors contributed equally to this work.}

\author*[2]{\fnm{Raphael} \sur{Kim}}\email{raphaelkim@fas.harvard.edu}
\equalcont{These authors contributed equally to this work.}

\author[6]{\fnm{Prasidh} \sur{Chhabria}}\email{chhabria@g.harvard.edu}
\author*[1,3,4]{\fnm{Raaz} \sur{Dwivedi}}\email{raaz@g.harvard.edu}
\author[5]{\fnm{Predrag} \sur{Klasnja}}\email{klasnja@umich.edu}
\author[6]{\fnm{Peng} \sur{Liao}}\email{pliao9122@gmail.com}
\author[1]{\fnm{Kelly} \sur{Zhang}}\email{kellywzhang@seas.harvard.edu}
\author*[1,3]{\fnm{Susan} \sur{Murphy}}\email{samurphy@fas.harvard.edu}

\affil*[1]{\orgdiv{Department of Computer Science}, \orgname{Harvard University}}

\affil[2]{\orgdiv{Department of Biostatistics}, \orgname{Harvard University}}

\affil[3]{\orgdiv{Department of Statistics}, \orgname{Harvard University}}

\affil[4]{\orgdiv{Department of Electrical Engineering and Computer Science}, \orgname{MIT}}

\affil[5]{\orgdiv{School of Information}, \orgname{University of Michigan}}

\affil[6]{\small{Work done by these authors while they were at Harvard University}}

\abstract{ %

 There is a growing interest in using reinforcement learning (RL) to personalize sequences of treatments in digital health to support users in adopting healthier behaviors. Such sequential decision-making problems involve decisions about when to treat and how to treat based on the user's context (e.g., prior activity level, location, etc.). Online RL is a promising data-driven approach for this problem as it learns based on each user's historical responses and uses that knowledge to personalize these decisions. However, to decide whether the RL algorithm should be included in an ``optimized'' intervention for real-world deployment, we must assess the data evidence indicating that the RL algorithm is actually personalizing the treatments to its users. Due to the stochasticity in the RL algorithm, one may get a false impression that it is learning in certain states and using this learning to provide specific treatments. We use a working definition of personalization and introduce a resampling-based methodology for investigating whether the personalization exhibited by the RL algorithm is an artifact of the RL algorithm stochasticity. We illustrate our methodology with a case study by analyzing the data from a physical activity clinical trial called HeartSteps, which included the use of an online RL algorithm. We demonstrate how our approach enhances data-driven truth-in-advertising of algorithm personalization both across all users as well as within specific users in the study.

}

\maketitle

\section{Introduction}
\label{sec:intro}
The use of reinforcement learning (RL) algorithms is becoming more prevalent in various applications such as mobile health (mHealth), online education, and online advertisements. The main objective of RL algorithms is to optimize the delivery of effective treatments to users based on their current state, with the goal of maximizing a weighted sum of rewards that represents their response to specific treatments. These rewards are designed in a way that higher values indicate more desirable responses. For instance, in an mHealth app, an RL algorithm may send mobile notifications to promote physical activity based on the user's state. In this case, the reward for the algorithm would be operationalized as the amount of physical activity the user performs over some period of time.

Numerous mHealth applications utilize RL algorithms to build interventions. Works using bandit algorithms to send mobile notifications/messages include increasing physical activity~\cite{yom2017encouraging,liao2020personalized}, promoting weight loss~\citep{forman2019can,forman2023using}, helping users quit smoking~\cite{albers2022addressing}, improving adherence for diabetes treatments \cite{Yom-Tov2017EncouragingSystem} and improving chronic pain care~\cite{piette2022artificial}. Partially observable Markov decision processes (POMDPs) have also been used in mHealth applications like helping the elderly using assistive technology \cite{DBLP:conf/aaaifs/HoeyPBM05}, and aiding people with dementia \cite{DBLP:conf/ijcai/BogerPHBFM05}. In another work~\cite{DBLP:conf/iui/ZhouMFGFKCA18}, inverse RL has been used to personalize mobile fitness applications for behavioral weight loss.
The use of RL algorithms in the above applications is based on the assumption that some actions may work better than other actions when a particular user is in a particular state. RL algorithms aim to learn the person-specific mappings between actions, states, and rewards in order to maximize the intervention's overall effectiveness (and, therefore, the obtained reward). Therefore, the designer aims to use an online RL algorithm that learns from user data and selects the most advantageous actions.  If the RL algorithm successfully learns the advantages of different actions for a user in a particular state, we say that the RL algorithm has \emph{personalized} for that user in that state\footnote{This is our informal working definition of personalization. In \cref{sub:personalization}, we formally define personalization and ways to measure personalization.}. The designer's objective is to design an RL algorithm that can personalize, and our goal is to conduct an exploratory data analysis to assess whether it has accomplished this objective.

\label{par:PersonalizationVsRegret}
This motivates analyzing the advantage forecasts across states produced by the algorithm and observing how these advantages change over time. It is worth noting that an online RL algorithm uses its forecasts to select actions, receive rewards, and then update the forecasts based on the resulting rewards. If a sequence of forecasts for similar states consistently trends in the same direction, it can indicate that the advantages are consistently leading to improved rewards. These trends tend to indicate that the algorithm is learning to personalize and using the learning to assign advantageous actions repeatedly. When we observe such trends, e.g., when visualizing them via a graph like in \cref{fig:interestingUser}, we call the user a potentially \emph{interesting user} and call the graph \emph{visually interesting}. However, observing visually interesting and consistent trends in advantage forecasts does not necessarily mean that the RL algorithm has personalized for that user. 

In this paper, our goal is to develop exploratory methods to promote \emph{truth-in-advertising} about the personalization achieved by the RL algorithm in real-life applications. That is, we want to ask if it is true that the RL algorithm is indeed personalizing or if the interesting trends that suggest that the RL algorithm is personalizing are arising just by chance due to the stochasticity of the RL algorithm. And we want to answer this question by analyzing the data at hand. 
Besides truth in advertising about the RL algorithm’s exhibited learning, our methodology also aims to inform the study designers about how to tweak and improve the RL algorithm for future usage (see \cref{sub:motivation}).

Personalization in this work is closely related to cumulative reward maximization, as the algorithm would have achieved personalization if it managed to choose more advantageous actions in different states visited by the user. While a theoretical guarantee on the standard regret bound for the RL algorithm is useful, it is not verifiable in practice since the optimal policy (with respect to which the regret is measured) is not readily available. In real-life applications, like the HeartSteps trial considered in this work, RL algorithms with low regret guarantees often serve as the starting point. However, one acknowledges that the chosen RL algorithm may not achieve low regret for the real-life problem at hand for a variety of reasons, ranging from the invalidity of assumptions, misspecified models, low signal-to-noise ratio, the number of decision times, or even design tweaks to the RL algorithm, made due to (e.g., computational or stability) constraints of the problem at hand.

\subsection{A motivating phenomenon from HeartSteps}
\label{sub:motivation}
Our work is inspired by our participation in the HeartSteps mobile health trial. Throughout this paper, we use this trial to clarify our ideas. The  HeartSteps RL algorithm is a generalization of the popular linear Thompson-Sampling algorithm~\cite{DBLP:journals/ftml/RussoRKOW18}. In a linear Thompson-Sampling algorithm, rewards are represented using a Gaussian linear model with a Gaussian prior on the model parameters, leading to a Gaussian posterior for the model parameters. The HeartSteps RL algorithm selects one of two treatment actions, i.e., sending a physical activity suggestion tailored to the user's current setting or not, at five user-specific times per day for each user. Moreover, the algorithm selects the action in a stochastic manner using posterior sampling. In particular, in a given state, the algorithm sends the suggestion with the posterior probability that the reward from sending a suggestion is higher than the reward if no suggestion is sent in that state.

When designing the HeartSteps RL algorithm, the team hypothesized that three state features---\variation, \engagement, and \location  (see \cref{sec:heartsteps_intro,tab:state_features} for their definitions)---would determine the advantage of sending an activity suggestion. After the trial, it is of interest to the team to assess whether these state features did help in forecasting the advantage of sending an activity suggestion. More generally, an assessment about which features are useful for personalization is important from a domain science point of view for multiple reasons: (1) In planning the next iteration of the RL algorithm, scientists can reason about which state features to include; (2) scientists can use these analyses to reason about health science theories; (3) if the RL algorithm personalizes based on some features, this would indicate that it is worthwhile to continue development of an RL algorithm that takes into account state as opposed to, for example, a simple multi-armed bandit (which is simpler to implement in an online setting).

\begin{figure}[ht!]
\centering
\resizebox{\textwidth}{!}{
\begin{tabular}{cc}
     \includegraphics[width=0.49\textwidth]{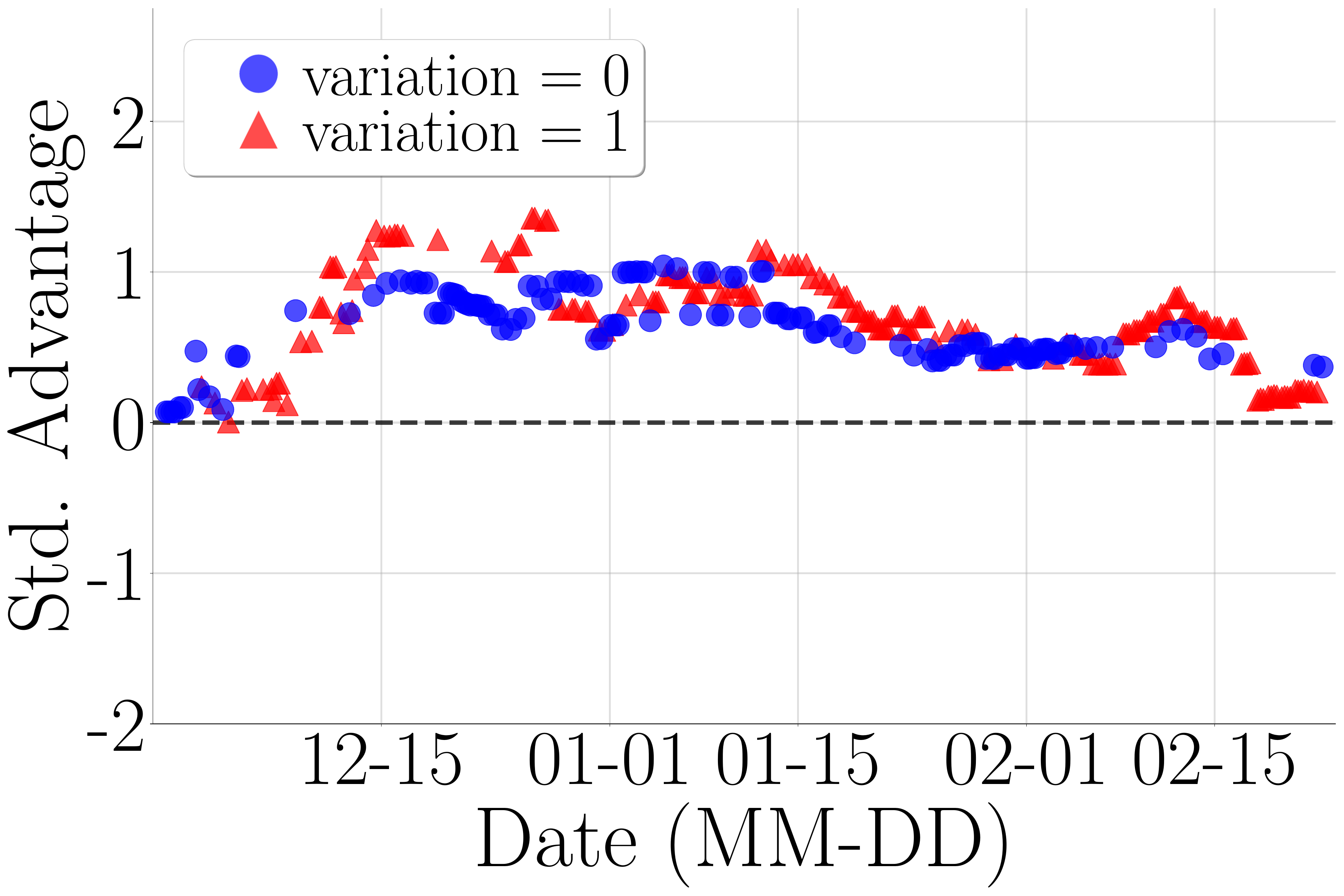} 

     & \includegraphics[width=0.49\textwidth]{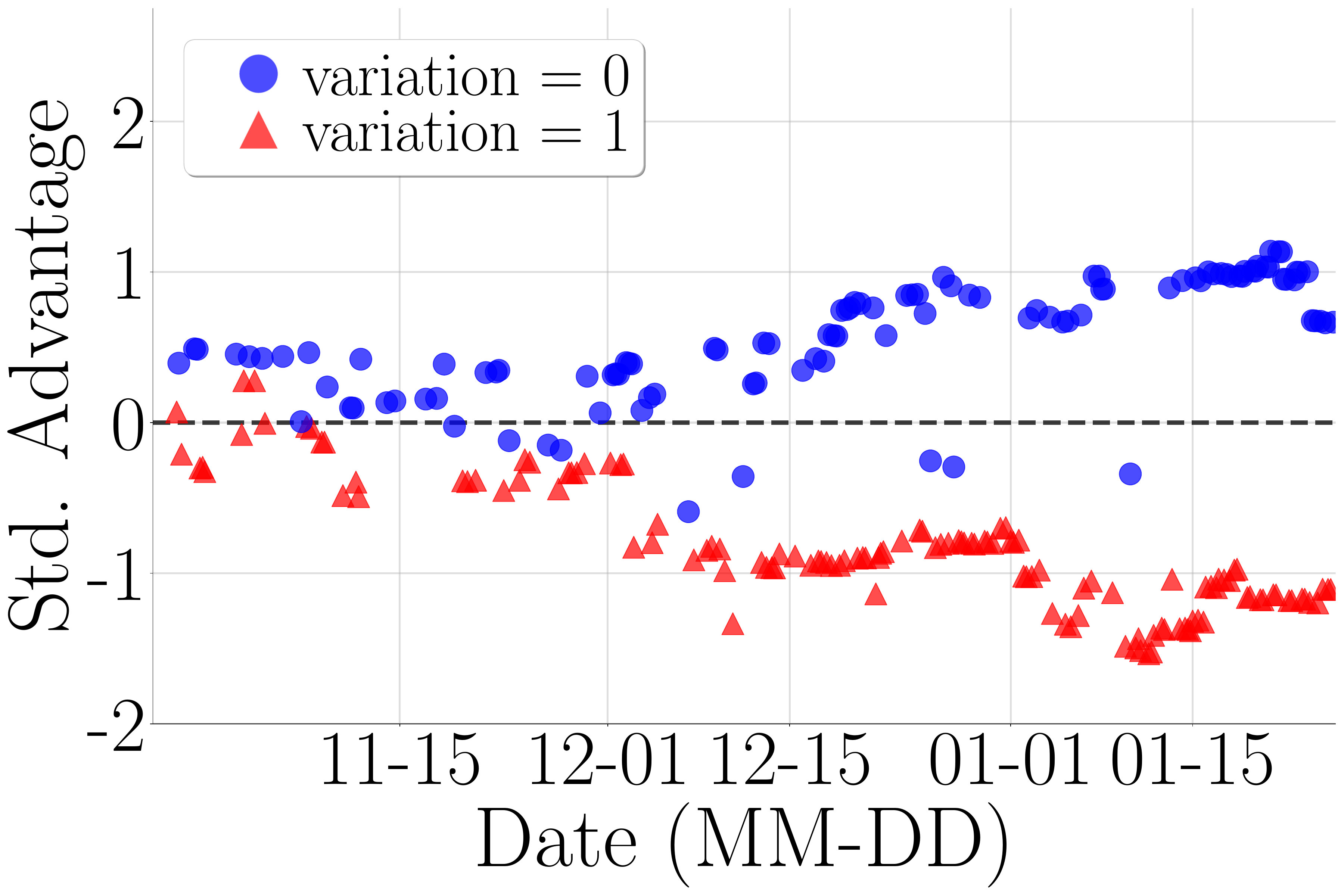}  \\

     (a) User 1 & (b) User 2
\end{tabular}
}
\caption{\tbf{Two instances of ``interesting'' advantage forecasts: Are we seeing evidence of personalization?} The two panels plot results for two distinct users from the HeartSteps trial. The value on the vertical axis represents the RL algorithm's forecast of the standardized advantage of sending an activity message for the user in the user's current state (note each day has $5$ decision times). The forecasts are marked as blue circles if \variation= 0 and red triangles if \variation= 1 at the decision time. %
\label{fig:interestingUser}}
\end{figure}

We elaborate our motivating questions of interest for the HeartSteps RL algorithm by first focusing on assessing whether feature \variation is useful for personalization. This feature takes a value of either 0 or 1 in a given timeslot (remember that there are five times per day) if the user's past 7-day step in that time slot is less or more variable than the average historical data\footnote{Details on this feature and others are provided further in \cref{tab:state_features}}. In \cref{fig:interestingUser}, we illustrate the algorithm's advantage forecasts for two of the 91 users in the study. The horizontal axis shows the user's study day, while the vertical axis shows the ``standardized posterior advantage'' of sending an activity suggestion at each time. The points on the panels are color-coded based on the value of \variation. The standardized posterior advantage at time $t$ is the algorithm's prediction of the advantage of sending an activity suggestion at that time based on pre-time $t$ data on the user. Roughly speaking, the advantage is computed by subtracting the posterior means of the rewards for when an activity suggestion is sent versus not sent and then dividing it by the posterior standard deviation of the advantage in the user's current state. Furthermore, starting from the 8th day in each of the panels, the standardized advantage also roughly translates into treatment selection probabilities at each time in a monotonic manner (see \cref{rem:std_advg,eq:probability}). 

\paragraph{Interesting type $1$}
In panel (a), we observe an intriguing pattern for user $1$. The advantage forecasts stay consistently above zero after the initial few days, regardless of any feature's value, including \variation. It is tempting to conclude that the RL algorithm personalized quickly for this user, learning that sending activity suggestions is always beneficial in the states experienced by this user. %

\paragraph{Interesting type $2$}
For user $2$ in panel (b), we observe a striking difference in the advantage forecast when \variation is $0$ compared to when it is $1$. The standardized advantage when the \variation feature is $1$ (user's past 7-day step in that timeslot is \emph{more} variable than on average in the historical data) is mostly negative and lower than when the feature is $0$ (user's past 7-day step in that timeslot is \emph{less} variable than on average in the historical data), where it is mostly positive and higher. The consistency across time exhibited in panel (b) is even more striking because higher standardized advantage forecasts result in the algorithm delivering activity suggestions with a higher probability when the user has \variation = 0, and the reward subsequent to the delivery of the treatment is used to update the standardized advantage that would be used at the next time. We notice that this user's graph appears interesting only when contrasted based on the feature \variation (see \cref{sec:user_2_details} for this user's graph colored by \location and \engagement). We might be tempted to conclude that the RL algorithm personalized for user $2$ by learning that sending an activity suggestion is better when \variation is $0$. Additionally, if we find that multiple users exhibit similarly striking graphs involving the \variation feature, we might be tempted to conclude that the \variation feature is useful for personalization more broadly.

\subsection{Key questions}
It is important to exercise caution when interpreting the effectiveness of the RL algorithm and whether a feature is useful for personalization. Apparent personalization (as in~\cref{fig:interestingUser}) may arise purely by chance due to algorithmic stochasticity. 
In this paper, we consider a first start at  truth-in-advertising by assessing whether stochasticity in the RL algorithm can lead to misleading results.   We focus on truth-in-advertising limited to this consideration and conduct/present exploratory analyses that address two questions:
\begin{enumerate}[label=(Q\alph*.), leftmargin=*]
\item\label{ques:numint} If we find multiple users with similar interesting graphs as shown in \cref{fig:interestingUser}(a) and/or \cref{fig:interestingUser}(b), does the data support the conclusion that this occurrence for multiple users is greater than chance due to the stochasticity of the RL algorithm?
\item\label{ques:userint} Does the data suggest that a specific user's graph, as shown in \cref{fig:interestingUser}(a) and/or \cref{fig:interestingUser}(b), is truly interesting and that this type of graph is unlikely to appear by chance due to the RL algorithm's stochasticity?
\end{enumerate}

 A positive answer to the first question would suggest that many users in the study could benefit from the RL algorithm and that the RL algorithm is effective at personalization when possible.
A positive answer to the second question would indicate that the RL algorithm is effectively personalizing for a particular user. 
Note that a variety of additional exploratory analyses might be considered, both concerning other measures of \lq\lq interestingness" as well as standard analyses for examining model misspecification. However here we limit the discussion to the above two questions.

\subsection{Contributions and organization}
We develop and present exploratory data analyses based on resampling~\cite{good2006resampling}\footnote{We remark that throughout this paper, we use resampling and resimulations interchangeably. 
In particular, our methodology resimulates user trajectories that we generate by resampling states and rewards using generative models, and resampling actions by re-running the RL algorithm.} for investigating whether a stochastic RL algorithm personalized treatment selection. 
At a high-level, our methodology proceeds in three steps: First, we operationalize whether RL exhibits personalization on a given trajectory via quantitative metrics, called interestingness scores and the number of interesting users; see \cref{sub:personalization}. Second, for a given notion of interestingness, we simulate user trajectories several times, with carefully selected generative models for states and rewards, and re-running the RL algorithm to resample actions. These generative models are coupled with the notion of interestingness under investigation; the RL algorithm should not exhibit personalization, i.e., the generated trajectories should not exhibit interesting patterns. We compute the corresponding interestingness scores for each of these resimulated trajectories. Finally, if the interestingness scores on the original trajectories differ significantly from the distribution of these scores on the resimulated trajectories, then we say that the personalization exhibited by the RL algorithm is not likely to arise solely due to the RL algorithm's stochasticity in sampling actions. Otherwise, we conclude that the personalization in the original data might have arisen just due to the RL algorithm's stochasticity in sampling actions. We summarize our methodology in \cref{subsec:resampling}.

The remainder of the paper is organized as follows.
In \cref{sec:problem_setup}, we review RL, provide a working definition of personalization by an RL algorithm, and develop a resampling methodology to investigate whether a stochastic RL algorithm was able to personalize. We provide further details concerning the HeartSteps RL algorithm and trial and apply the methods to the resulting data in \cref{sec:heartsteps_intro}. Finally, we discuss related work in~\cref{sec:related_work} and conclude with limitations, and future directions in \cref{sec:discussion}. Additional technical details about HeartSteps and our methodology are provided in the appendix.

\newcommand{\simplex}{\Delta}
\section{A resampling-based framework for truth-in-advertising}
\label{sec:problem_setup}

We start with a brief review of RL in \cref{subsec:background}, followed by an overview of our strategy in \cref{subsec:truth_advertise}. \cref{sub:personalization} presents our working definition of personalization by an online stochastic RL algorithm and operationalizes the key questions \cref{ques:numint,ques:userint}.  \cref{subsec:resampling} presents our resampling-based framework to investigate whether an RL algorithm was able to personalize. Note that we apply the resampling methodology to address \cref{ques:numint,ques:userint} for HeartSteps in \cref{sec:heartsteps_intro}.
 
\subsection{Notation and background on RL}
\label{subsec:background}
We consider a standard setup from the RL literature~\cite{Sutton1998ReinforcementIntroduction}, that of a Markov decision process (MDP)~\cite{bellman1957markovian} wherein an RL algorithm, e.g., the mobile app, interacts with the environment, e.g., the user. In this work, MDP is given by the tuple $(\mc S, \mc A, T, \mbb P, r)$, where $\mc S$ denotes the state space $\mc S$, $\mc A$ the treatment action space, $T$ the total decision times, and $\mbb P$
the transition dynamics with $\mbb P (\cdot \vert s, a) \in \simplex_{\mc S}$ denoting the (time-homogeneous) state transition probabilities at decision time when taking action $a \in \mc A$ in state $s \in \mc S$.
A user trajectory is given by $\{ S_{t}, A_{t}, R_{t} \}_{t= 1}^{T}$, where $S_t$ denotes the state at decision time $t$, $A_t$ the action assigned by the RL algorithm at time $t$, and $R_t$ the reward collected after selection of the action.  We define $r(s, a) \defeq \E[R_t|S_t=s, A_t=a]$  as the (time-homogeneous) mean reward function under state $s$ and action $a$.

\paragraph{RL algorithm}
The goal of the RL algorithm is to optimize treatment action delivery to maximize the sum of rewards $\E[\sum_{t=1}^{T} R_t]$. Towards this goal, an online RL algorithm learns a sequence of \emph{policies} $\sbraces{\pi_t}_{t=1}^{T}$, where a policy $\pi$ denotes a collection of probability distributions $\pi(\cdot|s) %
$ one per state $s \in \mc S$.\footnote{Note that the underlying problem might restrict the allowed actions depending on the value of $s$. For example, in HeartSteps, sending a notification is not allowed when the user is driving.
} An online RL algorithm learns the policy $\pi_t$ from the user data $(S_{t'}, A_{t'}, R_{t'})_{t'=1}^{t-1}$ till time $t-1$. Then given user state $S_t$ at time $t$, it samples treatment $A_t \sim \pi_t(\cdot \vert S_t)$, and collects reward $R_t$. The process repeats itself all over again at time $t+1$.

\paragraph{Advantage function}
Let $\E_{\pi}[\cdot]$ denote the expectation with respect to the trajectory induced by selecting actions using policy $\pi$ at each decision time. Then for each $t\in [T]$, the state-action-value function $Q_t^\pi$ for policy $\pi$ is defined as $Q_t^{\pi}(s, a) \defeq \E_{\pi}[\textsum_{t'=t}^{T} R_{t'} | S_t=s, A_t = a]$ for $s \in \mc S, a \in \mc A$. A natural quantity for evaluating the performance of the RL algorithm is the advantage function of action $a$ over $a'$ in state $s$ at time $t$, which we define as $\Delta_{t}^{\pi}(s, a, a') \defeq Q_t^{\pi}(s, a) - Q_t^{\pi}(s, a')$. Here we focus on binary actions and define the advantage as $Q_t^{\pi}(s,1)-Q_t^{\pi}(s,0)$. An online RL algorithm would often estimate this advantage and construct a policy $\pi_t$ to sample the more advantageous actions with higher probability. For a bandit setting, $S_{t+1} \perp (S_{t}, A_{t})$, and with a binary action space $\mc A = \sbraces{0 ,1}$, the advantage function of a policy $\pi$ at time $t$ in state $s$ and actions $1$ and $0$ is independent of $\pi$. In particular, the advantage for this setting is policy and time-invariant and equal to the difference in mean reward function, i.e., $\Delta_{t}^{\pi}(s, 1, 0) = r(s, 1)-r(s, 0)$, where $r(s,a)= \E[R_t | S_t=s, A_t=a]$.

\paragraph{Example of RL algorithms}
We focus on online stochastic RL algorithms that select actions at time $t$ based on a combination of prior $t$ data and external randomness. Such a design results in a stochastic selection, with probabilities in $(0,1)$, of treatment actions. Many commonly used RL algorithms assign treatment stochastically, such as Thompson sampling~\cite{thompson1933likelihood, russo2018tutorial}, Boltzmann exploration (softmax)~\cite{Sutton1998ReinforcementIntroduction}, $\epsilon$-greedy exploration~\cite{Sutton1998ReinforcementIntroduction}, information-directed sampling \cite{DBLP:conf/nips/RussoR14}, and hyper model based exploration~\cite{DBLP:conf/iclr/DwaracherlaLIOW20}. 
HeartSteps uses posterior sampling to stochastically select treatment actions (\cref{subsec:rlalg}).   In the development of digital interventions there is usually a sequence of trials (i.e., in HeartSteps there were 3 with more ongoing) with each trial  being used to inform the development of the RL algorithm for subsequent trial. Stochasticity in the RL algorithm facilitates between-study causal inferences, especially when the analyses are not pre-specified.  Further one can generally expect some  distribution drift, that is potential non-stationarity between trials.  The stochasticity in the RL algorithm facilitates the use of the resulting data  to allow for this potential non-stationarity~\cite{yang2020targeting}. 
Note that, in contrast, a range of RL algorithms that use Upper Confidence Bounds (UCB) and its variants~\cite{DBLP:journals/ml/AuerCF02, DBLP:journals/pmh/AuerO10}, select actions deterministically based on prior data. We do not consider such deterministic algorithms here. Unless otherwise specified, an RL algorithm in this work refers to an online stochastic RL algorithm.

\subsection{Truth-in-advertising using resampling}
\label{subsec:truth_advertise}

As mentioned earlier, to determine whether an online stochastic RL algorithm is personalizing, one way is to examine graphs showing the estimated advantage or its standardized version on the vertical axis and time on the horizontal axis, as seen in \cref{fig:interestingUser}. However, simply observing visually interesting patterns in these graphs is not enough to conclude that the RL algorithm personalized for that user. To address this issue, we propose a resampling-based approach to assess personalization and improve truth-in-advertising.

The approach consists of three steps: First, we quantify the interestingness of a graph of advantage forecasts. {Second, we simulate user trajectories by resampling treatments using the RL algorithm under different generative models. We call the trajectories ``resampled trajectories'' throughout this paper.  Finally, we compare the interestingness score in the user graph from the actual study with the interestingness scores from the resampled trajectories under different models. If the score in the actual study differs significantly from the scores on the resampled trajectories, we say that the RL algorithm is potentially personalizing; otherwise, the exhibited personalization might arise solely due to the stochasticity in the RL algorithm. We now describe each step in detail.

\subsection{Personalization and interestingness}
\label{sub:personalization}
We start by providing a definition of personalization for an online stochastic RL algorithm used throughout this work. 

\paragraph{Working definition of personalization}
 Personalization by an online stochastic RL algorithm occurs when the algorithm (1) learns which action(s) in a given state lead to higher subsequent total rewards, and (2) continues to select the better action(s) in that state with higher probability. Note that such an assessment of personalization is indeed concerned with cumulative reward maximization. In particular, our methodology to follow assesses if the RL algorithm seems to be learning to differentially select actions to maximize rewards, whether such learning is spurious and can be attributed to the stochastic variation for the actions selected by the RL algorithm.

\paragraph{Simplified notation}  To simplify the presentation, we assume binary action space. We use $\hat{\Delta}_{t}(s)$ to denote the RL algorithm's forecast of the advantage (or its standardized version) at time $t$ under policy, i.e., $\hat{\Delta}_{t}(s)$ denotes a forecast of $\Delta_{t}^{\hat{\pi}_t}(s, 1, 0)$ where  $\hat\pi_t$ denote the policy learned by the RL algorithm using data till time $t-1$. For example, in the case of HeartSteps, this forecast would be given by the difference in posterior mean of rewards in state $s$ for treatment $1$ vs $0$ (see \cref{eq:hts} for the standardized version).
We use $\mc U = \sbraces{(S_t, A_t, R_t, \hat{\Delta}_{t})}_{t=1}^{T}$ to denote a generic user's trajectory including the advantage forecast functions. Given a set of $n$ users, we use $\mc U_i$ to denote the trajectory of user $i$.

With the notation in place, we are ready to operationalize our working definition for personalization, based on observed data. Here we focus on RL algorithms that make use of advantage forecasts. Various choices exist to measure why a user trajectory of advantages might indicate personalization and thus be interesting.  Here we present two ways to capture personalization or equivalently (throughout this work) interestingness, representing the two panels in \cref{fig:interestingUser}. 

\paragraph{Operationalizing interestingness of type 1}
To operationalize whether the advantage forecasts in a user trajectory $\mc U$ are consistently positive irrespective of the observed current state (like in \cref{fig:interestingUser}(a)), we use the score
\begin{align}
\label{eq:intscore1}
    \intscore[1](\mc U) \defeq \frac{\sum_{t= 1}^T \indicator( \hat{\Delta}_{t}(S_t)>0)}{T}.
\end{align}
That is, $\intscore[1](\mc U)$ measures the fraction of times the advantage forecasts are positive in the trajectory $\mc U$. For a user trajectory with no advantage on average, this score would concentrate around $0.5$. Thus, a trajectory should be deemed interesting (indicative of personalization) if this score deviates away from $0.5$. We say that a user trajectory $\mc U$ is interesting of type 1 if $|\intscore[1](\mc U)-0.5| \geq \delta$ for a suitable choice of $\delta\in(0, 0.5)$. A larger threshold $\delta$ would require the user \intscore[1] to be more extreme to qualify as an interesting user. For a set of $n$ user trajectories $\sbraces{\mc U_i}_{i=1}^{n}$, we count the number of interesting users of type $1$ as
\begin{align}
    \label{eq:numint1}
    \numint[1] \defeq \sum_{i=1}^n \indicator(|\intscore[1](\mc U_i)-0.5| \geq \delta).
\end{align}
In other words, $\numint[1]$ denotes the number of users for whom their advantage forecasts signal personalization by the RL algorithm in the sense that one of the two actions are consistently more advantageous.

\paragraph{Operationalizing interestingness of type 2}
To operationalize whether (a binary feature) $\var$ consistently leads to differential advantage forecasts, (like in \cref{fig:interestingUser}(b)), we use the score:
\begin{align}
\label{eq:intscore2}
    \intscore[2, \var](\mc U) \defeq \frac{\sum_{t= 1}^T \indicator( \hat{\Delta}_{t}(S_t(\var=1))>\hat{\Delta}_{t}(S_t(\var=0))) }{T},
\end{align}
where $S_t(\var=1)$ denotes a state that takes same value as $S_t$ for all features except $\var$, which is set to $1$. When the advantage function is independent of the feature $\var$, the score $ \intscore[2, \var](\mc U)$ would concentrate around $0.5$. Hence, we say that a user trajectory $\mc U$ is interesting of type $2$ for feature $\var$ if $|\intscore[2, \var](\mc U)-0.5| \geq \delta$ for a suitable choice of $\delta\in(0, 0.5)$. That is, for such a user, we say that the advantage forecasts suggest that the RL algorithm is potentially personalizing based on feature \var. Moreover, for a set of $n$ user trajectories $\sbraces{\mc U_i}_{i=1}^{n}$, we count the number of interesting users of type $2$ for feature $\var$ as
\begin{align}
    \label{eq:numint2}
    \numint[2,\var] \defeq \sum_{i=1}^n \indicator(|\intscore[2,\var](\mc U_i)-0.5| \geq \delta).
\end{align}
In words, $\numint[2,\var]$ denotes the number of users for whom their advantage forecasts signal that the RL algorithm is personalizing, by learning to treat differentially based on the value of feature $\var$. (Note that the choice of $\delta$ in \cref{eq:numint1,eq:numint2} can be different.)\\

\begin{remark}
    We use $\intscore[]$ and $\numint[]$ throughout the paper to respectively denote a generic interestingness score and the associated number for interesting users.
\end{remark}

\paragraph{Operationalizing questions~\cref{ques:numint,ques:userint}} 
Consider an interestingness type quantified by \intscore[] with an associated count of interesting users given by \numint[].  Then we operationalize \cref{ques:numint} as asking how likely a count as high as $\numint[]$ can appear due to the RL algorithm's stochasticity. Furthermore, we operationalize \cref{ques:userint} for a specific user with trajectory $\mc U$ as asking how likely a value as high as $|\intscore[](\mc U)-0.5|$ can appear due to algorithmic stochasticity.

\subsection{Resampling framework}
\label{subsec:resampling}
Given a question of interest, we use a resampling-based strategy to investigate if an RL algorithm is personalizing. 

We state a general resampling approach in \cref{resim} that can be applied to more general settings than considered in this work (see \cref{future}). Our approach mimics the original data-generating process using a generative model and resamples a user trajectory given (i) a (parametric) model $\mc M$ for the user's data-generating process and (ii) the RL algorithm under consideration. Here the generative model $\mc M$ includes a distribution $\rho_1$ for the initial state, the state transition distribution $\mbb P$,
the mean reward model $r$,
and the (additive reward) noise distribution $\mbb Q$.
The questions of interest would govern the choice of the model $\mc M$. For example, to investigate the interestingness of type $1$ in HeartSteps (\cref{fig:interestingUser}(a)), we resample using a generative model for which sending an activity suggestion has no advantage in any state. For simplicity in our notation, we also include the initial policy $\pi_1$ for the RL algorithm as part of the model $\mc M$.  We call this procedure to resample a user trajectory as \resimtraj.

 Next, we use \resimtraj to generate new trajectories for each of the users multiple (say $\numbootstrap$) times. For each user, we use a suitable generative model $\mc M$ depending on the type of interestingness to generate state and rewards and resample treatments using the RL algorithm. Notably, for a given question and user, the generative model remains fixed across the $\numbootstrap$ resamples. For each trajectory, the RL algorithm is initialized and updated in an online manner at each iteration within the trajectory. We call each such trajectory a \emph{resampled trajectory}. 

\begin{algorithm2e}[t!]
\SetKwInOut{Input}{Input}
\SetKwInOut{Output}{Output}
\small{
 \Input{RL algorithm, user models $\sparenth{\mc M_{i}}_{i=1}^{n}$, simulator \resimtraj, resim count $\numbootstrap$
 }
 
 \Output{$\numbootstrap$ resampled trajectories for each user}

  \For{$i= 1$ \KwTo $n$}
  {

    \For { $b= 1$ \KwTo $\numbootstrap$} {
    
        $\mc U_i^{(b)}\!\gets \resimtraj(\mc M_i, \trm{RL Algorithm})$
        \setcomment{Resample $i$-th user trajectory}
    }

  }
  \Return $((\mc U_i^{(b)})_{b=1}^{\numbootstrap})_{i=1}^n$
  \\\hrule
\SetKwProg{myproc}{Algorithm}{}{}
     \myproc{\resimtraj(Initial state distribution $\rho_1$, 
     State transition model $\P$, Mean reward model $r$, Noise model $\Q$, Initial policy $\pi_1$, RL Algorithm)}{
 $S_1^\resimtag \sim \rho_1$ \\
 \For{$t= 1$ \KwTo $T$}
  {
    Sample action $A_t^\resimtag \sim \pi_t(\cdot\vert S_{t}^{\resimtag})$ \\ 
    Generate reward $R_t^\resimtag = r(S_t^\resimtag, A_t^\resimtag) + \vareps_{t}^{\resimtag}$ where $\vareps_t^{\resimtag}\sim \Q$ \\ 
    Generate state $S_{t+1}^\resimtag \sim \P(\cdot\vert S_{t}^{\resimtag}, A_{t}^{\resimtag})$ \\
    Update RL algorithm's policy to $\pi_{t+1}$ \setcomment{update advantage forecasts to $\hat{\Delta}^{\resimtag}_{t+1}$}
  }
     }
     \Return $(S_t^\resimtag,A_t^\resimtag, R_t^\resimtag, \hat{\Delta}_{t}^\resimtag)_{t=1}^{T}$
\caption{Resampling user trajectories using \resimtraj}\label{resim}
}
\end{algorithm2e}

 \subsubsection{Answering \cref{ques:numint}: Number of interesting users} To answer population-level questions like \cref{ques:numint}, we use resampled trajectories of all users (under suitable generative models) to generate a distribution for \numint[] due to algorithmic stochasticity. We then compute how extreme the observed \numint[] is with respect to this distribution via the fraction
\begin{align}
\label{eq:percentile_numint}
    \frac{1}{\numbootstrap} \sum_{b=1}^{\numbootstrap} \indicator(\numint[]\leq \numint[]^{(b)})
    \qtext{where} \numint[]^{(b)} \defeq \sum_{i=1}^{n} \isint[](\mc U_i^{(b)}),
\end{align}
and $\sbraces{\mc U_i^{(b)}}_{b=1}^{\numbootstrap}$ denote the resampled trajectories for user $i$ generated by \cref{resim}. When the fraction~\cref{eq:percentile_numint} is small, we conclude that the data provides evidence that the number of interesting users is not likely to be as high as those observed in the study just by chance (caused by the stochasticity of the RL algorithm) and the RL algorithm is potentially personalizing. On the other hand, when the fraction~\cref{eq:percentile_numint} is large, we conclude that the number of interesting users observed in the study could have arisen due to algorithmic stochasticity.

\subsubsection{Answering \cref{ques:userint}: User-specific interestingness}
To answer a question like \cref{ques:userint} for a fixed user $i$, we use the $\numbootstrap$ resampled trajectories for that user to generate a distribution for \intscore[] due to algorithmic stochasticity. We then compute how extreme the observed \intscore[] for user $i$ is with respect to this distribution via the fraction
\begin{align}
\label{eq:percentile_isint}
    \frac{1}{\numbootstrap} \sum_{b=1}^{\numbootstrap} \indicator\big(\intscore[](\mc U_i) \leq  \intscore[](\mc U_i^{(b)})\big),
\end{align}
where $\mc U_i$ denotes user $i$'s trajectory in the original data and $\sbraces{\mc U_i^{(b)}}_{b=1}^{\numbootstrap}$ denote user $i$'s resampled trajectories from \cref{resim}. 
When the fraction~\cref{eq:percentile_isint} is small, we say that the data suggests that the user $i$ is not likely to be as interesting as that observed in the study just by chance. Or, equivalently, we can conclude that the RL algorithm is potentially personalizing for user $i$. When the fraction~\cref{eq:percentile_isint} is large, we can conclude that the user's interestingness might be an artifact of algorithmic stochasticity.

\section{HeartSteps case study}
\label{sec:heartsteps_intro}
In this section, we take HeartSteps as a case study to illustrate our resampling-based methodology (from \cref{subsec:truth_advertise}). HeartSteps was a mobile health clinical trial with the goal of developing a mobile application to improve physical activity for users with stage I hypertension. Here we focus on one intervention component of the trial, where the goal was to enhance user's physical activity in terms of active step counts, and the reward of the RL algorithm was a monotone (logarithmic) transformation of the users' step counts. Before diving into the case study, we specifically define the states, actions, and rewards in \cref{subsec:sar} that was used by the HeartSteps RL algorithm, which we describe in \cref{subsec:rlalg}. Then, we discuss our choices for the generative model used in \resimtraj for HeartSteps in \cref{subsec:parasim_heartsteps} followed by the resampling results in \cref{subsec:hs_int1,subsec:hs_int2}. 
 
\subsection{State, action, and reward}
\label{subsec:sar}
In HeartSteps, 91 users went through three periods. The first period was one week of wearing a Fitbit tracker. At the end of the first period, the users were instructed to install a mobile application powered by a generalization of the Thompson Sampling algorithm~\cite{Liao2020PersonalizedActivity}. 
The second period was one week of the user interacting with the algorithm programmed to send contextually tailored physical activity notifications with fixed probabilities.
The third period lasted the rest of the study, where the RL algorithm (described in \cref{subsec:rlalg}) was deployed to assign notifications. In our case study, we focus on the data for the first 90 days from the beginning of the second period for each of the 91 users. 

For the second and third periods, there are 5 user-determined decision times per day, representing the user's mid-morning, mid-day, mid-afternoon, mid-evening, and evening. The treatment variable $A_t$ (also referred to as \emph{action} in the RL setup) is binary: $A_t=1$ when a contextually tailored activity suggestion is sent at time $t$, and $A_t=0$ when the activity suggestion is \emph{not} sent at time $t$. The state at time $t$ is defined as $S_t=(I_t, Z_t, X_t)$. Here $I_t$ is the binary \emph{availability} indicator. We set $I_t=0$ when $A_t=0$ (no activity suggestion) is the only ethical or feasible action available at time $t$, and $I_t=1$  otherwise. For example, if the sensors on the phone detect that the person may be driving a vehicle, the availability indicator is set to 0, i.e., $I_t=0$. The vector $Z_t$ denotes the features that represent the current context information  at  decision time $t$; these features include the user's app \engagement score (defined as a function of the number of app screens encountered by the user), step-count \variation indicator (a measure of how active the user has been around the current decision time over the last week), current \location indicator, prior 30-minute step count, yesterday’s daily step count, and current location temperature. For a more detailed description of all the state features, please refer to \cref{tab:state_features} in the Appendix. Finally, $X_t$ denotes the \dosage variable, constructed using user's historical data till time $t$ as follows:
 \begin{align}
\label{eq:dosage}
    X_{t}=\lambda X_{t-1} +  \max(A_{t-1}, B_{t-1}),
\end{align}
where $B_{t-1}$ denotes another binary indicator, corresponding to whether or not an anti-sedentary suggestion (another intervention component in the mobile application) was sent between time $t-1$ and $t$, and $\lambda =0.95$. 
The HeartSteps reward, $R_t$, is  the log-transformed step count in the 30 minute window after the decision time.

\subsection{The RL algorithm}
\label{subsec:rlalg}
We provide an overview of the RL algorithm from HeartSteps and refer the reader to \cite[Sec. 6]{liao2020personalized} for details. 
In HeartSteps, the RL algorithm was initialized with a prior~\cite{Liang2016}, using data from a pilot study~\cite{walter2015}, and deployed separately on each user. Furthermore, it used the data collected during the second period of the study to learn and update its posterior for each user, but did not assign actions. Starting from the third period of the study, it was allowed to assign actions for each user. 
For the action assignment, the algorithm (see \cite[Sec. 2-3]{liao2020personalized}) used a generalization of a contextual linear Thompson-Sampling bandit based on the Gaussian Bayesian linear regression model for the reward~\cite{Liang2016}: 
\begin{align}
\label{eq:working_model}
    R_t = g(S_t)\tp \alpha +  A_t f(S_t)\tp \beta + \varepsilon_t \qtext{if} \avail_t = 1,
\end{align}
with low dimensional parameters $\alpha, \beta$ (to be learned), feature vectors, $f(s), g(s)$; and $\varepsilon_t$ denoting \iid noise from $\mc N(0, \sigma^2)$. 
 The feature vector $g(s)$ includes an intercept, all features that are part of $Z_t$, and the \dosage variable. Meanwhile, the feature vector $f(s)$ includes an intercept, and a subset of the features from $g(s)$, namely \engagement, \variation, \location, and \dosage. For a detailed description of these features, please refer to \cref{tab:state_features}.
 The HeartSteps team selected these features using domain science and  data from the pilot study. Under a bandit environment, that is, when $A_t$ does not affect the distribution of future states, the term $f(s)\tp \beta$ represents the advantage of sending an activity message in state $S_t=s$ with $I_t=1$.

The posterior Gaussian distribution for $\beta$ was updated each night resulting in  the posterior mean $\mu_{d, \beta}$ and variance matrix $\Sigma_{d, \beta}$,  on day $d$.  Thus at decision time $t$ (on day $d = \ceil{\frac{t-1}5}$), for an available user ($\avail_t=1$) in state $S_t=s$ with \dosage $x$, the posterior distribution of advantage of sending an activity suggestion, $\beta^\top f(s)$ is Gaussian, with mean $\mu_{d, \beta}^\top f(s)$ and variance, $f(s)^\top\Sigma_{d, \beta}f(s)$.  In classical Thompson Sampling~\cite{thompson1933likelihood,russo2018tutorial}, actions are selected based on the posterior probability that the advantage exceeds $0$. HeartSteps deployed a slight generalization, replacing $0$ with a  threshold, $\eta_d$, (updated nightly) which was intended to  capture negative delayed effects for sending a treatment at the \dosage level $x$. See \citet[Section 5.4.2]{liao2020personalized} for a discussion connecting this threshold to RL for Markov Decision Processes. Overall, the posterior probability used in HeartSteps is given as:
\begin{align}
\label{eq:hts}
    \P \brackets{\beta^\top f(s) > \eta_{d}(x)
    ; \beta \sim \mc N(\mu_{d, \beta}, \Sigma_{d, \beta})}
    = \Phi(\hat{\Delta}_{d}(s))
     \qtext{where} \hat{\Delta}_{d}(s) = \frac{(\mu_{d, \beta}\tp f(s)-\eta_d(x))}{\sqrt{f(s)\tp \Sigma_{d, \beta}f(s)}} 
\end{align}
denotes the standardized advantage forecast with delayed effects taken into account; here, $\Phi$ is the standard normal distribution function.
In addition, the above posterior probability was clipped to the range $[0.2,0.8]$ using a non-decreasing function $h$ (see \cref{eq:h_clip}); that is the algorithm assigns $A_t = 1$ with probability 
\begin{align}
\label{eq:probability}
    \hat{\pi}_t = 
    \pi(S_t; \mu_d, \Sigma_d, \eta_d) \defeq
     h(\Phi(\hat{\Delta}_{d}(S_t))).
\end{align}
so that the probability of sending $A_t=1$ by the RL algorithm is a monotone function of the standardized advantage forecast $\hat{\Delta}_{d}(S_t)$ in state $S_t$. \\

\begin{remark}
\label{rem:std_advg}
We highlight that the standardized advantage on the vertical axis in \cref{fig:interestingUser} is equal to $\hat{\Delta}_{d}(S_t)$ defined in \cref{eq:hts}. Note that the RL algorithm uses the probability to assign treatments starting from the third period in the study and the horizontal axis in \cref{fig:interestingUser} starts on the first day of the user's second period (in which treatments are assigned using constant probabilities), which lasts 7 days. Hence, the probability of treatment is a monotone function of the standardized advantage forecast $\hat{\Delta}_{d}(S_t)$ starting day 8 for the two panels in \cref{fig:interestingUser}.
\end{remark}

\subsection{\resimtraj for HeartSteps} 
\label{subsec:parasim_heartsteps}
For any question of interestingness, we resample each user trajectory $\numbootstrap=500$ times for the second and third periods of the study. We treat the following algorithm as the ``complete intended RL algorithm'': One week of treatment assignment with fixed probabilities followed by the treatment assignment by the RL algorithm. Recall that the RL algorithm is updated in both of these periods.

We now describe the choices for our generative model as needed by \resimtraj: we use a  Gaussian linear reward model~\cref{eq:working_model}---same as that used by the RL algorithm. To obtain this model for a given user, first, we estimate the parameters $(\hat{\alpha}_T, \hat{\beta}_T)$ on its trajectory using $\ell_2$-regularized least squares (see \cref{sub:model_fit}). Let $\hat{\vareps}_{t} \defeq R_{t}- \hat\alpha_T\tp g(S_t) - A_t \hat\beta_T\tp f(S_t)$ denote the residuals from this fitted model. Then we modify this fitted model to generate the mean rewards depending on the interestingness question. For example, when characterizing the interestingness of type $1$, i.e., consistently positive advantage forecasts, we set $\alpha=\hat{\alpha}, \beta=0$ in \cref{eq:working_model} to generate the mean rewards so that there is no advantage for any action in any state. Intuitively, the number of interesting users (\numint[1]~\cref{eq:numint1}) should be smaller when the advantage is zero in all states versus when the advantage is non-zero in some states. When characterizing interestingness of type $2$ for a feature $\var$ (no differential advantage between $\var=1$ and $\var=0$), we set $\alpha = \hat\alpha$, $\beta_{\var^c} = \hat{\beta}_{\var^c}$, and $\beta_{\var}=0$, where $\beta_{\var}$ denotes the coordinate of $\beta$ corresponding to the feature $\var$ and $\beta_{\var^c}$ denotes all other features. Intuitively, the number of interesting users (\numint[2, \var]~\cref{eq:numint2}) should be smaller when the differential advantage for different values of $\var$ is zero versus when the differential advantage is non-zero.

For noise, we set $\hat{\vareps}_{t}^\resimtag=\hat{\vareps}_t$ %
to mimic possible time-varying trends in the misspecification of the assumed model for the real data.

We set all state features, except \dosage, equal to the observed value in the study to match the state distribution in the HeartSteps trial. On each user, the RL algorithm is rerun (resulting in resampled actions) with
the state features $(I_t, Z_t, B_t)$  set equal to the observed value and  the \dosage variable $X_t$ recomputed using \cref{eq:dosage} starting with $X_0=0$. The RL algorithm omits decision times that have missing rewards or states.

Finally, we initialize the RL algorithm policy $\pi_1$ using the same prior parameters $(\mu_1, \Sigma_1)$ (\cref{eq:prior,eq:prior_sigma}) as in the original RL algorithm. During the first 7 days (corresponding to the second period) of the HeartSteps study, the actions are resampled using a constant probability of $\pi_t \equiv 0.25$ as in the original study. For the remaining days, the actions are resampled using the RL algorithm, as in the third period of the HeartSteps study.

We summarize the \resimtraj with these choices in \cref{resimHeartSteps}, where for brevity we describe the resampling only for the third period of the study. With the above choices, the tuple $ \{ (I_t, Z_t, B_t, \hat{\vareps}_t)_{t=1}^{T}, (\alpha, \beta), (\mu_1, \Sigma_1)\}$ suffices to define the input of the generative model $\mc M$ needed by \resimtraj in \cref{resim}. Moreover, the posterior parameters $(\hat\mu^{\resimtag}_{d}, \hat\Sigma^{\resimtag}_{d}, \hat\eta^{\resimtag}_{d})$ on day $d=\ceil{\frac{t-1}{5}}$ suffice to compute the advantage forecast $\hat{\Delta}_{t}$ in \cref{eq:hts} for the resimulated trajectory. We highlight that for a given user, amongst the input arguments to \resimtraj, only the reward model parameters $(\alpha, \beta)$ vary and all other input arguments remain the same across different types of interestingness.

\begin{algorithm2e}[ht!]
\SetKwInOut{Input}{Input}
\SetKwInOut{Output}{Output}

\SetKwComment{tcp}{//}{}
 \Input{Model $\mc M = \{ (I_t, Z_t, B_t, \hat{\vareps}_t)_{t=1}^{T}, (\alpha, \beta), (\mu_1, \Sigma_1, \eta_1)\}$, Thompson Sampling \cref{eq:updates}
 }
 \Output{Resimulated user trajectory $\mc U^\resimtag$}

   Initialize $X_0^\resimtag=0$, $(\hat{\mu}_{1}^\resimtag, \hat{\Sigma}_{1}^\resimtag, \hat\eta_1^\resimtag)\gets (\mu_1, \Sigma_1, \eta_1)$ \\
  \For {$t= 1$ \KwTo $T$} 
  {
    $ d \gets \ceil{(t-1)/5}$ \setcomment{compute the day in the study} \\
    \setcomment{generate state}\\
    $S_t^\resimtag \gets (I_t, Z_t, X_t^{\resimtag})$ where $X_t^\resimtag \gets \lambda X_{t-1}^\resimtag +\max(A_{t-1}^\resimtag, B_{t-1})$ as in \cref{eq:dosage}\\
    \setcomment{proceed with randomization if the user is available at decision time $t$}
    \\
    \If{$\avail_t=1$}{
     \setcomment{sample new action} \\
    $A_t^\resimtag \gets \mathrm{Bernoulli}(\pi(S_t^{\resimtag}; \hat{\mu}_{d}^\resimtag, \hat{\Sigma}_{d}^\resimtag, \hat\eta_{d}^\resimtag))$ \setcomment{computed using \cref{eq:probability}}\\
    \setcomment{generate reward}    \\
    $ R_t^\resimtag \gets  \alpha^\top g(S_t^\resimtag)+ A_{t}^\resimtag \beta^\top f(S_{t}^\resimtag) + \hat{\varepsilon}_t$
    } 
    \setcomment{Update the posterior for the reward model on a daily basis using \cref{eq:updates}}\\
    \If{$t$ is divisible by $5$ and $\sum_{\l=t-4}^{t}\avail_{\l}\geq 1$}{
    $(\hat{\mu}_{d+1}^\resimtag, \hat{\Sigma}_{d+1}^\resimtag, \hat\eta_{d+1}^\resimtag,)\gets \texttt{PosteriorUpdate}(\hat{\mu}_{d}^\resimtag, \hat{\Sigma}_{d}^\resimtag, \hat\eta_{d}^\resimtag, (S_{\l}^\resimtag,A_{\l}^\resimtag, R_{\l}^\resimtag)_{\l=t-4}^{t})$
    }
    }
\Return{$(S_t^\resimtag, A_{t}^\resimtag, R_{t}^\resimtag, (\hat\mu^{\resimtag}_{d}, \hat\Sigma^{\resimtag}_{d}, \hat\eta^{\resimtag}_{d})_{d=\ceil{(t-1)/5}} )_{t = 1}^{T}$}
\caption{\small{\resimtraj for HeartSteps}}\label{resimHeartSteps}
\end{algorithm2e}

\subsection{Results for interestingness of type $1$}
\label{subsec:hs_int1}

Recall that to assess interestingness of type 1, our procedure proceeds by comparing personalization metrics (\numint[1], \intscore[1]) observed in the real data from the study with those observed on resimulated trajectories, which in turn are resampled by the RL algorithm with a generative model that has no advantage of the action in any state. When the metrics from the real data \emph{differ} significantly from the metrics observed on resimulated trajectories, we have greater confidence that our RL algorithm is \emph{not} exhibiting spurious personalization solely due to the stocahsticity in the RL algorithm.

In \cref{fig:numint1_obs_hist}, we plot a histogram of the values of $\intscore[1]$ across all users in the original study. To add stability, we compute the interestingness scores using smoothened advantage forecasts on a daily basis rather than at each decision time. Furthermore, we do not consider users with low availability whose estimates are not reliable (see \cref{sec:int_details,rem:gamma} for details). Consequently, total users in \cref{fig:numint1_obs_hist} are 63 (and not 91). We note that scores away from $0.5$ lead to interesting graphs. So a relatively high count near $0$ or $1$ indicates that many user graphs appear interesting.
 
\begin{figure}[ht!]
\centering
    \begin{tabular}{c}
    \includegraphics[width=0.35\linewidth]{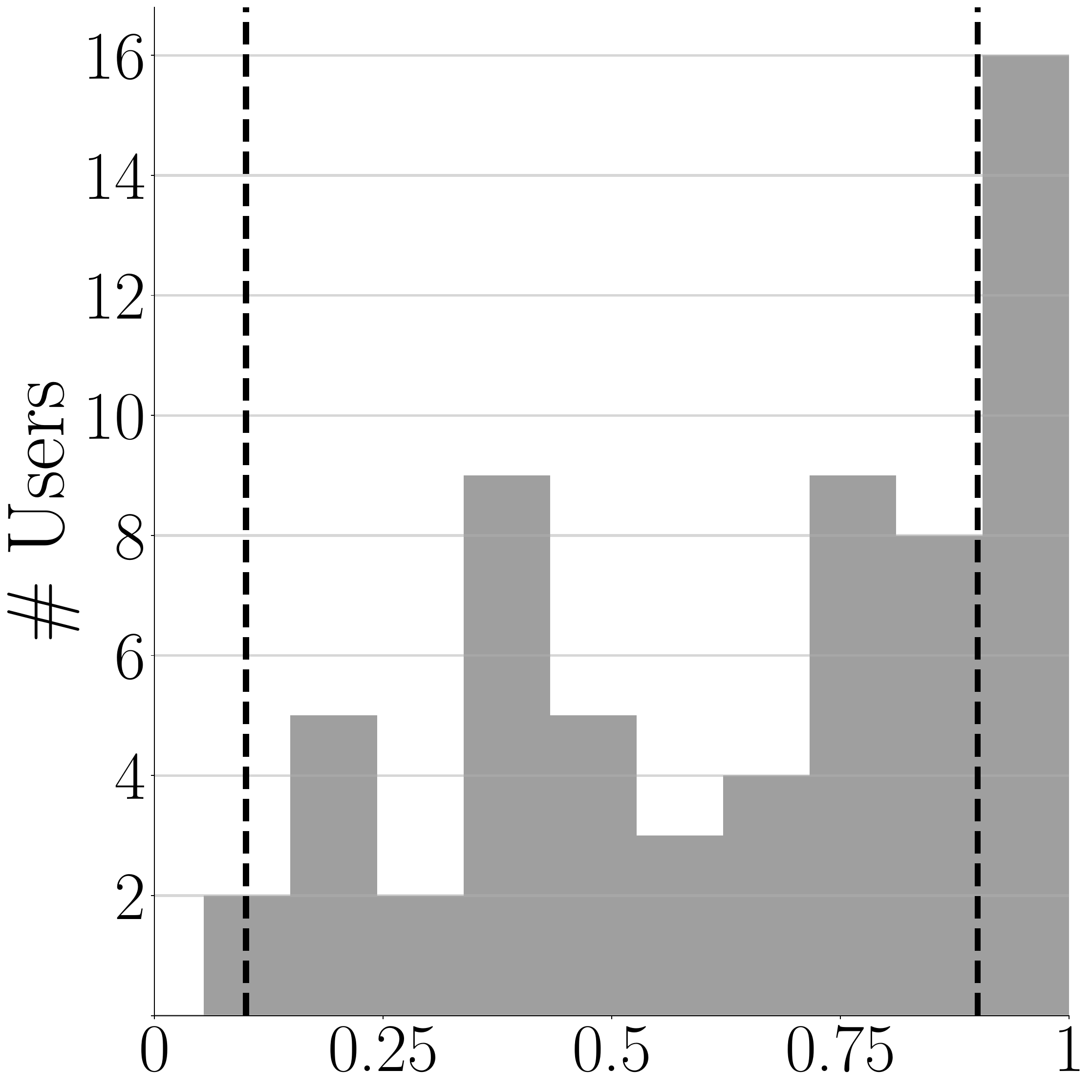}  \\[-2mm]
    \qquad $\intscore[1]$
    \end{tabular}
    \caption{\tbf{Histograms of \intscore[1]~\cref{eq:intscore1} for $63$ users in the original data for HeartSteps.} The count on the vertical axis represents the number of users with the value of $\intscore[1]$ on the horizontal axis. The vertical dashed lines denote the cutoffs ($\intscore[1] \geq .9$ or $\intscore[1] \leq .1$) to denote whether a user is interesting, obtained by setting $\delta=0.4$ in \cref{eq:numint1}.
    \label{fig:numint1_obs_hist}}
\end{figure}

We set $\delta=0.4$ in \cref{eq:numint1} to define an interesting user and find that $\numint[1]=18$, i.e., $18$ out of 63 users in \cref{fig:numint1_obs_hist} appear potentially interesting of type $1$.\footnote{We demonstrate the stability of our results to the choice of $\delta$ in \cref{sec:unpack_int1}. Also, note that the number of users with $\intscore[1]\geq 0.9$ and $\intscore[1]\leq 0.1$ are 17 and 1 respectively; these counts are slightly obscured in the histogram~\cref{fig:numint1_obs_hist}.} We might hence conjecture that the RL algorithm seems to personalize by learning that these 18 users benefit from sending the message in nearly all states. We now use our resampling strategy to investigate if the data provides evidence in favor of this conjecture.

\paragraph{Answering \cref{ques:numint}: Number of interesting users}
We start by investigating the number of interesting users for $\numbootstrap=500$ simulated trials generated with an alternate generative model. In particular, every trial is composed of an independently resampled trajectory for each of the $63$ users such that there is no true advantage in any state. Recall a ``resampled trajectory'' for a user is one in which the RL algorithm is rerun, resulting in a resampling of the treatments across time within the user. \cref{fig:numint1_null_hist} shows a histogram for \numint[1] constructed from $500$ trials. \cref{fig:numint1_null_hist} shows that around 20\% of these trials had a more extreme count for the number of interesting users of type $1$ than the count of $18$ observed in the original data (vertical blue dashed line). Thus we can conclude that the data presents evidence in favor that the number of interesting users can be as high as 18 simply due to algorithmic stochasticity. 

To further refine the conclusions from \cref{fig:numint1_null_hist}, we consider one-sided variants of the definition~\cref{eq:numint1} for the number of interesting users and present the respective analogs of \cref{fig:numint1_null_hist} in \cref{fig:int1_oneSided_histograms}. From \cref{fig:int1_oneSided_histograms}, we find that the number of users with $\intscore[1] \geq 0.9$ is $17$ in the original data, which is significantly larger than any value observed in the 500 trials. On the other hand, the number of users with $\intscore[1]\leq 0.1$ is $1$ in the original data, and all the trials have strictly more users with $\intscore[1]\leq 0.1$. (See \cref{sec:unpack_int1} for details.) Putting the pieces together, we can conclude that the data presents evidence in favor that the RL algorithm is potentially personalizing by learning that many users benefit from sending an activity message. Moreover, the number of interesting users with $\intscore[1] \geq 0.9$ would not be as high as 17 (the observed value in the original data) just due to algorithmic stochasticity. However, many users might exhibit $\intscore[1] \leq 0.1$ (i.e., sending the message is less beneficial than not sending for them) just due to chance and hence we might see that the number of interesting users with $|\intscore[1]-0.5| \geq 0.4$ can be as high as 18 (the observed value in the original data) due to algorithmic stochasticity.

We highlight that these conclusions are also stable to small perturbations to the hyperparameter choice for $\delta$, as we elaborate in \cref{sec:unpack_int1}.

\begin{figure}[ht!]
\centering
    \begin{tabular}{c}
    \includegraphics[width=0.35\linewidth]{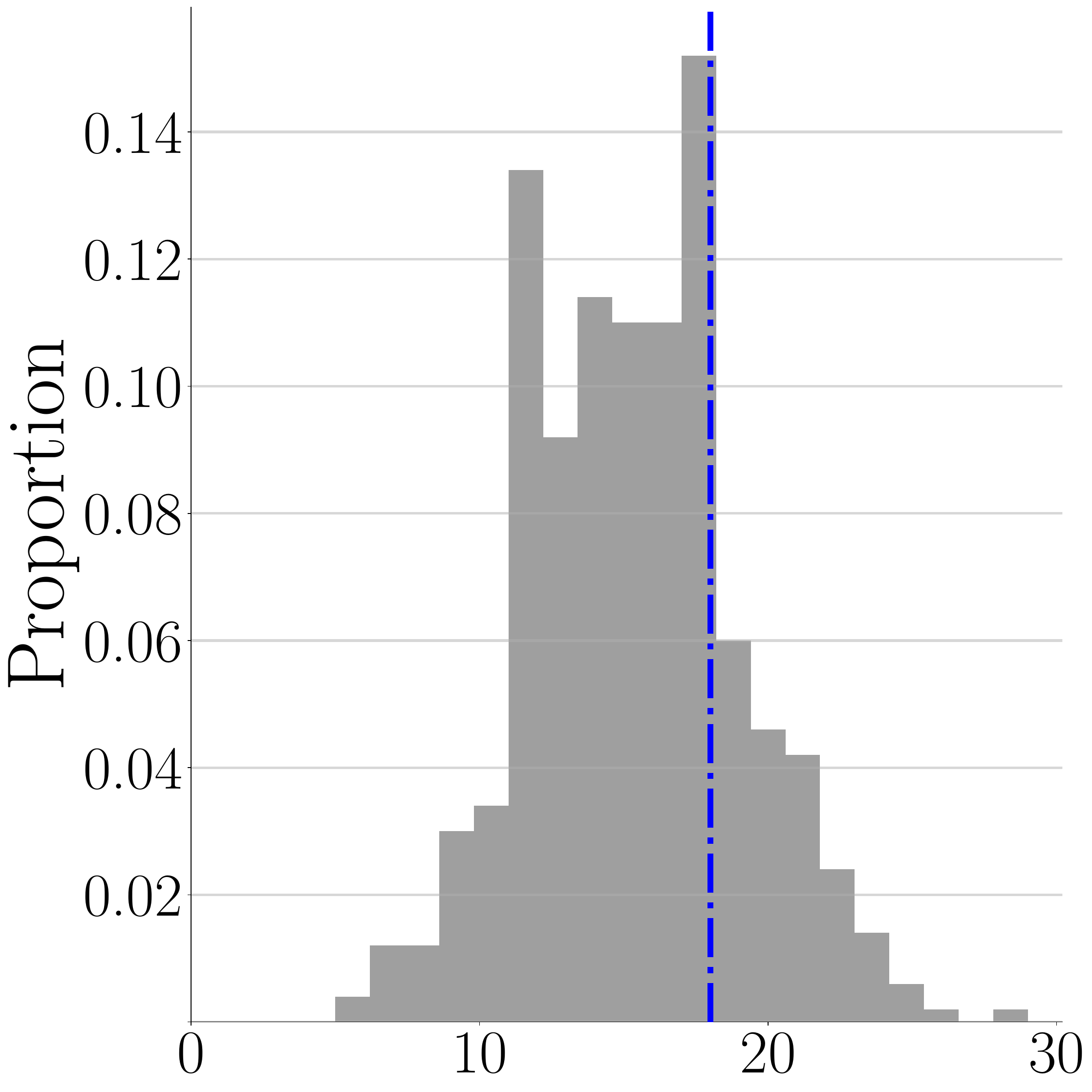}\\[-2mm]
    \qquad $\numint[1]$
    \end{tabular}
    \caption{\tbf{Histogram of $\numint[1]$~\cref{eq:numint1}, the number of interesting users of type $1$, across $500$ trials. 
}  Each trial is composed of 63 resampled trajectories, which are  generated such that  the true advantage is zero.  The proportions on the vertical axis represent the fraction of the 500 trials, with the value of $\numint[1]$ on the horizontal axis.  The vertical blue dashed line marks $\numint[1]=18$, the value observed in the original data.\label{fig:numint1_null_hist}}
\end{figure}

\paragraph{Answering \cref{ques:userint}: User-specific interestingness}
Now, we turn to answer how likely the graph for  User 1 in \cref{fig:interestingUser}(a), with $\intscore[1]=1.0$, would appear just by chance due to algorithmic stochasticity. 
Recall that here the generative model, under which the resampled trajectory is constructed, is that there is no advantage of sending a message for this user. Panels (a) and (b) of  \cref{fig:user_int1} plot two resampled trajectories of user 1 (chosen uniformly from user 1's 500 resampled trajectories) generated under this generative model. For consistency, we use the same color coding as in \cref{fig:interestingUser}, namely, mark the forecast as red if \variation = 1 and blue if \variation = 0. In panel (c) of \cref{fig:user_int1}, we plot the histogram for the \intscore[1] for this user across all 500 resampled trajectories and denote the observed value in the original data as a vertical dotted line.

From the figure, we observe that the two (randomly chosen) resampled trajectories do not appear interesting of type $1$ as in \cref{fig:interestingUser}. The trajectories in panels (a) and (b), respectively, have $\intscore[1]=$ 0.52, and 0.25. (Recall that the $\intscore[1]=1$ for the original trajectory in \cref{fig:interestingUser}(a).) Moreover, panel (c) shows that the interestingness score of $1$, which was observed for the user in the original data, never appears across all the resampled trajectories. Thus we can conclude that the data presents evidence that the RL algorithm potentially personalized for user~1 as the user's observed interestingness score would not likely arise due to algorithmic stochasticity.

\begin{figure}[ht!]
\centering
\begin{tabular}{ccc}
     \includegraphics[width=0.31\textwidth]{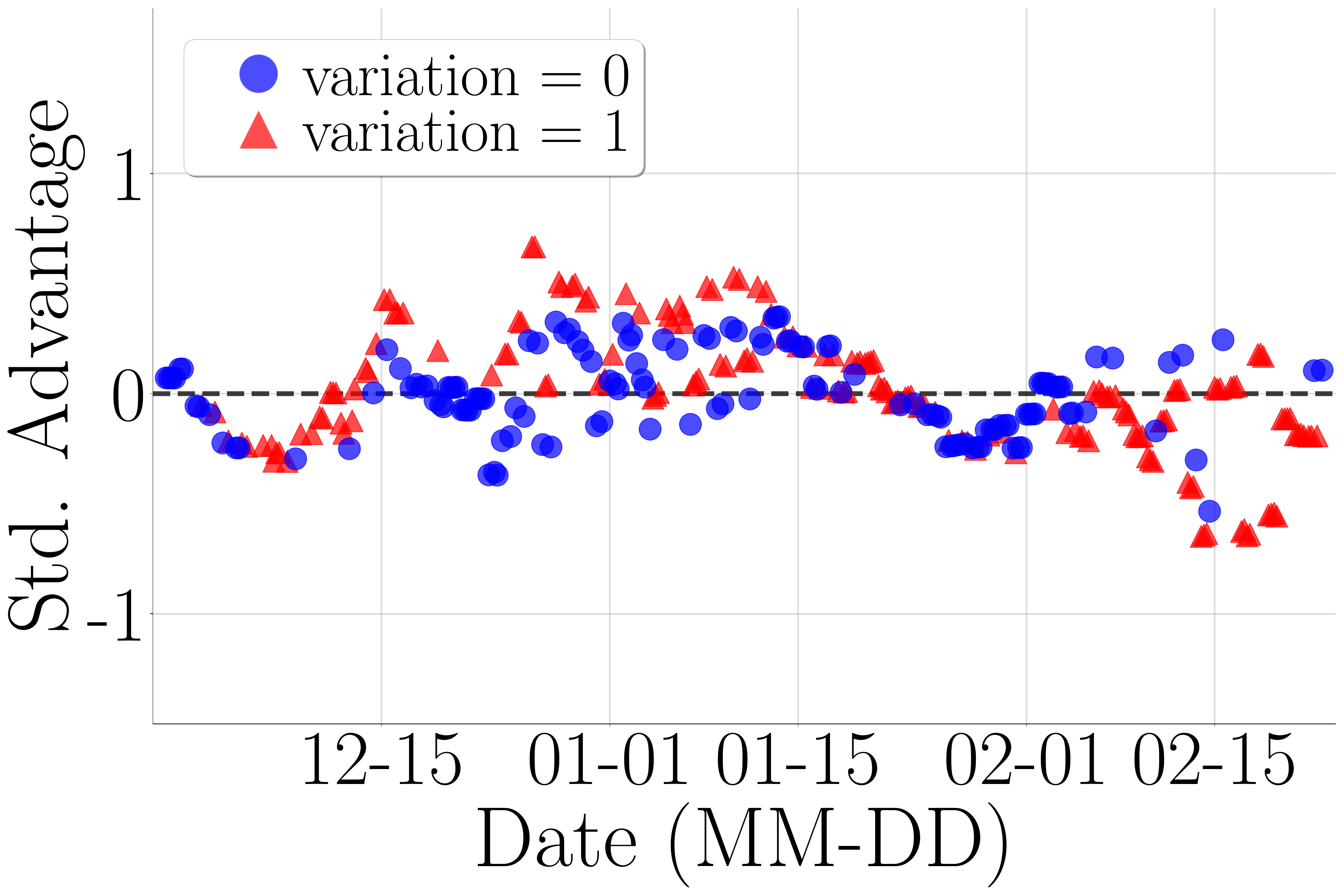} 
     & \includegraphics[width=0.31\textwidth]{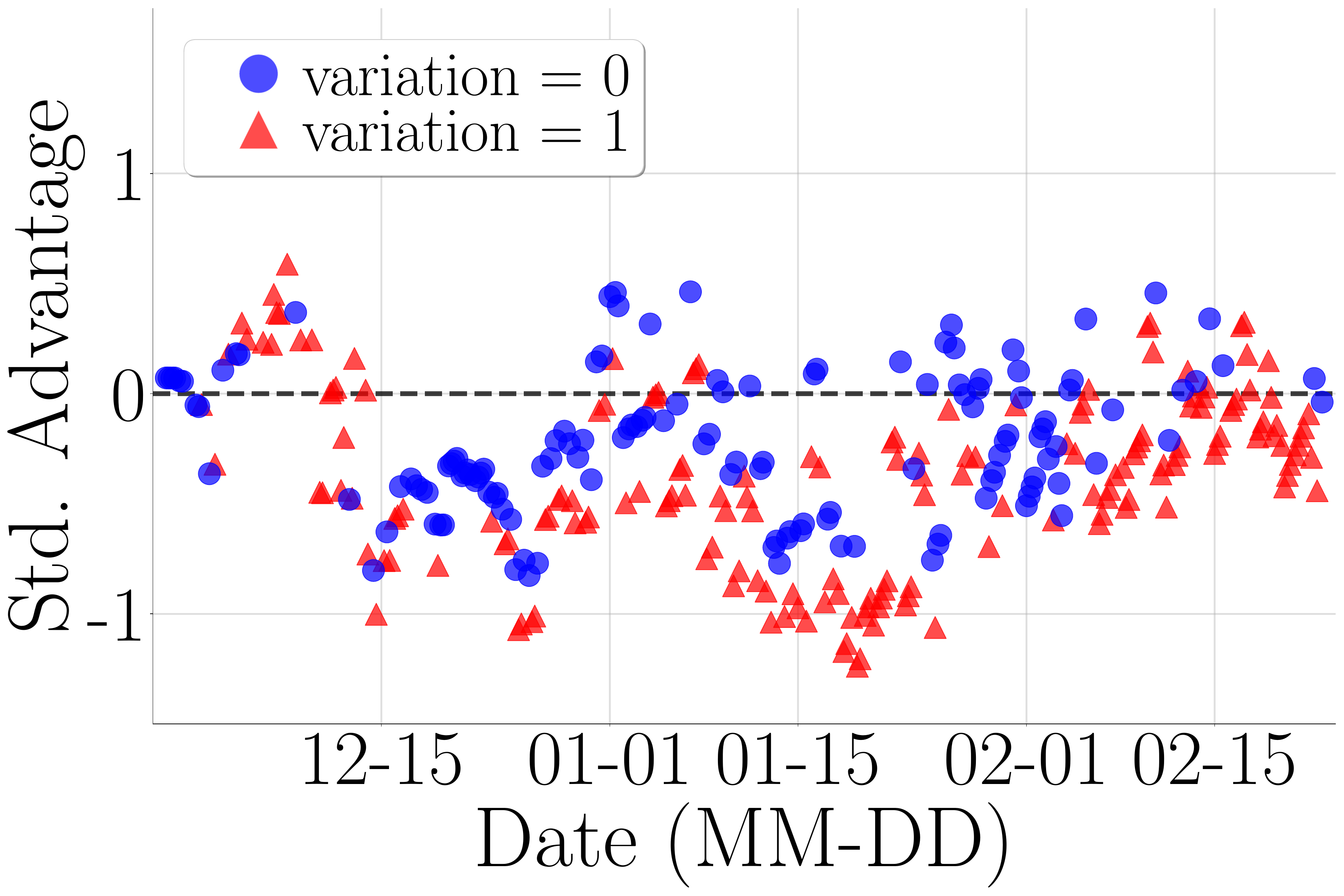} 
     & \includegraphics[width=0.31\textwidth]{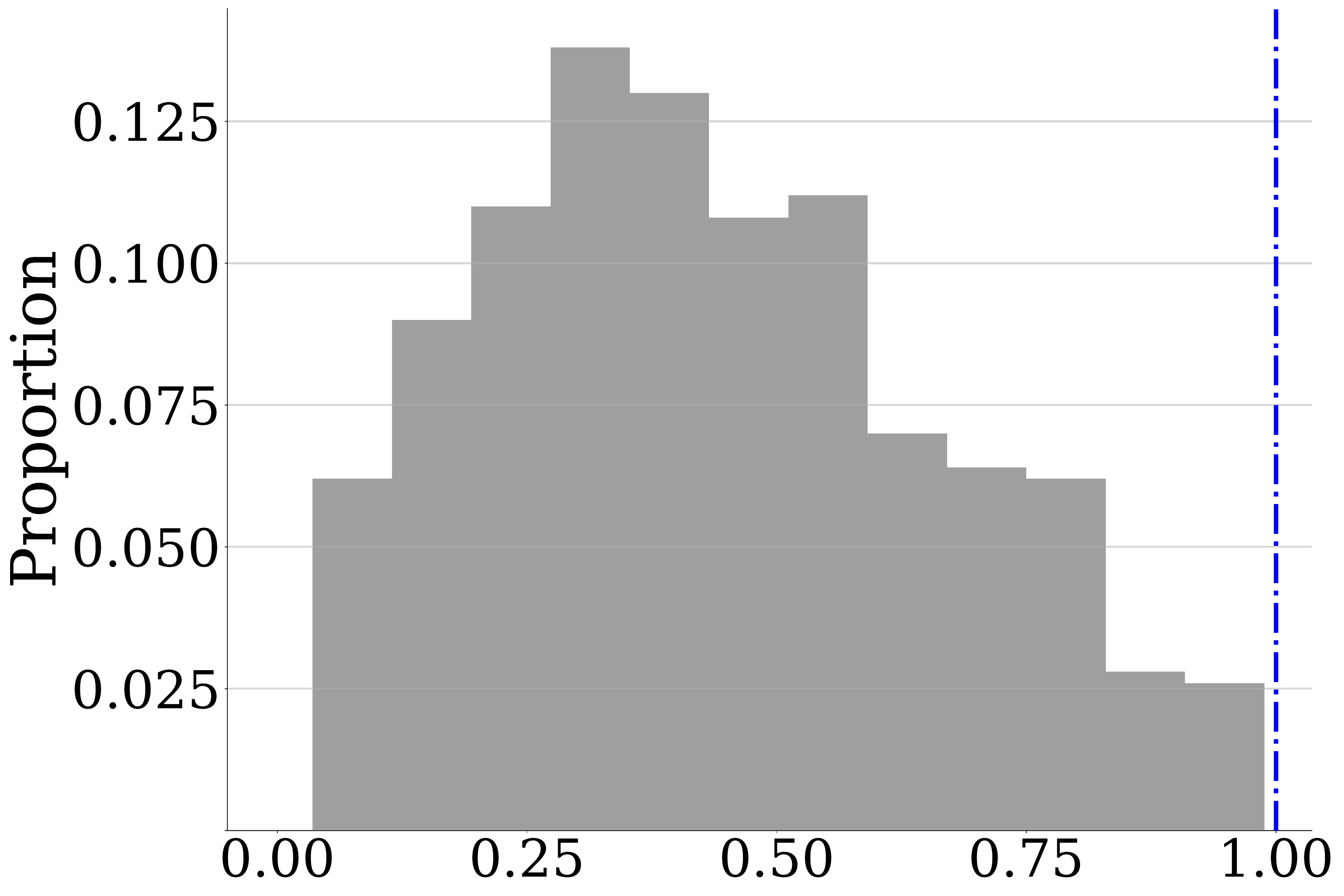}  \\
      \qquad Resampled trajectory 1 &
     \qquad Resampled trajectory 2 &
     \quad \quad $\intscore[1]$ \\
     (a)  & (b)  & (c)
\end{tabular}
\caption{\tbf{Resampling results for user $1$ for interestingness of type $1$.}   Panels (a) and (b) plot two randomly chosen (out of 500) resampled trajectories generated with zero advantage; the two trajectories, respectively, have $\intscore[1]=$ 0.52, and 0.25. In panel (c), the vertical axis represents the fraction of the 500 resampled trajectories for this user with the value of $\intscore[1]$ on the horizontal axis; and the vertical blue dashed line marks $\intscore[1]$ (value 1) for the original trajectory from \cref{fig:interestingUser}(a). \label{fig:user_int1}}
\end{figure}

There are $17$ other users with $|\intscore[1]-0.5|\geq 0.4$, for whom the RL algorithm appears to have personalized by learning which treatment results in better rewards and using this learning to select treatments. To answer if this apparent personalization is simply due to algorithm stochasticity, we compute the fraction~\cref{eq:percentile_isint} of the resampled trajectories for which the $|\intscore[1]-0.5|$ is at least as extreme as the corresponding value in the original data. That is, for user $i$  with trajectory $\mc U_i$ in the original data, we compute the fraction
\begin{align}
\label{eq:p1}
    \lval[1] \defeq \frac{1}{500} \sum_{b=1}^{500} \indicator\big(|\intscore[1](\mc U_i)-0.5| \leq  |\intscore[1](\wtil{\mc U}_i^{(b)})-0.5|)\big), 
\end{align}
where $\wtil{\mc U}_i^{(b)}$ denotes a resampled trajectory for user $i$ under the generative model that there is no advantage in any state.
A lower value of the fraction in \cref{eq:p1} means that the personalization for user $i$ is unlikely to appear just due to the algorithmic stochasticity.

\begin{figure}[ht!]
\centering
    \begin{tabular}{c}
    \includegraphics[width=0.35\linewidth]{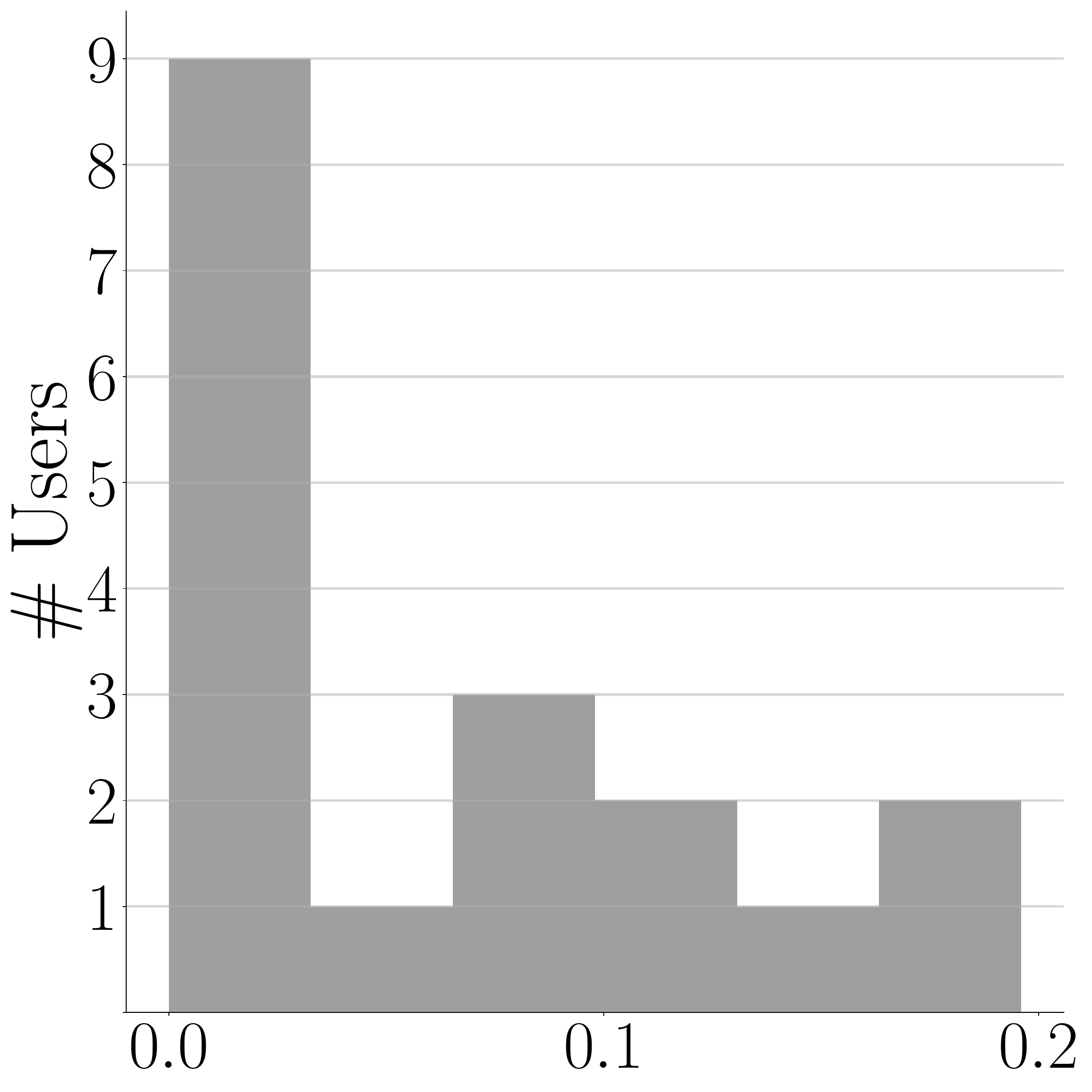}
    \\[-2mm] 
        \quad \quad $\lval[1]$
    \end{tabular}
    \caption{\tbf{Histogram of $\lval[1]$~\cref{eq:p1} for the 18 potentially interesting users with $|\intscore[1]-0.5|\geq 0.4$ in \cref{fig:numint1_obs_hist}.} The count on the vertical axis represents the number of users with the value of $\lval[1]$ on the horizontal axis. \label{fig:int1_pvalues}}
\end{figure}

\cref{fig:int1_pvalues} provides the histograms of $\lval[1]$ for all potentially interesting users, i.e., the 18 users with 
 $|\intscore[1]-0.5|\geq 0.4$ in \cref{fig:numint1_obs_hist}. 
 Out of these 18 users, 13 user exhibit a $\lval[1]$ near $0$, leading us to conclude that these potentially interesting users can indeed be deemed interesting of type $1$ and that they would \emph{not} appear as interesting just due to algorithmic stochasticity. That is, the RL algorithm is potentially personalizing for the users (with small values of $\lval[1]$) by learning when it is advantageous to treat. For the other 5 users, we observe a relatively larger value of $\lval[1]$, leading us to conclude that their interestingness score of type 1 might appear extreme due to algorithmic stochasticity.

\subsection{Results for interestingness of type $2$}
\label{subsec:hs_int2}
Next, we turn to investigate interestingess of type $2$ for different features.

\cref{fig:numint2_obs_hist} provides three histograms for $\intscore[2,\var]$ across all users, one per feature, $\var \in \sbraces{\variation,\location,\engagement}$. As before, to add stability 
we compute the interestingness scores using smoothened advantage forecasts on a daily basis rather than at each decision time. Furthermore, we do not consider users that either have low diversity in the values of feature $\var$ or their estimates are not reliable (see \cref{sec:int_details,rem:gamma} for details); consequently, total users in panels (a), (b), and (c) are, respectively, 60, 12, and 43 (and not 91). We note that scores away from $0.5$ lead to interesting graphs. So a relatively high count near $0$ or $1$ indicates that many user graphs appear interesting.

We set $\delta=0.4$ in \cref{eq:numint2} to quantitatively identify an interesting user; that is, we call a user interesting if their score satisfies $|\intscore[2, \var]-0.5|\geq 0.4$. We mark these cutoffs in \cref{fig:numint2_obs_hist} as vertical dotted lines in each of the panels. (We check the stability of the results to follow with respect to the choice of $\delta$ in \cref{sec:additional_results_heartsteps}.) 

\begin{figure}[ht!]
\centering
    \begin{tabular}{ccc}
     \includegraphics[width=0.31\textwidth,trim={0 0.0cm 0 0},clip]{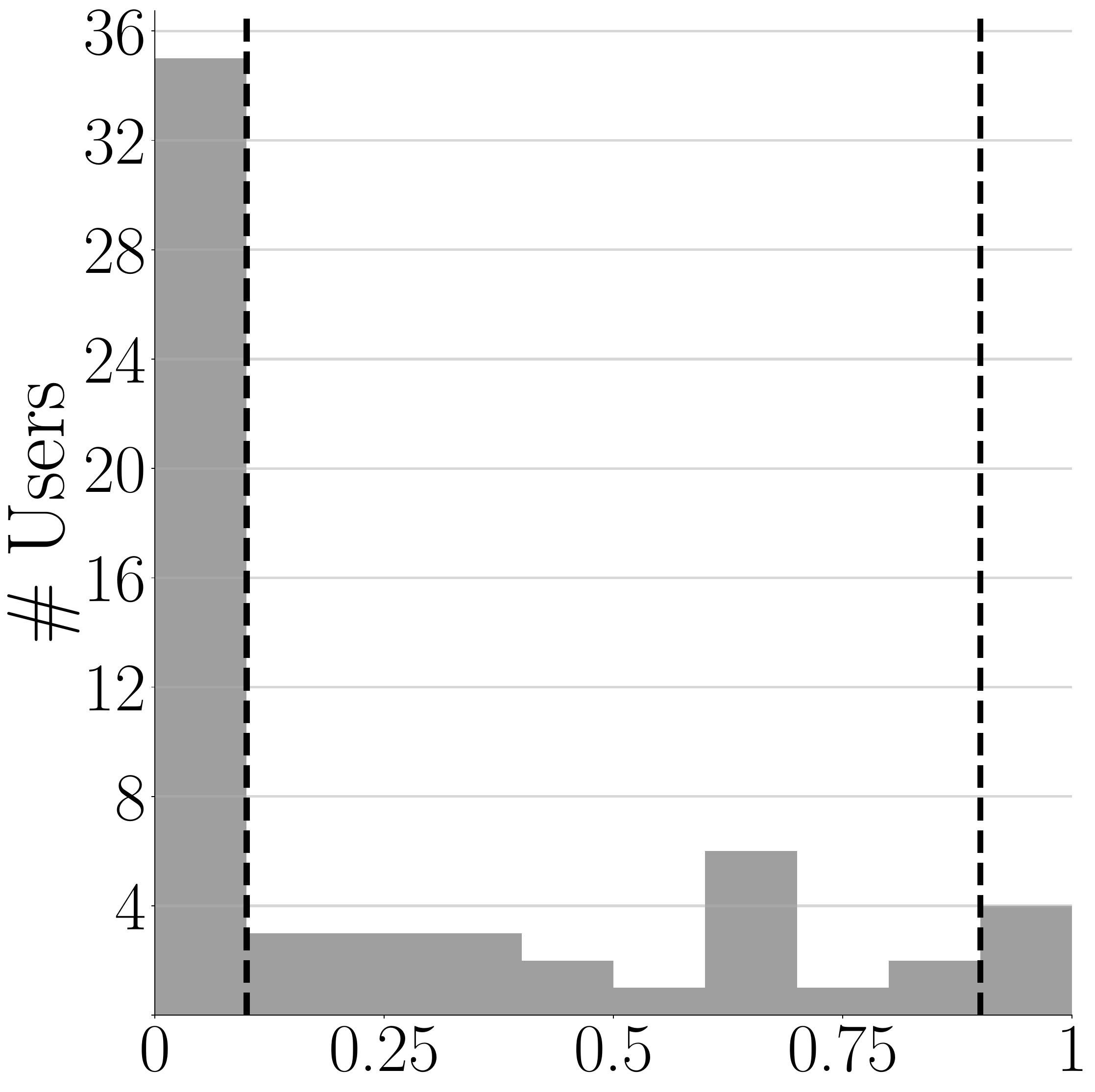} & \includegraphics[width=0.31\textwidth,trim={0 0.0cm 0 0},clip]{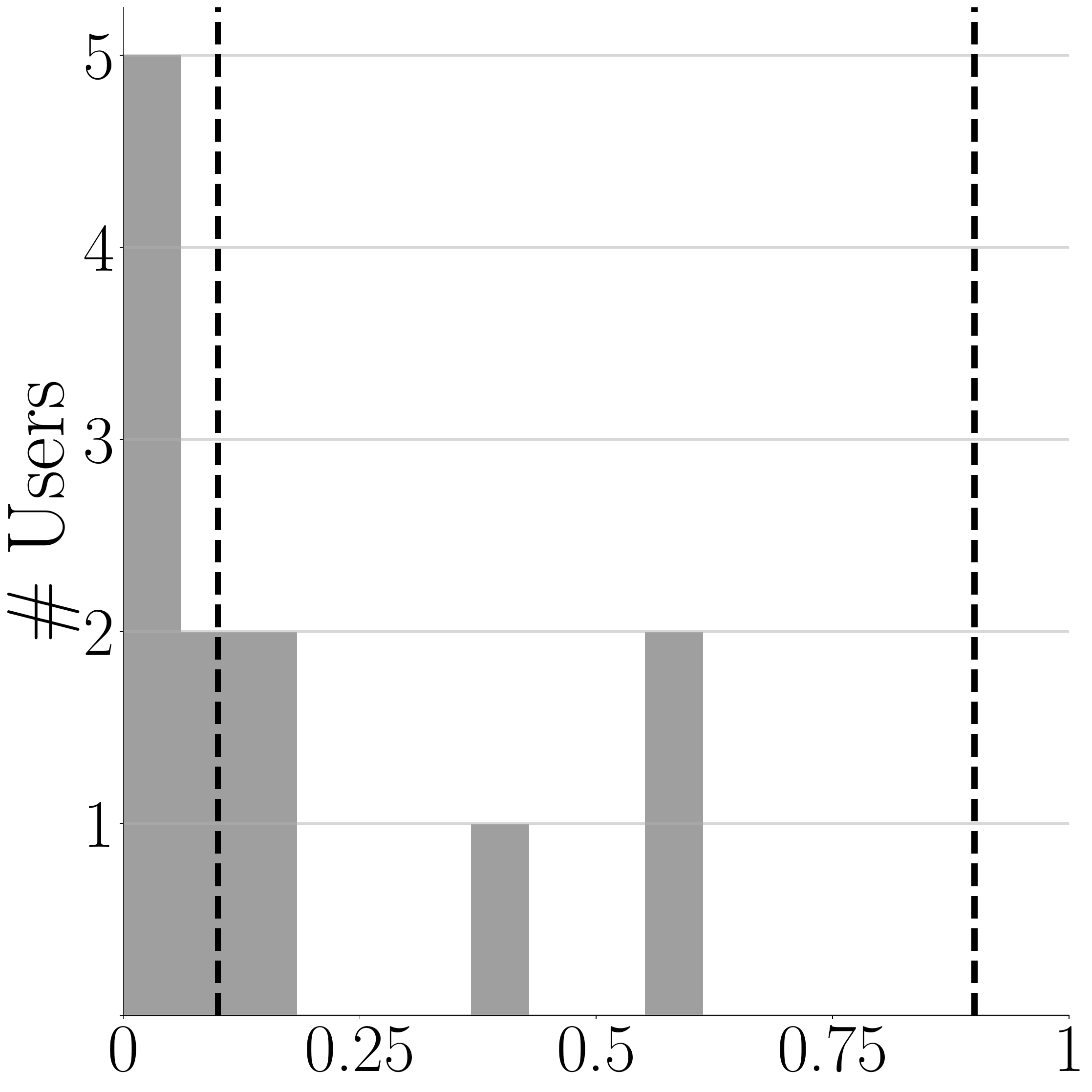}  
     & \includegraphics[width=0.31\textwidth,trim={0 0.0cm 0 0},clip]{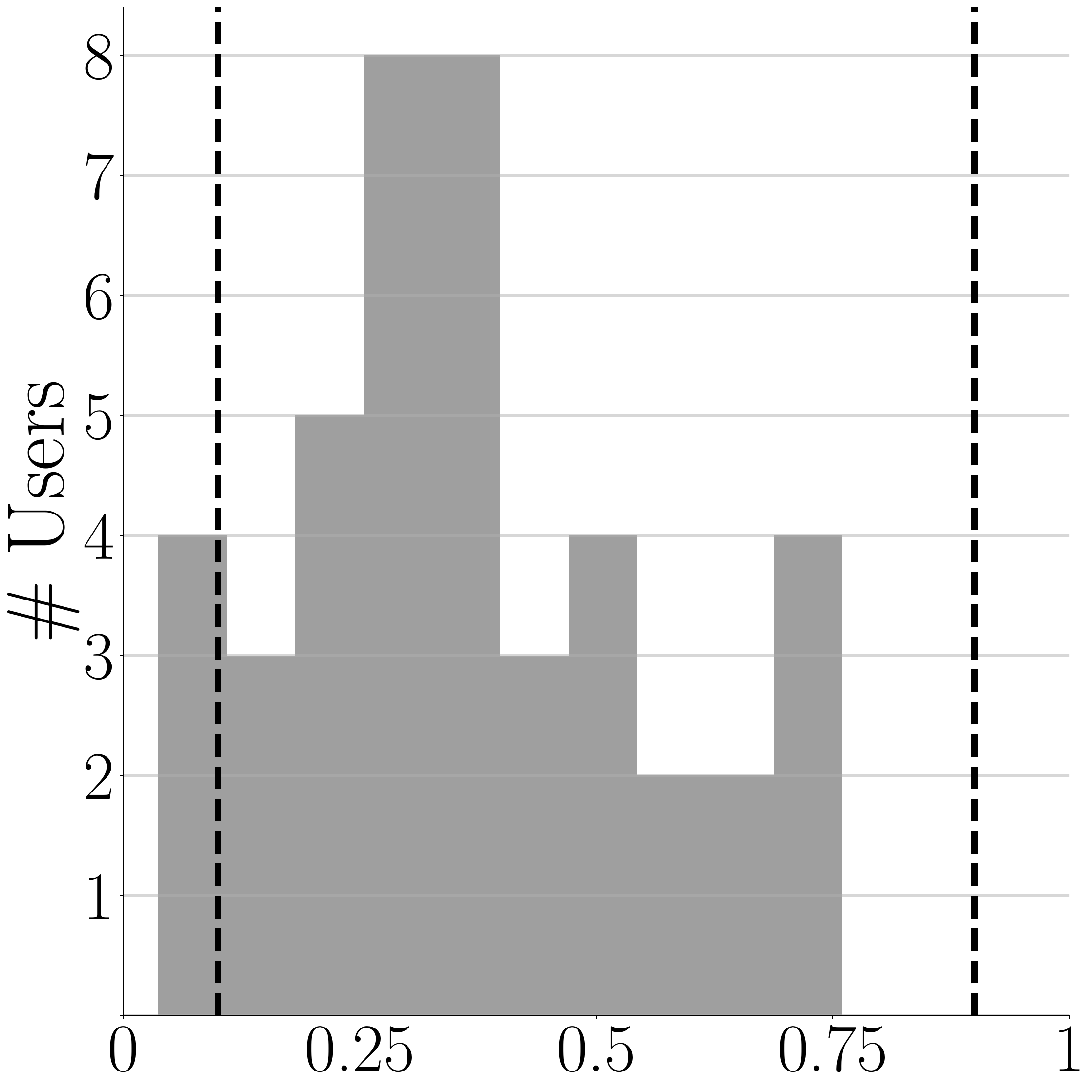} 
    \\[-1mm] 
        \quad \quad $\intscore[2,\variation]$
        & \quad \quad  $\intscore[2,\location]$
        &  \quad \quad $\intscore[2,\engagement]$ \\ %
        (a) & (b) & (c)
    \end{tabular}
    \caption{\tbf{Histograms of \intscore[2, \var]~\cref{eq:intscore2} for the users in the original data for HeartSteps.} The count on the vertical axis represents the number of users with the value of $\intscore[2, \var]$ on the horizontal axis. The vertical dashed lines denote the cutoffs ($\intscore[2,\var] \geq .9$ or $\intscore[2,\var] \leq .1$) to denote whether a user is interesting, obtained by setting $\delta=0.4$ in \cref{eq:numint2}. To add stability to these score computations, we do some pre-processing and filtering (see \cref{sec:int_details}), leading to a total of $60$, $12$, and $43$ users in panels (a), (b), and (c), respectively.\label{fig:numint2_obs_hist}}
\end{figure}
 From \cref{fig:numint2_obs_hist}(a), we find that $\numint[2, \variation]=39$, i.e., $39$ out of 60 users appear potentially interesting of type $2$ for feature \variation. We also note that the majority of the users have their $\intscore[2, \variation]$ close to $0$, and only a few users have $\intscore[2, \variation]$ close to $1$. Hence, we conjecture that the RL algorithm is potentially personalizing for these users and learning that they benefit differentially from sending the message when their \variation takes value $0$ (user's past 7-day step count for a daily timeslot is \emph{less} variable than on average; see \cref{tab:state_features}) compared to when it takes value $1$.

Similarly, the disproportionate share of low values for $\intscore[2, \location]$ in \cref{fig:numint2_obs_hist}(b) (total 8 users with $|\intscore[2, \location]-0.5|\geq 0.4$) leads to the conjecture that the RL algorithm is personalizing for these users by learning that they benefit differentially depending on whether their \location takes value $0$ (user is at home/work) as compared to when it takes value $1$ (user is not at home/work).

Finally, the values of $\intscore[2, \engagement]$ are rather spread in \cref{fig:numint2_obs_hist}(c), which suggests that \engagement feature is perhaps not as critical for RL personalization. However, the scores are surprisingly concentrated on the left of $0.5$, suggesting that the RL algorithm is learning that  some users tend to benefit from an activity message when their \engagement $=0$, i.e., when they are less engaged with the mobile app.

Overall, these plots motivate our conjectures that the RL algorithm seems to personalize by learning to  differentially treat the users based on the values of the features \location and \variation. 
We now employ our resampling strategy to investigate whether the data does provide evidence in favor of these conjectures.

\paragraph{Answering \cref{ques:numint}: Number of interesting users and relevance of features for personalization} Next, we consider histograms similar to those in \cref{fig:numint2_obs_hist}, but constructed under  different generative models for the three panels.  That is, the user resampled trajectories are generated under the generative model that the true advantage does not depend on feature $\var$.  
In \cref{fig:numint2_null_hist}, we plot the histograms for \numint[2, \var] constructed from  $\numbootstrap=500$ trials separately for each $\var \in \sbraces{\variation, \location, \engagement}$. Each trial is composed of resampled trajectories for $60$, $12$, and $43$ users, respectively, for \variation, \location, and \engagement. Recall a ``resampled trajectory'' for a user is one in which the RL algorithm is rerun with an alternative generative model, resulting in a resampling of the treatments across time within the user.  

From the panels (a)  and (b) of \cref{fig:numint2_null_hist}, we notice that the user count for interestingness of type $2$ for $\variation$ as well as $\location$ observed in the original HeartSteps study (vertical blue dashed line) is much higher than the maximum count that appears in the resampled data under each of the two different generative models. Thus we can conclude that the data presents evidence in favor that the RL algorithm is potentially personalizing by learning to differentially treat the users based on their \variation and \location values. 

However, the observed number of interesting users for $\var=\engagement$ in the actual HeartSteps study  (vertical blue dashed line) is smaller than that in the resampled datasets under the generative model of no impact of engagement on the advantage.  This  observation suggests that the concentration of users on the left of \cref{fig:numint2_obs_hist}(c) is likely due to stochastic variation in the RL algorithm.  Since the HeartSteps design team had thought that engagement would be a useful feature for personalization, this observation might also  motivate the team to reconsider the definition of the feature \engagement in order to make it effective for personalization. 

We highlight that the above conclusions are also stable to small perturbations to the hyperparameter choice for $\delta$, as we elaborate in \cref{sec:additional_results_heartsteps}.

\begin{figure}[ht!]
\centering
    \begin{tabular}{ccc}
     \includegraphics[width=0.31\textwidth,trim={0 0.0cm 0 0},clip]{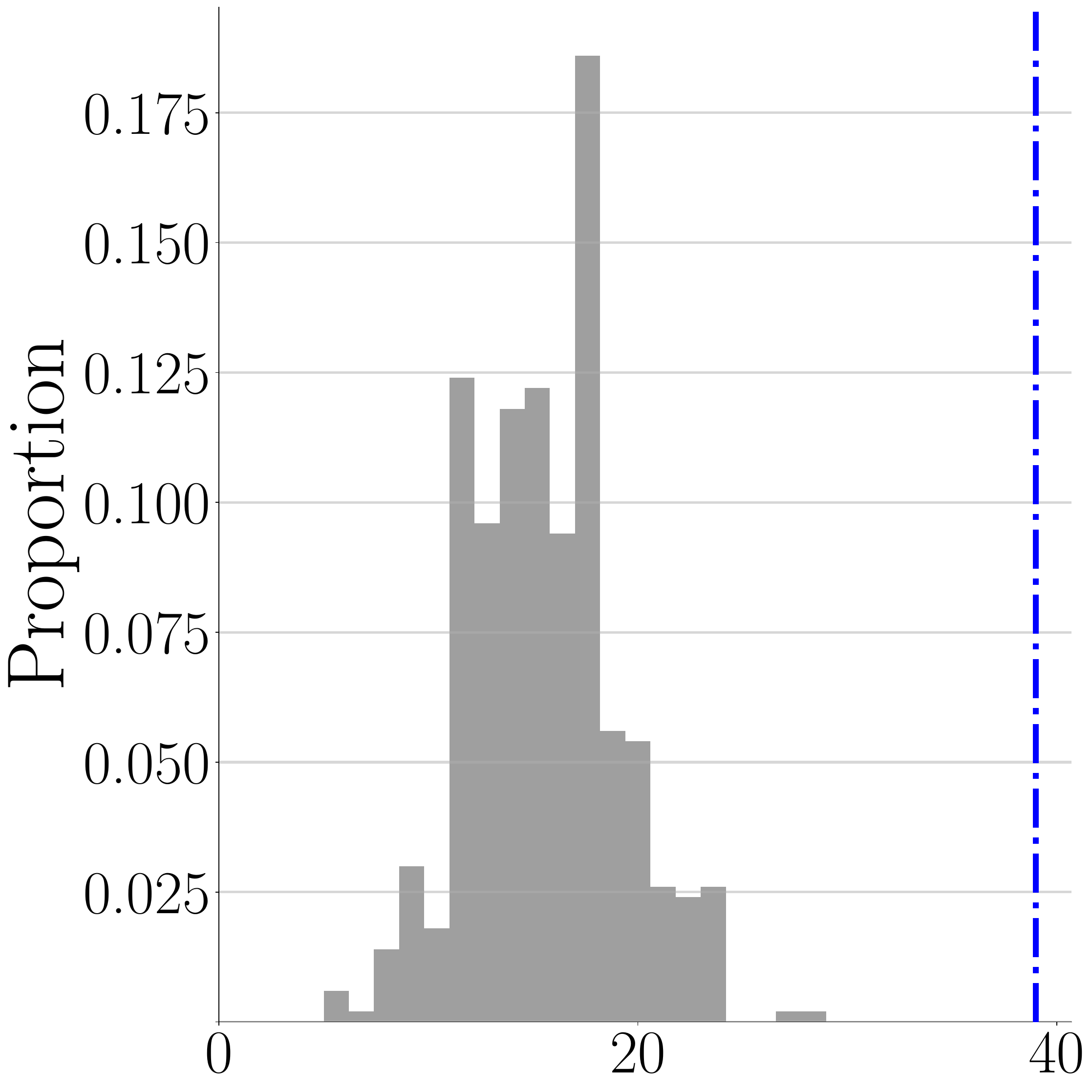}
     &
     \includegraphics[width=0.31\textwidth,trim={0 0.0cm 0 0},clip]{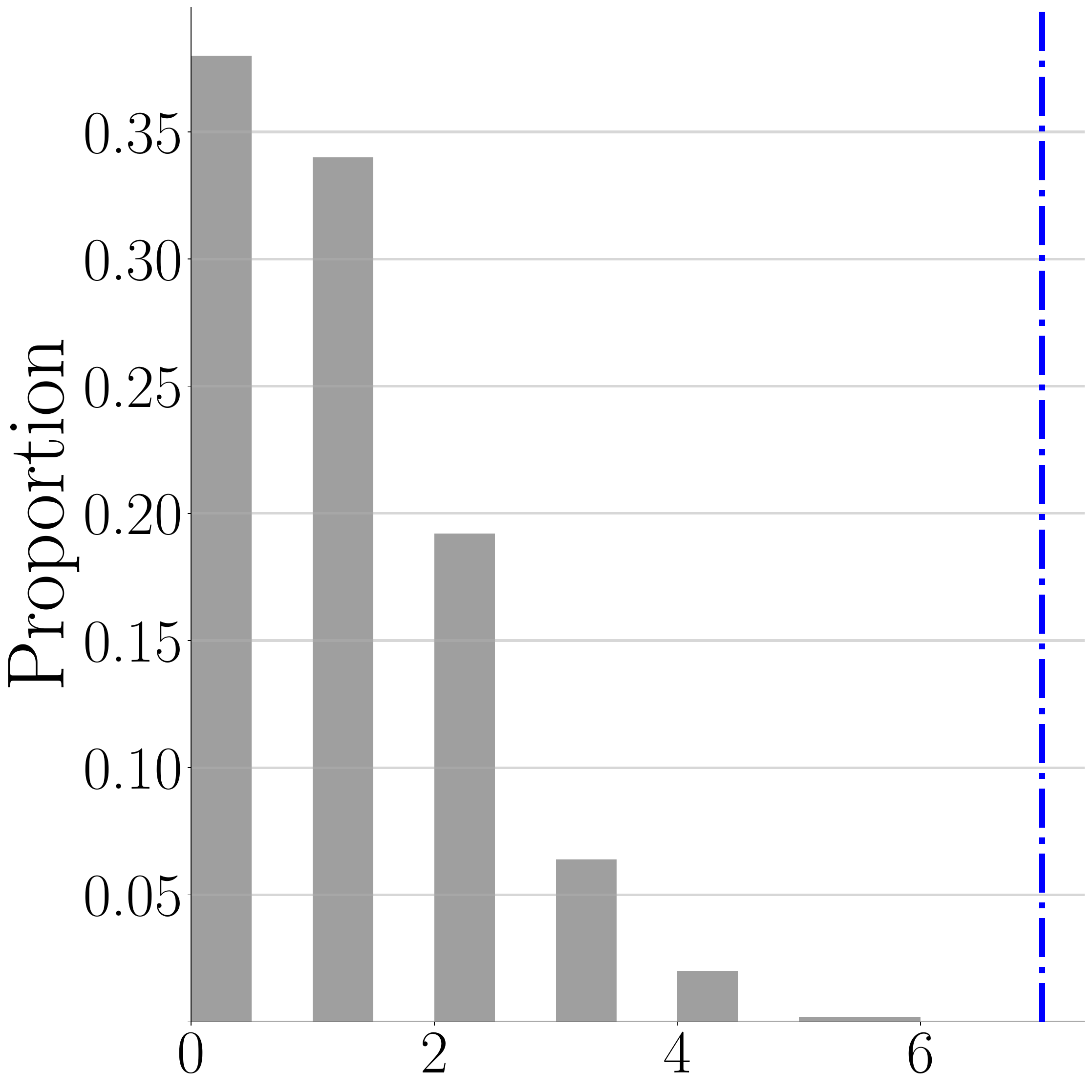}
     & 
     \includegraphics[width=0.31\textwidth,trim={0 0.0cm 0 0},clip]{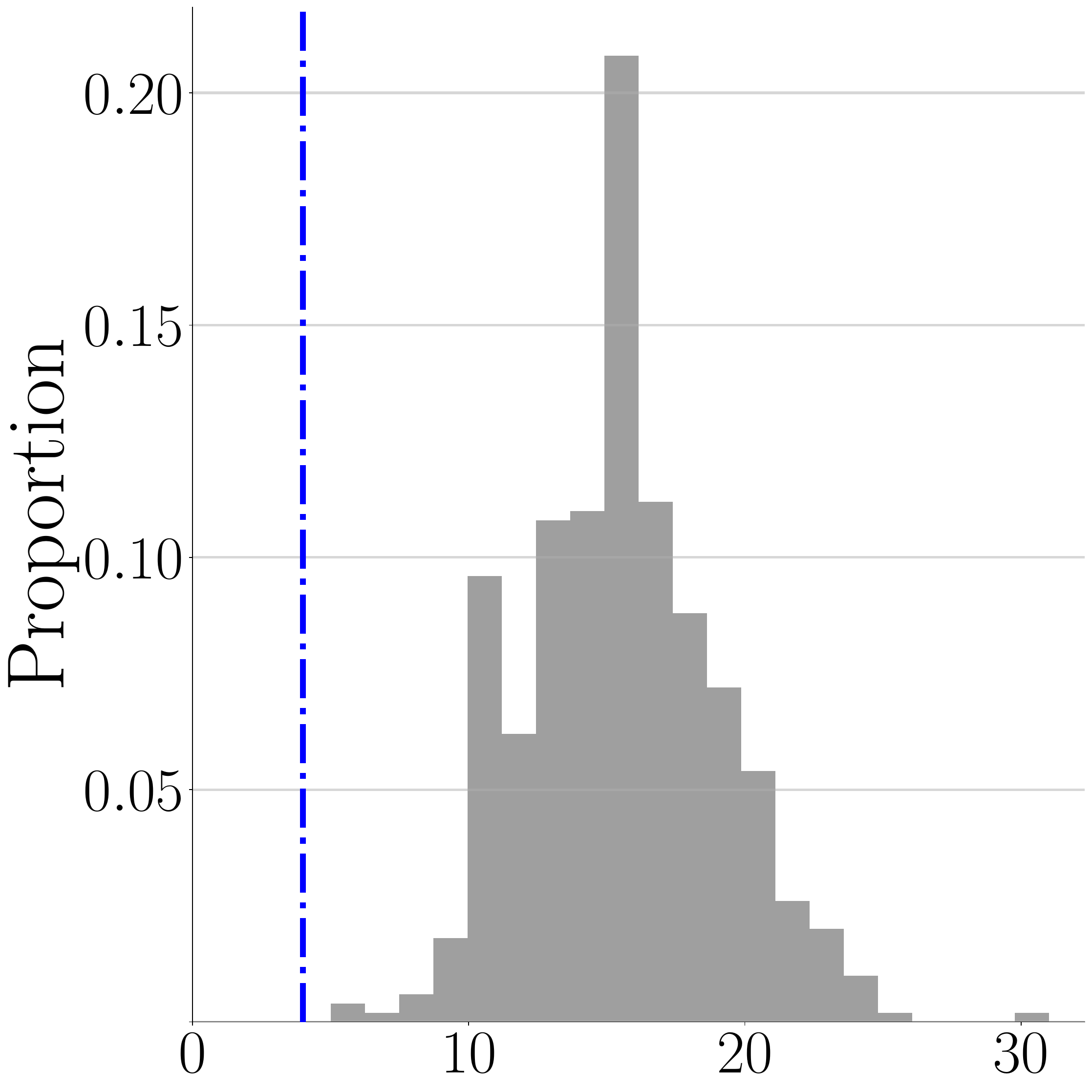} 
     \\[-3mm] 
        \quad \quad \footnotesize{$\numint[2,\variation]$}
        & \quad \quad \footnotesize{$\numint[2,\location]$}
        &  \quad \quad \footnotesize{$\numint[2,\engagement]$ }\\ %
        \footnotesize{(a)} & \footnotesize{(b)} & \footnotesize{(c)}
        \vspace{-3mm}
    \end{tabular}
    \caption{\tbf{Histogram of $\numint[2, \var]$~\cref{eq:numint2}, the number of interesting users of type $2$ for feature $\var$, across $500$ trials.}  Each trial is composed of 60, 12, and 43 resampled trajectories, respectively, in panels (a), (b), and (c).  The trajectories are generated under the generative model that the true advantage does not depend on feature $\var$.  The proportions on the vertical axis represent the fraction of the 500 trials, with the value of $\numint[2, \var]$ on the horizontal axis.  The vertical blue dashed line at 39, 7, and 4, respectively, in panels (a) to (c), denotes the value of $\numint[2, \var]$ observed in the original data. \label{fig:numint2_null_hist}}
\end{figure}

\paragraph{Answering \cref{ques:userint}: User-specific interestingness}
Next, we turn to answer how likely the user graph in \cref{fig:interestingUser}(b) with $\intscore[2,\variation]=0$, would appear just by chance.  Panels (a) and (b) of  \cref{fig:user_int2_variation} visualize two resampled trajectories of user 2 (chosen uniformly from user 2's 500 resampled trajectories) generated under the generative model that there is no differential advantage of sending a message based on the value of \variation. The color coding is as in \cref{fig:interestingUser}, namely, the forecasts are marked in red triangles if \variation = 0 and blue circles if \variation = 1. In panel (c) of \cref{fig:user_int2_variation}, we plot the histogram for the \intscore[2, \variation] for this user across all 500 resampled trajectories and denote the observed value in the original data as a vertical dotted line.  

\cref{fig:user_int2_variation}(a,b) show that the resampled trajectories do not appear interesting of type $2$ for \variation as in \cref{fig:interestingUser}(b). In particular, the trajectories in panels (a) and (b), respectively, have $\intscore[2,\variation]=$ 0.088, and 0.66. Moreover, panel (c) shows that the interestingness score of $0$, which was observed for user 2 in the original data (\cref{fig:interestingUser}(b)), appears only once across all the resampled trajectories.
Thus we can conclude that the data presents evidence that the RL algorithm potentially personalized for user $2$ by learning to treat the user based on \variation differentially, and this personalization would not likely arise simply due to algorithmic stochasticity.

\begin{figure}[ht!]
\centering
\begin{tabular}{ccc}
     \includegraphics[width=0.31\textwidth]{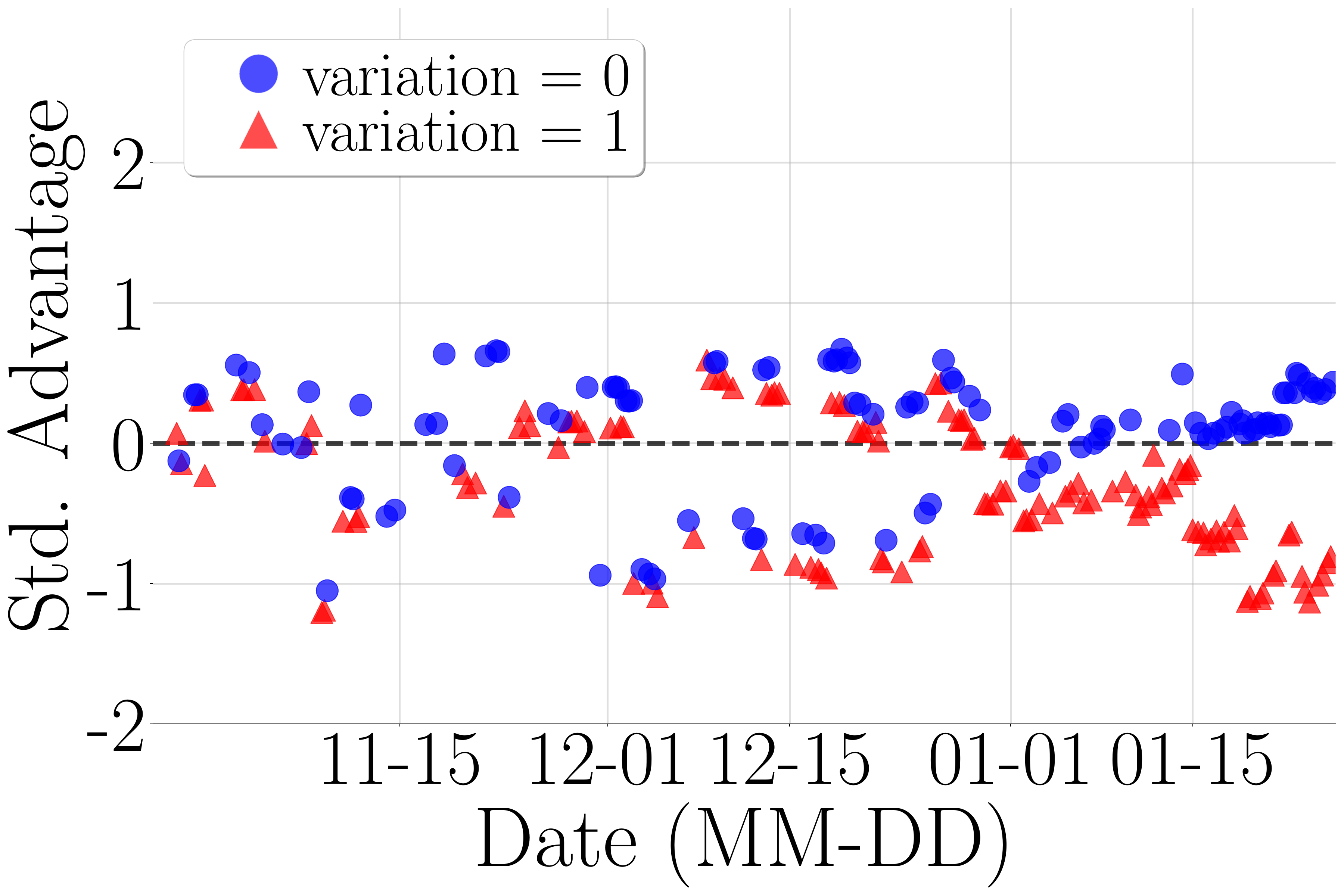} 
     & \includegraphics[width=0.31\textwidth]{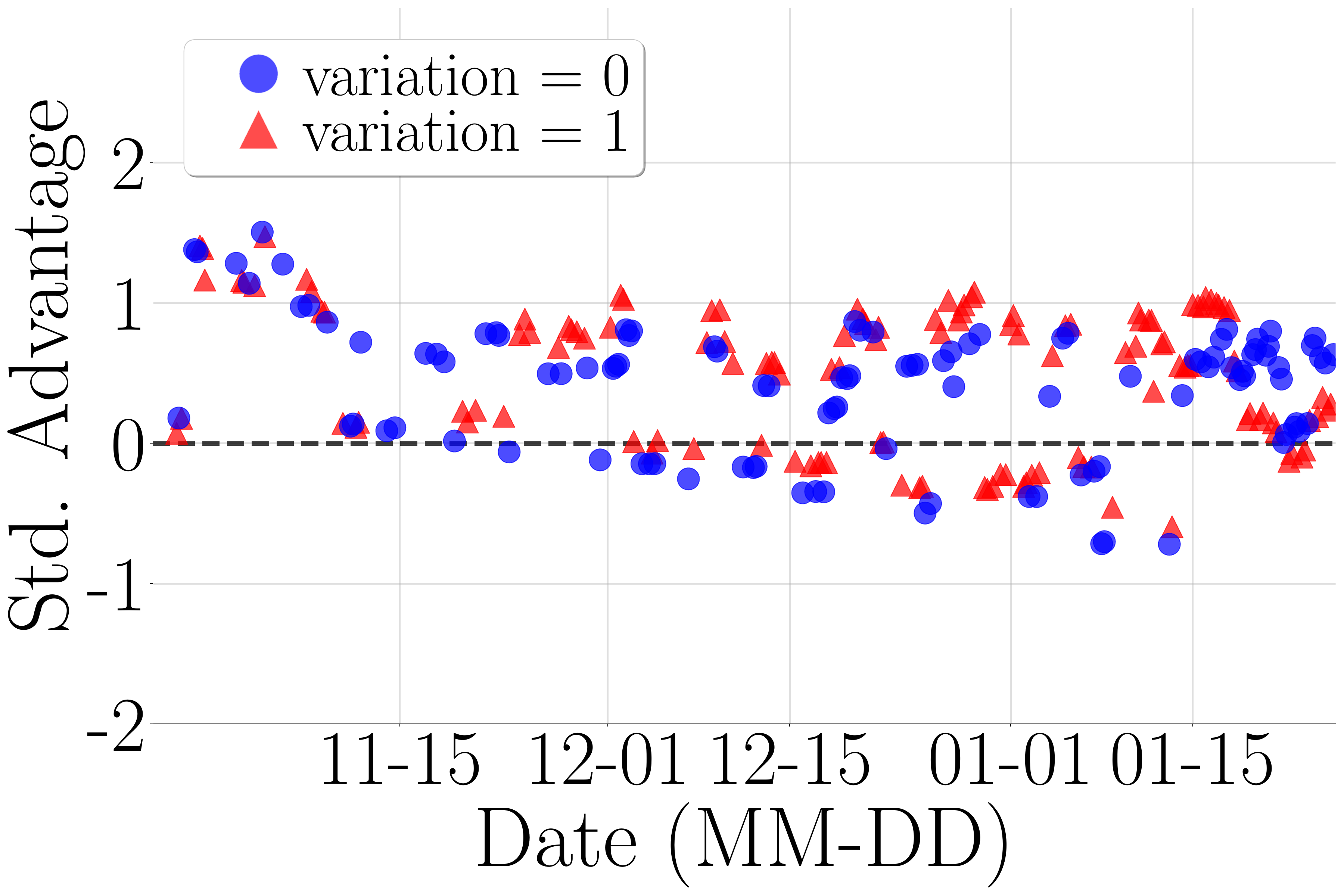} %

     & \includegraphics[width=0.31\textwidth]{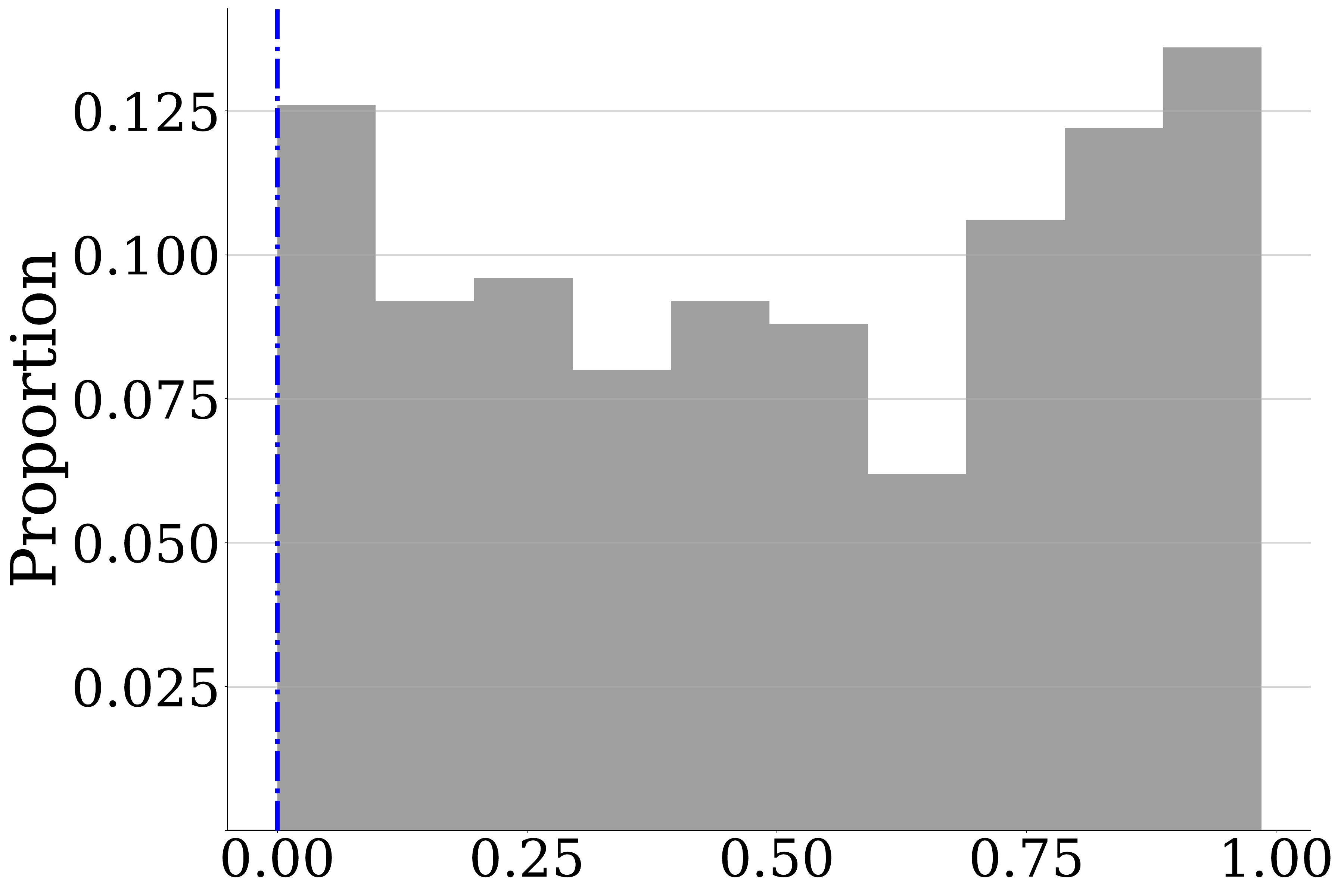}  \\
     \qquad Resampled trajectory 1 &
     \qquad Resampled trajectory 2 &
     \quad \qquad $\intscore[2,\variation]$ \\
     (a)  & (b) & (c)
\end{tabular}
\caption{\tbf{Resampling results  for user $2$ for interestingness of type $2$ for $\var=\variation$.} Panels (a) and (b) plot two randomly chosen (out of 500) resampled trajectories generated under a generative model with zero advantage; the two trajectories, respectively, have $\intscore[2,\variation]=$ 0.088, and 0.66. In panel (c), the vertical axis represents the fraction of the 500 resampled trajectories for this user with the value of $\intscore[2,\variation]$ on the horizontal axis; and the vertical blue dashed line marks $\intscore[2,\variation]$ (value 0) for the original trajectory from \cref{fig:interestingUser}(b). \label{fig:user_int2_variation}}
\end{figure}

To investigate whether other specific users indeed exhibit interestingness of type $2$, for each user we compute the fraction~\cref{eq:percentile_isint} of its resampled trajectories for which the value $|\intscore[2,\var]-0.5|$ is at least as extreme as the user's corresponding value in the original data. That is, for user $i$ with trajectory $\mc U_i$ in the original data, we compute the fraction 
\begin{align}
\label{eq:p2}
    \lval[2,\var] \defeq
        \frac{1}{500} \sum_{b=1}^{500} \indicator\big(|\intscore[2,\var](\mc U_i)-0.5| \leq  |\intscore[2,\var](\wtil{\mc U}_i^{(b)})-0.5|\big),
\end{align}
 where $\wtil{\mc U}_i^{(b)}$ denotes a resampled trajectory  for user $i$ 
 under the generative model that the advantage does not depend on feature $\var$. A lower value of the fraction in \cref{eq:p2} means that the personalization for user $i$ is unlikely to appear just by chance due to the algorithmic stochasticity. For instance, we have $\lval[2,\variation]=0$ for user $2$ from \cref{fig:interestingUser,fig:user_int2_variation}. 

\cref{fig:int2_pvalues} provides the histograms of $\lval[2,\var]$ for 39, 7, and 4 potentially interesting users for $\var=\variation, \location$, and \engagement, respectively, each with $|\intscore[2,\var]-0.5|\geq 0.4$ in \cref{fig:numint2_obs_hist}. The high proportion of values near $0$ from panels (a) and (b) lead us to conclude that several of these potentially interesting users can indeed be deemed interesting of type $2$ for $\var=\variation$ and $\location$ and that they would \emph{not} appear as interesting just due to algorithmic stochasticity. That is, the RL algorithm is potentially personalizing for the users (with small values of $\lval[2, \var]$) by learning to treat them differentially based on the values of these two features. However, from panel (c), we find that for all 4 potentially interesting users of type 2 for \engagement, their interestingness score might appear extreme simply due to algorithmic stochasticity.

\begin{figure}[ht!]
\centering
    \begin{tabular}{ccc}
        \includegraphics[width=0.31\linewidth,trim={0 0cm 0 0},clip]{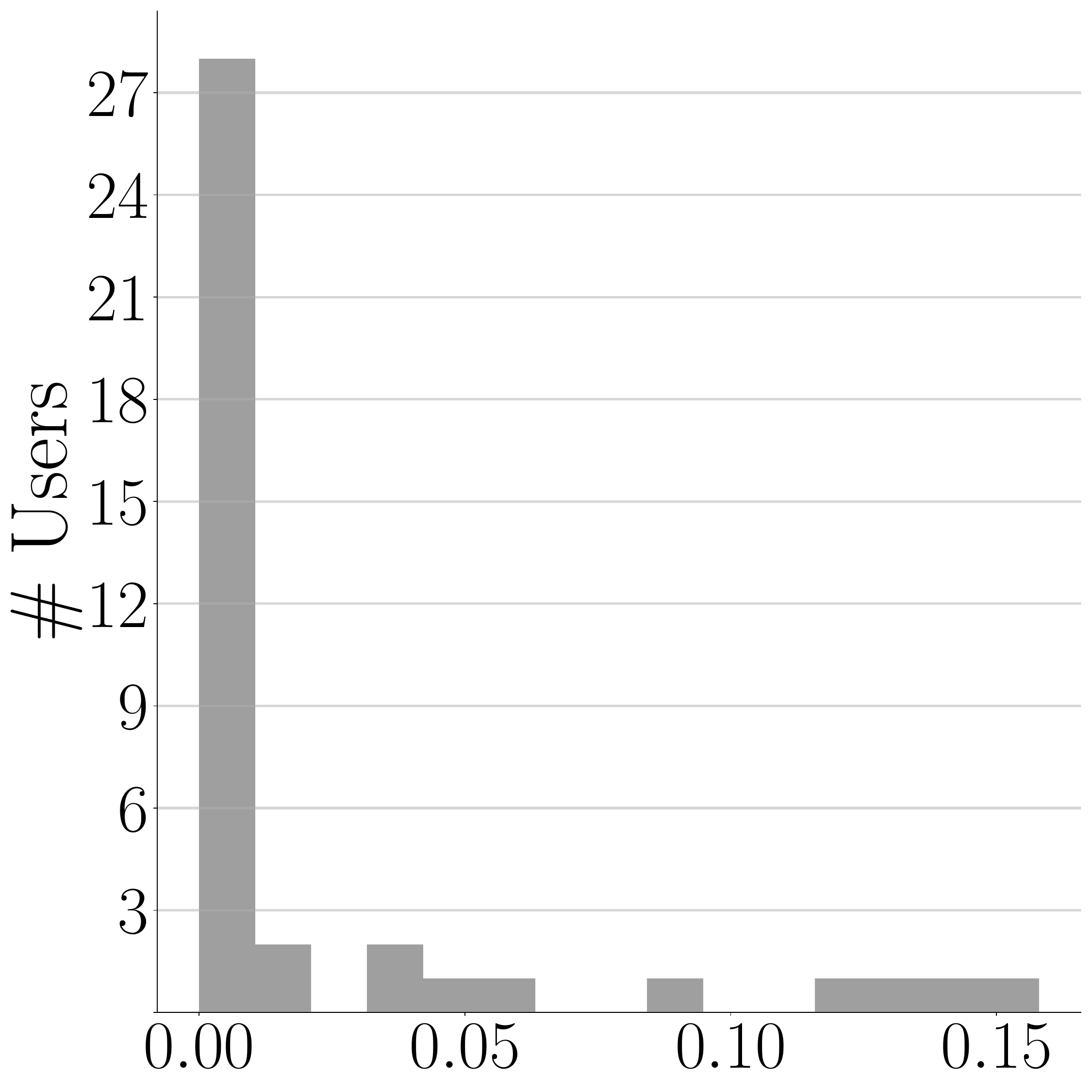} 
       & \includegraphics[width=0.31\linewidth,trim={0 0cm 0 0},clip]{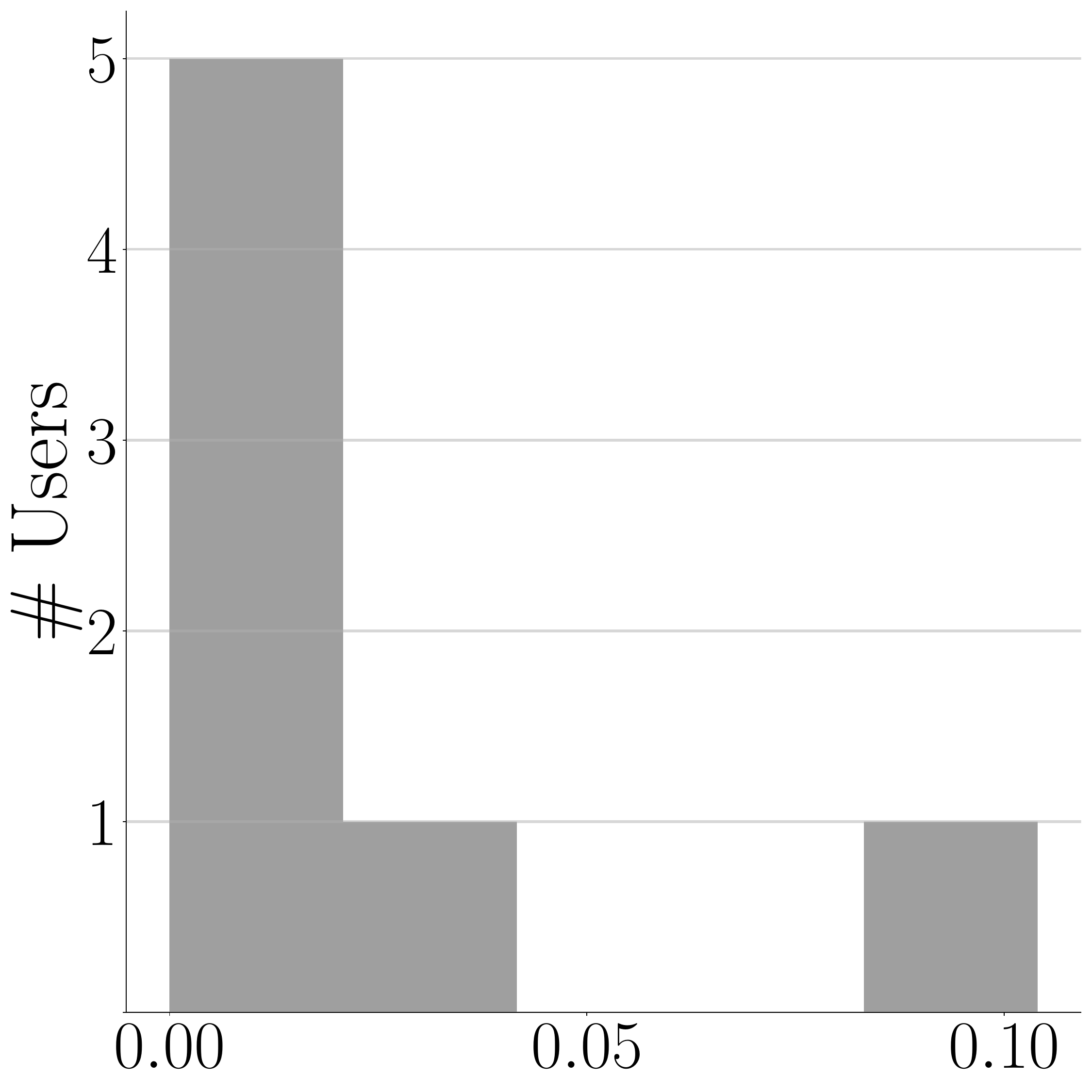} 
       & \includegraphics[width=0.31\linewidth,trim={0 0cm 0 0},clip]{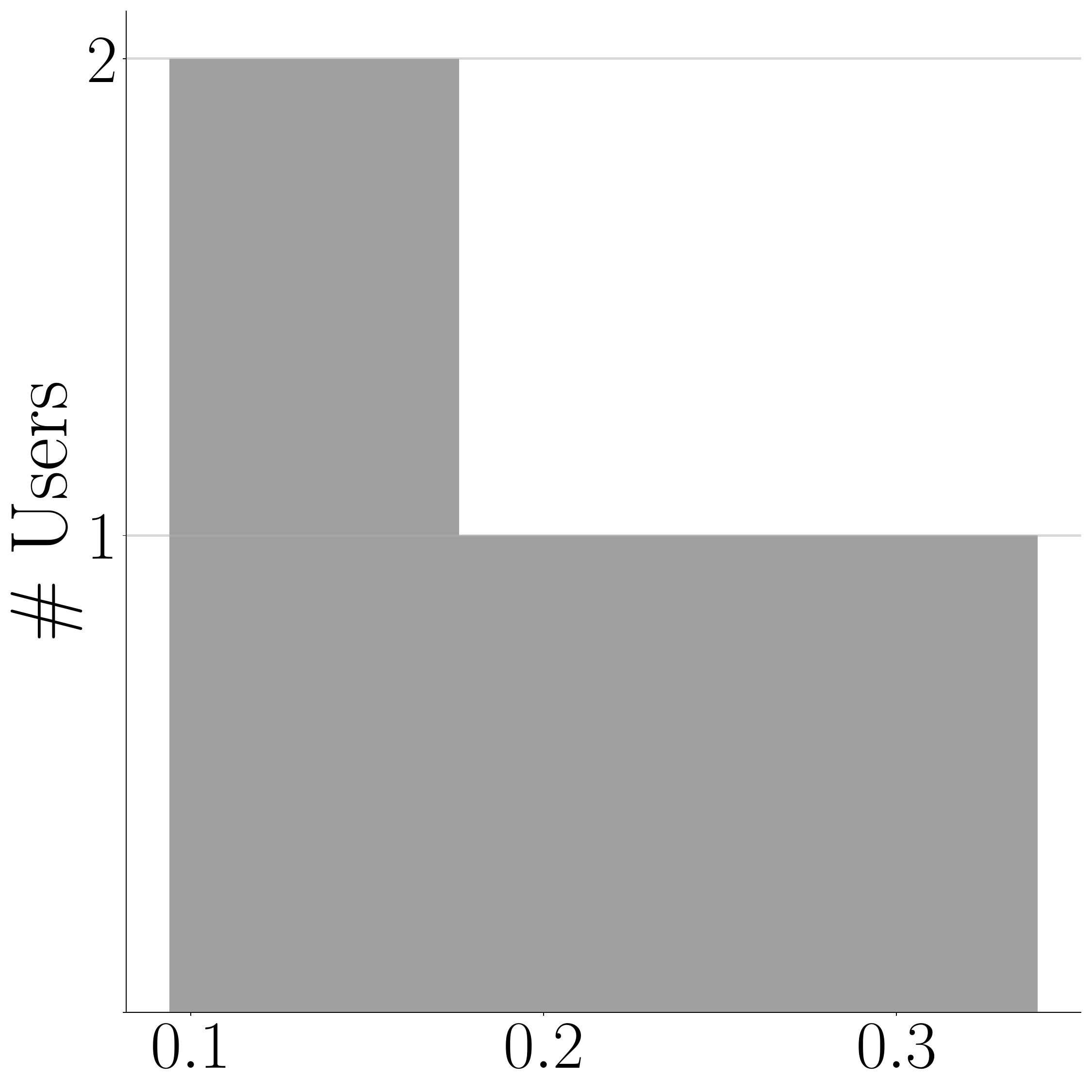}
       \\[-2mm] 
        \quad \quad $\lval[2,\variation]$
        & \quad \quad $\lval[2,\location]$
        &  \quad \quad $\lval[2,\engagement]$ \\ %
        (a) & (b) & (c) 
    \end{tabular}
    \caption{\tbf{Histogram of the fractions~$\lval[2, \var]$~\cref{eq:p2} for 39, 7, and 4 potentially interesting users of type 2 for $\var=\variation, \location$, and \engagement, respectively, each with $|\intscore[2,\var]-0.5|\geq 0.4$ in \cref{fig:numint2_obs_hist}.} The count on the vertical axis represents the number of users with the value of $\lval[2, \var]$ on the horizontal axis.
     \label{fig:int2_pvalues}}
\end{figure}

\begin{table}[ht]
\centering
\resizebox{\textwidth}{!}{
   \begin{tabular}{ccccccc}
   \toprule
      \tbf{\thead{Interestingness \\ type}} & \tbf{Question}& \tbf{\thead{Observed metric}}  & \tbf{Personalized?} & \tbf{Conclusions stable?}  \\
      \midrule
      Type 1& \cref{ques:numint} & $\numint[1]=18$ %
      & \xmark\  (\cref{fig:numint1_null_hist})  & \xmark\ (\cref{fig:int1_oneSided_heatmaps}(a)) \\
      
        & \thead{\cref{ques:userint} for user 1 \\ \cref{fig:interestingUser}(a)}  & $\intscore[1]=1.0$  &  \checkmark\ (\cref{fig:user_int1}(c))
        & -- \\

      \midrule
      Type 1$^{+}$ & \cref{ques:numint} & $\numint[1]^+=17$  %
      & \checkmark\ (\cref{fig:int1_oneSided_histograms}(b)) & \checkmark\ (\cref{fig:int1_oneSided_heatmaps}(b))  \\
      
      Type 1$^{-}$ & \cref{ques:numint} & $\numint[1]^-=1$ & \checkmark\ (\cref{fig:int1_oneSided_histograms}(c)) %
       & \checkmark\ (\cref{fig:int1_oneSided_heatmaps}(c)) \\

      \midrule
      \thead{Type 2: $\var =\variation$ } & \cref{ques:numint} & $\numint[2,\var]=39$ & \checkmark\ (\cref{fig:numint2_null_hist}(a)) & %
      \checkmark\ (\cref{heatmaps}(a)) \\
        & \thead{\cref{ques:userint} for user 2 \\ \cref{fig:interestingUser}(b)} & $\intscore[2,\var]=0.0$ & \checkmark\ (\cref{fig:user_int2_variation}(c)) %
          & --  \\

      \midrule
      \thead{Type 2: $\var=\location$ } & \cref{ques:numint} & $\numint[2,\var]=7$ & \checkmark\ (\cref{fig:numint2_null_hist}(b)) %
      & \checkmark\ (\cref{heatmaps}(b)) \\
        & \thead{\cref{ques:userint} for user 3 \\ \cref{fig:interestingUserLocation}(b)}  & $\intscore[2,\var]=0.0$ & \checkmark\ (\cref{fig:user_int2_location}(c))%
        & -- \\

      \midrule
      \thead{Type 2: $\var= \engagement$ } & \cref{ques:numint} & $\numint[2,\var]=4$  & \xmark\ (\cref{fig:numint2_null_hist}(c)) %
      & \checkmark\ (\cref{heatmaps}(c)) \\
        & \thead{\cref{ques:userint} for user 4 \\ \cref{fig:interestingUserEngagement}(c)} & $\intscore[2,\var]=0.037$ & \xmark\ (\cref{fig:user_int2_engagement}(c)) %
         & -- \\
   \bottomrule \hline
   \end{tabular}
   }
    \caption{\tbf{Summary of results from our resampling-based exploratory data analyses for HeartSteps data.} The first column denotes the type of interestingness considered (\cref{sub:personalization}); second column denotes the question of interest (\cref{ques:numint}, \cref{ques:userint}); third column states the observed metric (\numint[], \intscore[])  corresponding to the first two columns on the original data; fourth column answers whether our resampling based strategy provides evidence in the favor of personalization ($\checkmark$) or not (\xmark); and the final column states whether the personalization results are stable ($\checkmark$) or not (\xmark) to the choice of hyperparameters used in the interestingness scores (we did the stability analysis only for \cref{ques:numint}; see \cref{sec:unpack_int1,sec:additional_results_heartsteps}). Note that while our analysis for interestingness of type 1 for \cref{ques:numint} does not indicate evidence of personalization, we find evidence when looking at the one-sided variants (see \cref{sec:unpack_int1}). \label{tab:results}}
\end{table}

\section{Related work}
\label{sec:related_work}
Our work is an exploratory data analysis~\cite{tukey1977exploratory} for data collected by RL algorithms. We create visualizations to compare various statistics computed from our observed and resampled trajectories. Our chance distributions are computed by rerunning the learning algorithm and assigning rewards according to a true parametric model. We focus on secondary data analyses and hypothesis generation for further investigation, and thus our work is complimentary to the growing literature for primary data analysis with adaptively collected data~\cite{boruvka2018assessing,qian2021estimating,Zhang2020InferenceBandits, Bibaut2021Post-Contextual-BanditInference, Hadad2019ConfidenceExperiments}.

Overall our work builds on several areas of research on exploratory methods (a) for secondary analyses with adaptively collected data and (b) for diagnosing the performance of an RL algorithm. 
The classical works on EDA use resampling \cite{Buja1996InteractiveVisualization, Gelman2004ExploratoryModels,good2006resampling} to compare an observed statistic in the data to that from a replication distribution. This replication distribution can be generated via permutation testing, parametric bootstrap, or posterior predictive checking. While such approaches have been used successfully for exploratory as well as confirmatory assessment for many statistical problems, here we use a parametric model that is fitted on the observed adaptively collected data.

We note that resampling methods, like bootstrap, have been primarily used in RL for producing confidence intervals around value functions \cite{Hanna2017BootstrappingEvaluation, Ramprasad2021OnlineLearning,White2010IntervalDomains,Hao2021BootstrappingInference}, or to guide online learning policies \cite{Hao2019BootstrappingBound, Wang2020ResidualAlgorithms, Elmachtoub2017ATrees, Eckles2019BootstrapSciences}. To the best of our knowledge, RL literature does not apply resampling methods to perform exploratory analysis on assessing the personalization achieved by an RL algorithm as we do here.

Our resampling approach is also related to randomization inference~\cite{fisher1935design} and permutation tests that have been used to estimate treatment effects in causal inference~\cite{rosenbaum2002observational,ding2016randomization}. Our set-up here differs in a couple of ways: First, the treatments are adaptively assigned in our setting using an RL algorithm. Second, we use a parametric model to generate rewards under different treatments instead of the common choice of fixing the rewards to the observed values as in a ``strong null''.

Finally, our motivating questions \cref{ques:userint,ques:numint} are implicitly asking about the stability of the personalization achieved by the RL algorithm. Stability implies generalization is now folklore in supervised machine learning~\cite{vapnik1974theory,bousquet2002stability}. Our framework is based on a similar principle: If the RL algorithm is learning while being stable with respect to the underlying generative process, the advantage forecasts would likely not appear very interesting just by chance. Thus, the resampled trajectories are used to estimate the stability of the personalization of RL algorithm by rerunning it under different generative models. A mathematical framework connecting the two viewpoints is an exciting future direction.

\section{Discussion}
\label{sec:discussion}

In this work, we take a first step and introduce an exploratory methodology to assess personalization by a stochastic RL algorithm using resampling and promote data-driven truth-in-advertising about the success of RL methods. We operationalized personalization and the two questions based on the interesting trends that consistently appear in the RL. We applied our methodology with a case study on data from a mobile health trial to collect evidence for the conjectures made by their design team and generate hypotheses that can inform further data collection and probing into feature design for the RL algorithm. 
\subsection{Limitations and extensions for HeartSteps}\label{limitations}
We now acknowledge several limitations that should be taken into account while interpreting our results. First, there were a few subtle differences between the real algorithm implementation and the RL algorithm used in our resampling. Our resampling procedure (namely \cref{resimHeartSteps}) uses the RL algorithm \emph{as-intended} by the HeartSteps design team.  During the actual clinical trial, there were instances when the latest user sensor data and features did not arrive in time for the posterior update. In such cases, the data was imputed~\cite{liao2020personalized}, and the posteriors were updated using the imputed data. Later, when the real data arrived, the trial database was updated. We utilized the (up-to-date) real data recorded in the database. In addition, during the trial, transformations to the randomization probabilities~\cref{eq:h_clip} changed partway through for a small number of users. We use~\cref{eq:h_clip} in order to stay true to the intended RL algorithm. During the resampling, we followed the original implementation, dropping data points where state or reward was not observed. Technically this may call for a refined analysis as these data points were likely not missing at random.

We reiterate that there are multiple ways to mathematize a certain kind of personalization. For example, the interestingness score for type $1$ can be altered to measure the consistency of standard advantage in being positive rather than the frequency of it being positive.  Moreover, one might consider other types of interestingness. For instance, our interestingness of type $2$ for feature $\var$ can be modified to consider interactions of features.

Finally, our resampling procedure assumes that the (stationary) reward model~\cref{eq:working_model} is correctly specified. It would be interesting to assess the limitations of our methodology to model misspecification. 
The resampling procedure treats all features as fixed except for \dosage. However, all features except \temperature are partly endogenous. Consequently, one might consider a few different variants for \resimtraj. One might consider fitting and using a non-stationary reward model, a state transition model, as well as a noise distribution model. These models can then (with appropriate modifications) be used in the more general \resimtraj formulation in \cref{resim} depending on the choice of interestingness. However, a reliable fit for such models would require a large user trajectory with $T\to\infty$, a real challenge in several mHealth trials like HeartSteps.

\subsection{Future work}
\label{future}
Our work opens several avenues of future research. Here we focus solely on the stochasticity due to the RL algorithm and treat the users as fixed. However, our methodology can be extended to take into account several other  sources of stochasticity. For instance, when the users are being drawn from a super-population, to answer population-level analogs of \cref{ques:numint}, one can use statistical bootstrap~\cite{Efron1994AnBootstrap} to  account for stochasticity due to user sampling. Moreover, we can also statistical bootstrap to take into account within-user variability in the reward distributions. As noted above, our resampling strategy assumes that the working reward model by the RL algorithm is correctly specified. Addressing the limitations of such an assumption is also of broad interest.

Here we focused on stochastic RL algorithms due to our motivations from their popularity in mHealth applications. In other settings, non-stochastic methods like UCB and its variants might be more suitable. A future direction is to extend our methodology to assess the personalization achieved by such algorithms potentially by utilizing sampling variability across users, and the stochasticity inherent to the user in their state transitions and the noise in the observed reward.   

There is a growing interest in using RL algorithms that pool information across users to speed up learning. Developing a methodology to assess whether such pooling indeed helps speed up learning and personalization is a natural future step~\cite{dwivedi2022counterfactual, zhang2023statistical, tomkins2019intelligent}.
The mobile app in HeartSteps had multiple intervention components, here we focused on one and we treated another component (messages for anti-sedentary behaviors) as exogenous and fixed. Developing methods that take into account these multiple components together is yet another interesting direction.  

Finally, while our methodology is exploratory in nature, developing confirmatory tests and establishing their consistency under a theoretical framework is also of immense interest. In particular, our methodology uses a parametric model fitted on the entire user trajectory to resample trajectories to maintain similarities between observed and resampled trajectories. Theoretical analysis can shed light on the advantages and limitations of such an approach. 

\section*{Declarations}
\paragraph{Funding}
RK is supported by NIGMS Biostatistics Training Grant Program under Grant No. T32GM135117. SG and SM acknowledge support by NIH/NIDA P50DA054039, and NIH/NIBIB and OD P41EB028242. RD acknowledges support by NSF DMS-2022448, and DSO National Laboratories  grant DSO-CO21070. RD and SM also acknowledge support by NSF under Grant No. DMS-2023528 for the Foundations of Data Science Institute (FODSI). PK acknowledges support by NIH NHLBI R01HL125440 and 1U01CA229445. SM also acknowledges support by NIH/NCI U01CA229437, and NIH/NIDCR UH3DE028723. KWZ is supported by the Siebel Foundation and by NSF CBET–2112085 and by the NSF Graduate Research Fellowship Program under Grant No. DGE1745303.

\paragraph{Conflict of Interest/Competing Interests}
KWZ worked as a summer intern at Apple. The other authors declare no conflict of interest. %

\paragraph{Ethics Approval}
We adhere to the policies outlined in \href{https://www.springer.com/gp/editorial-policies/ethical-responsibilities-of-authors}{https://www.springer.com/gp/editorial-policies/ethical-responsibilities-of-authors}.

\paragraph{Consent to Participate}
The reported study was approved by the Kaiser Permanente Washington Institutional Review Board (IRB 1257484-16). All participants completed a written informed consent to take part in the study.

\paragraph{Consent for Publication}
Adhering to the IRB and consent forms, individual level data that could be used for identification are not released.

\paragraph{Availability of Data and Materials}
Under the current data policies for HeartSteps V2/V3, the research team cannot make the data publicly available.

\paragraph{Code availability} 
The code used for generating resampled trajectories and reproducing the plots is available at the \href{https://github.com/changon/resampling_heartsteps}{following link}.

\paragraph{Authors contributions}

All authors contributed to conceptualization and methodology. Data Preparation and Software: PC, PL, RK, and SG. Analysis: RK. Writing: RK, RD, SM, and SG. Funding and Administration: PK, SM. Supervision: RD, KZ, SM. All authors read and approved the final manuscript.

\bibliography{refs}
\appendix

\section{Details on the RL framework for HeartSteps}
\label{sub:defer_rl}
We now provide further details regarding the RL framework for the HeartSteps study discussed in \cref{subsec:rlalg}. 

\begin{table}[ht]
\centering
   \begin{tabular}{cp{0.45\textwidth}cc}
   \toprule
      \tbf{Variable}& \tbf{Description}  & \tbf{Range} & \tbf{Part of f(s)}\\
      \midrule
      \dosage & Function of walking suggestions and anti-sedentary messages pushed to the user, defined in \cref{eq:dosage} in \cref{subsec:sar} & $[0, 20]$ & \checkmark \\
      \engagement & Binary indicator. The indicator is $1$ if the number of screens encountered by the user in the app from the prior day is greater than 40\% quantile of the screens collected over the last seven days (after period one) or over the previous days (during period one); otherwise the indicator is $0$. & $\sbraces{0, 1}$ & \checkmark \\ 
      temperature & Temperature (in Celcius) at the current location of the user &  $[-15.6, 36.1]$ &\\
      \location & Binary indicator of whether the user is at home or work location (0), or not (1). & $\sbraces{0, 1}$ & \checkmark \\
      \variation & Binary indicator variable - for a study day $d+1$ and $k^{th}$ timeslot, the indicator is 1 if the standard deviation (past 7-days data) of 60-min step count of a user calculated at day $d$ for time slot $k$ is greater than or equal to median of the past standard deviations up to day $d$ for $k^{th}$ time slot, and 0 otherwise.  & $\sbraces{0, 1}$ & \checkmark \\
      prior 30-min step count & Log-transformed step count of user 30 mins prior to the current decision time & $[-0.69, 8.6]$ &\\
      yesterday step count & The square root of the step-count from the tracker, collected from 12 AM to 11:59 PM & $[0, 209]$ &\\
   \bottomrule \hline
   \end{tabular}
   \caption{Description of state features used in the HeartSteps study. All the feature values are standardized before being used in the regression models. Please refer to \citet[Table 3]{liao2020personalized} for an even more detailed description and information about the state features. }
   \label{tab:state_features}
\end{table}

\paragraph{Clipping probabilities}
The function $h$ appearing in \cref{eq:probability} is given by
\begin{align}
\label{eq:h_clip}
    h(p) \defeq \mathrm{min}\{ 0.8, 0.2+ \frac{0.8}{0.5} \cdot \mathrm{max}\{ p-0.5,0 \} \}
    \qtext{for} p \in [0, 1].
\end{align}

\paragraph{Prior and posterior formulation}
Using the notation $\theta\tp \defeq (\alpha_0\tp, \alpha_1\tp, \beta\tp)$, the prior for $\theta$ was specified as $\mc N(\mu_0, \Sigma_0)$,  where
\begin{align}
\label{eq:prior}
    \overline{\mu}_0 = \begin{bmatrix} \mu_{\alpha_0} \\ \mu_{\beta} \\ \mu_{\beta}\end{bmatrix} 
    \qtext{and}
    \overline{\Sigma}_0 = \begin{bmatrix} \Sigma_{\alpha_0} && \\ & \Sigma_{\beta} &\\ & &\Sigma_{\beta} \end{bmatrix} 
\end{align}
and $\sbraces{\mu_{\alpha_0}, \mu_\beta,  \Sigma_{\alpha_0},  \Sigma_{\beta}}$ were computed from the prior study (see~\cite[Sec.~6]{liao2020personalized} for details on how priors were constructed and \cref{eq:prior_sigma} for the specific values used by us; note that the HeartSteps team decided to update the priors used from those those presented in \cite{liao2020personalized}).
Given the Gaussian prior and the Gaussian working model, the posterior for $\theta$ on the day $d$ is also Gaussian and is given by $\mc N(\overline{\mu}_d, \overline{\Sigma}_d)$, where these posterior parameters are recursively updated as 

\begin{subequations}
\label{eq:updates}
\begin{align}
\label{eq:sigma_d}
   \overline{\Sigma}_{d} &= \parenth{\frac{1}{\sigma^2} \sum_{t=5(d-1)}^{5d} I_t \phi(S_t, A_t) \phi(S_t,A_t)^\top + \overline{\Sigma}_{d-1}^{-1}}\inv, \qtext{and}
   \\ 
\label{eq:mu_d}
   \overline{\mu}_{d}  &=  \overline{\Sigma}_d\parenth{\frac{1}{\sigma^2} \sum_{t=5(d-1)}^{5d} I_t \phi(S_t, A_t) R_t + \overline{\Sigma}_{d-1}^{-1} \overline{\mu}_{d-1}}
\end{align}
\end{subequations}
where $\phi(S_t, A_t)\tp \defeq [g(S_t)\tp, \pi_t f(S_t)\tp, (A_t-\pi_t)f(S_t)\tp]$ collects all the feature vectors from the working model~\cref{eq:working_model}. The updates~\cref{eq:sigma_d,eq:mu_d} are denoted by \texttt{PosteriorUpdate} in \cref{resimHeartSteps}. For $k$-dimensional $\beta$, the posterior parameters $\mu_{d, \beta}, \Sigma_{d, \beta}$ for $\beta$ are respectively given by the last $k$ entries of $\overline{\mu}_d$ and $k\times k$ sub-matrix formed by taking the last $k$ columns and rows of $\overline{\Sigma}_d$.

\paragraph{Model estimates used by \resimtraj for resampling trajectories}
For a user trajectory $(S_{t}, A_{t}, R_{t})_{t= 1}^{T}$ from the reward model~\cref{eq:working_model}, we estimate the parameters $(\alpha, \beta)$ using the updates~\cref{eq:updates} albeit without action centering. Hence the estimates $(\hat \alpha_T \tp , \hat \beta_T\tp)$ (that inform the model parameters used by \resimtraj after suitable modifications in \cref{subsec:parasim_heartsteps}) are given by
\begin{align}
\label{sub:model_fit}
    \begin{bmatrix}
        \hat \alpha_{T} \\ 
        \hat \beta_{T}
    \end{bmatrix} \!=\! \left(\frac{1}{\sigma^2} \sum_{t=1}^{T} I_t \til \phi(S_t, A_t) \til\phi(S_t,A_t)^\top + \begin{bmatrix} \Sigma_{\alpha_0} && \\ & \Sigma_{\beta} & \end{bmatrix} \inv\right)\inv \left(\frac{1}{\sigma^2} \sum_{t=1 }^{T} I_t \til\phi(S_t, A_t) R_t \!+\! \begin{bmatrix} \Sigma_{\alpha_0} && \\ & \Sigma_{\beta} & \end{bmatrix} \inv \begin{bmatrix} \mu_{\alpha_0} \\ \mu_{\beta} \end{bmatrix} \right)
\end{align}
where $\til{\phi}(S_t, A_t)\tp \defeq \left[g(S_t)\tp, A_t f(S_t)\tp\right]$.

\paragraph{Priors Means and Variances}
We now summarize the exact values of prior parameters used by the RL algorithm.
For $\l \in \N$, let $\diag(a_1, \ldots, a_{\l}) \in \real^{\l \times \l}$ denote an $\l\times\l$ diagonal matrix with its $j$-th diagonal entry equal to $a_j$ for $j=1, \ldots, \l$. Then the mean and variance parameters used in \cref{eq:prior} are given by
\begin{align}
\label{eq:prior_sigma}
\begin{split}
    \mu_{\alpha_0} &= [0.82, 1.95, 3.81, -0.19, 0.76, 0.0, -0.92, 0.0]\tp \in \real^{7}, \\
    \mu_{\beta} &= [0.47, 0.0, 0.0, 0.0, 0.0]\tp \in \real^{5}, \\ 
    \Sigma_{\alpha_0} &= \diag(14.24, 13.35, 3.24, 0.57, 19.00, 0.26, 17.00, 7.35) \in \real^{7\times 7}, \qtext{and} \\
    \Sigma_{\beta} &= \diag(4.93, 24.56, 4.95, 0.67, 0.82)  \in \real^{5\times 5},
    \end{split}
\end{align}
where the features of $\alpha_0$ are ordered as
(Intercept, temperature, prior 30 min step count, yesterday step count, \dosage, \engagement, \location, \variation) and that for $\beta$ and $\alpha_1$ are ordered as (Intercept, \dosage, \engagement, \location, \variation).
\section{Details on \intscore computation for HeartSteps}
\label{sec:int_details}

We now describe the smoothened  version of $\intscore[1]$~\cref{eq:intscore1} and $\intscore[2,\var]$~\cref{eq:intscore2} that we use to add stability in our HeartSteps results in \cref{subsec:hs_int1,subsec:hs_int2}. 

At a high level, we use the following steps: (i) We use moving windows to average out the advantage forecasts and use an \emph{averaged} forecast on a daily scale, i.e., for $d=\ceil{(t-1)/5}$ and not for each decision time $t\in [T]$ as in \cref{eq:intscore1,eq:intscore2}. (iii) While computing the interestingness scores, we omit days when the quality of data is not good due to low availability or low diversity of features. (iii) We compute the forecasts without changing any state feature from the observed data other than \dosage. That is, if the observed feature value $\var_t=1$, we do not compute a counterfactual forecast by artificially forcing $\var_t = 0$.  (iv) Finally, we do not consider a user's interestingness score if they have only a few days of good data.

\begin{figure}[ht!]
\centering
 \resizebox{\textwidth}{!}{
\begin{tabular}{cccc}
     \includegraphics[width=0.245\textwidth]{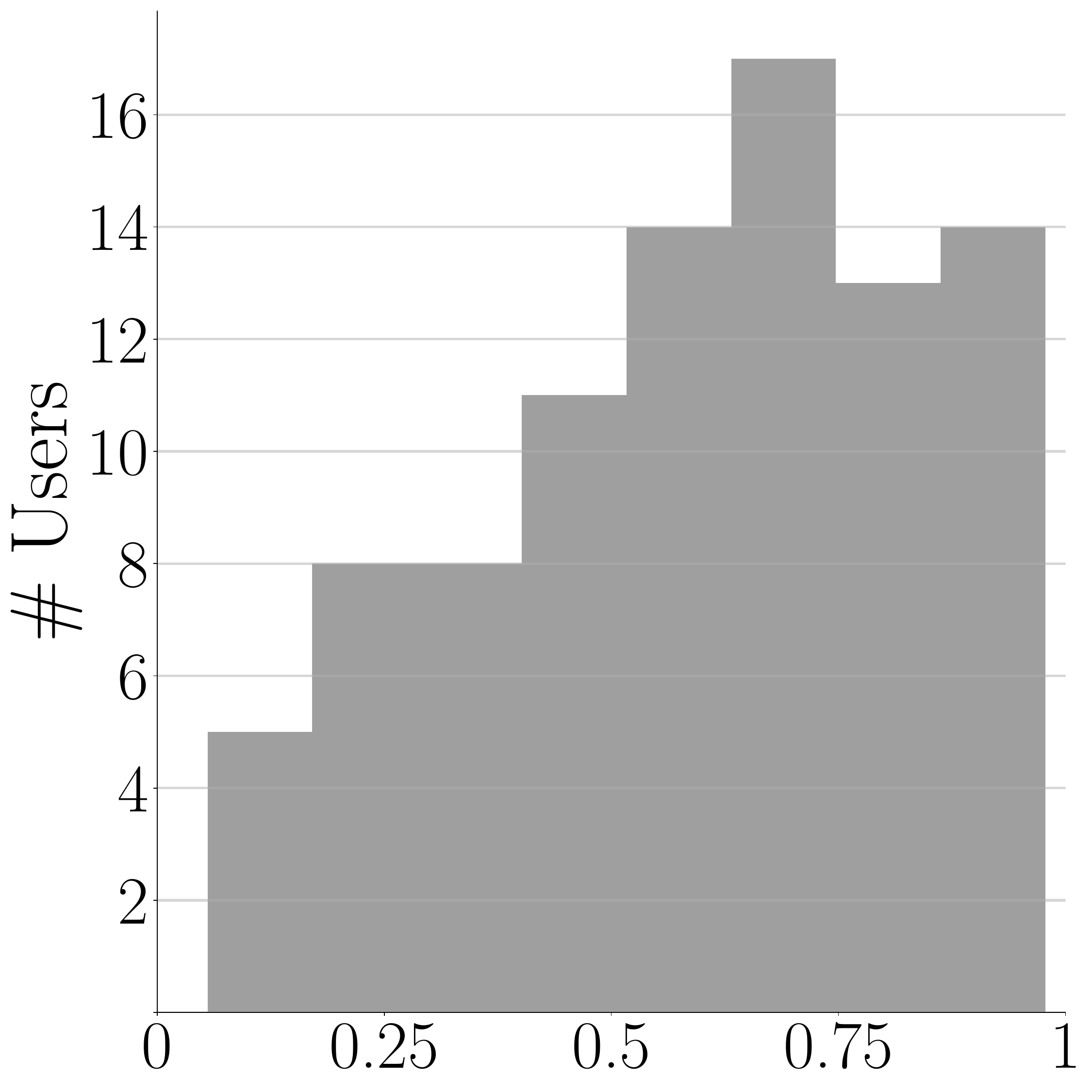}
     & 
     \includegraphics[width=0.245\textwidth]{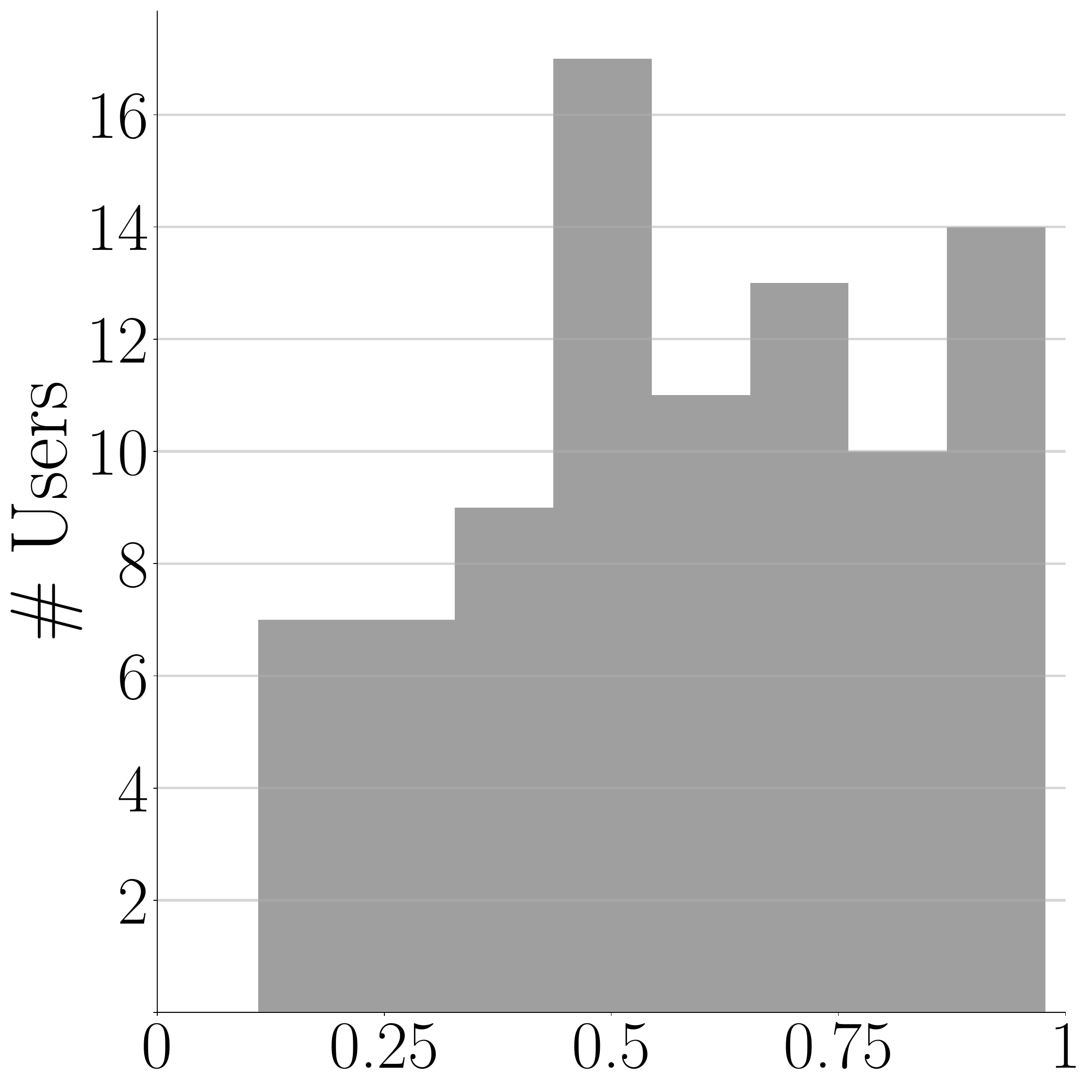} 
     &
     \includegraphics[width=0.245\textwidth]{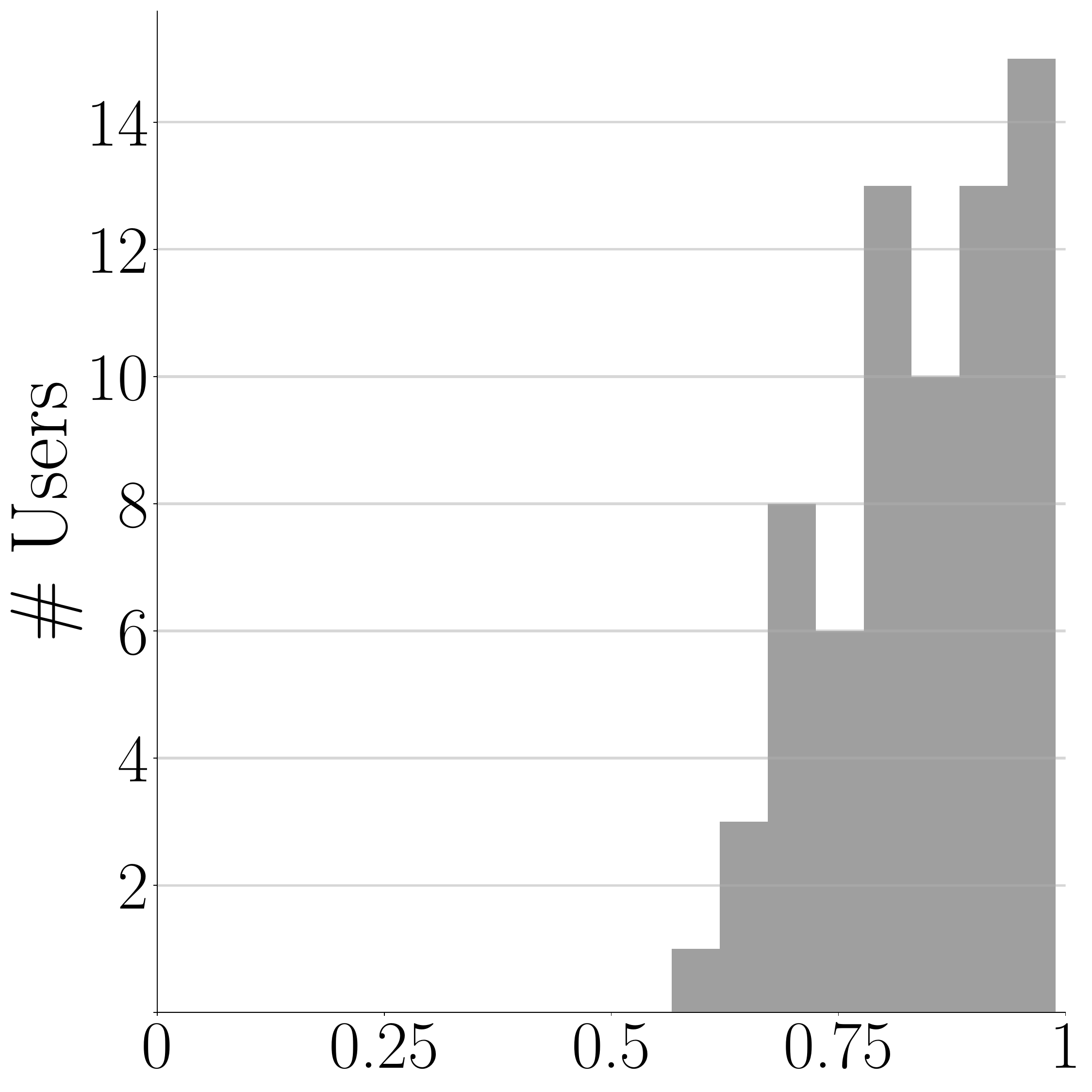}
     &
     \includegraphics[width=0.245\textwidth]{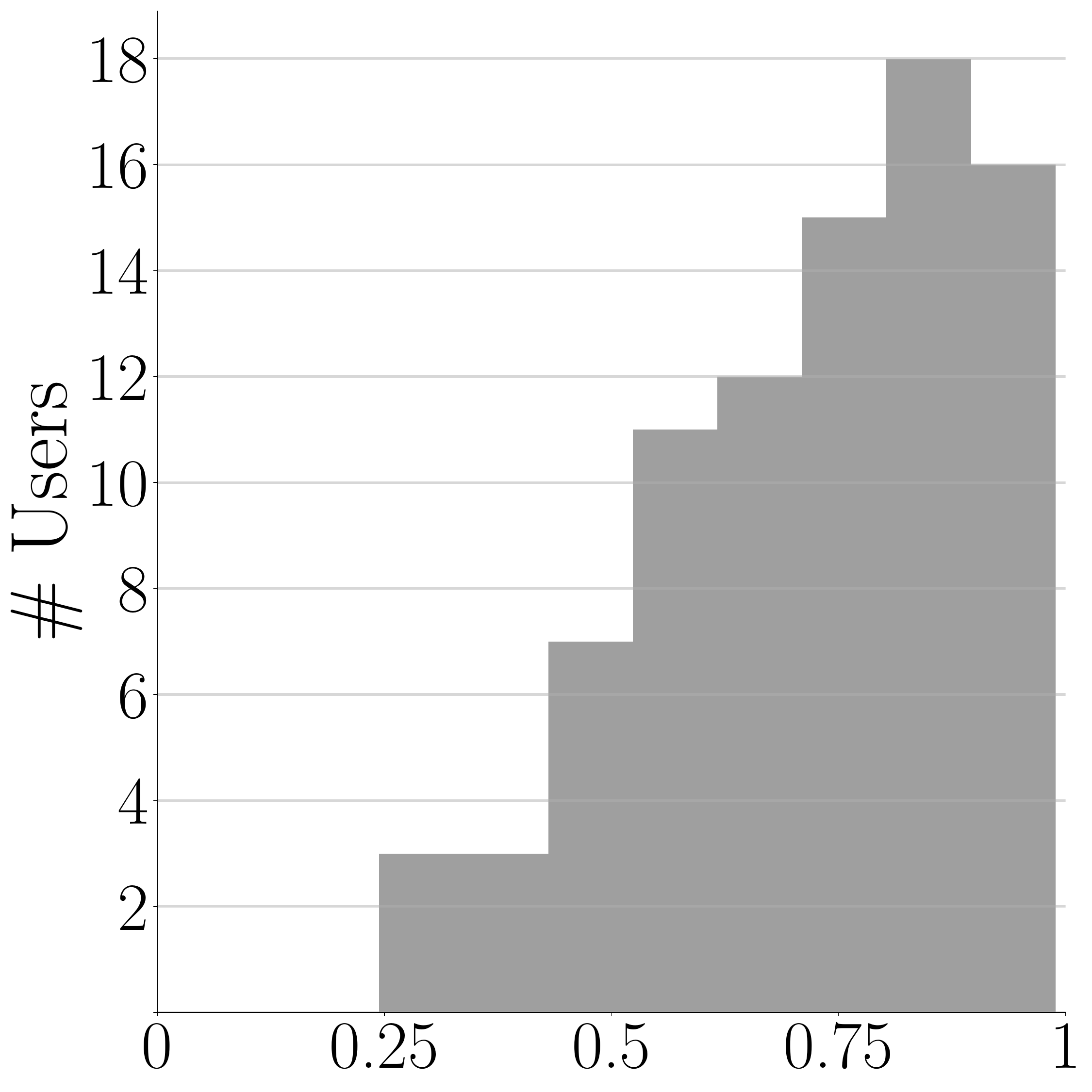}
     \\
     (a) Int. Type $1$ & (b) \variation  & (c) \location  & (d) \engagement
\end{tabular}
}
\caption{\tbf{Histogram of fraction of good days for the users in the original data for HeartSteps.} In panel~(a), the count on the vertical axis represents the number of users with the value of  $\frac1D \sum_{d=1}^{D} G_{d,1 }$ on the horizontal axis. In panels (b) to (d), the count on the vertical axis represents the number of users with the value of  $\frac1D \sum_{d=1}^{D} G_{d,2,\var}$ on the horizontal axis, respectively for $\var\in\sbraces{\variation, \location, \engagement}$. \label{gamma_hist}}
\end{figure}

We now describe these steps in detail for a user with total $T$ decision times in their data trajectory (with total $D\defeq\floor{T/5}$ days of data). We note that total decision times for each user might vary.
\begin{enumerate}[leftmargin=*]
    \item \tbf{Sliding window:} For each day $d \in \sbraces{1, \ldots, D}$, we define a sliding window $W_d$ using all $5$ decision times on day $d$ when computing \intscore[1] and all 5 decision times on days $\sbraces{d-1, d, d+1}$ (total $15$ decision times) when computing \intscore[2, \var]. That is,
    \begin{align}
        W_d \defeq \begin{cases}
        \sbraces{5(d-1), 5(d-1)+1, \ldots, 5d} \cap [T] &\qtext{for} \intscore[1], \\
        \sbraces{5(d-1)+1, 5(d-1)+2, \ldots, 5(d+1)} \cap [T] &\qtext{for} \intscore[2, \var].
        \end{cases}
    \end{align}
    \item \tbf{Characterizing good data day:} 
    Next, when considering \intscore[1], we define an indicator variable $G_{d, 1}$ to denote a \emph{good day}. It is set to $1$ if the following two conditions hold: (a) if the user was available for at least $2$ decision times in $W_d$, i.e., $\sum_{t\in W_d} \avail_t \geq 2$  and (b) the RL algorithm posterior was updated on the night of  day $d-1$; we impose this additional constraint to deal with the real-time and missing data update issues. We set $G_{d, 1}=0$ in all other cases.
    
    When considering \intscore[2, \var], we define the variable $G_{d, 2, \var}$ ta denote a good data day based on whether the user's observed states exhibit enough diversity in the value of the variable $\var$ for the decision times in $W_d$. 
    In particular, we set $G_{d, \var}=1$ when the following two conditions hold: (a) the feature $\var$ takes values $1$ and $0$ at least twice out of the decisions times in $W_d$ when the user was available for randomization ($\avail_t=1$), i.e., 
    \begin{align}
        \sum_{t\in W_d} \avail_t \var_t
        \geq 2
        \qtext{and}
        \sum_{t\in W_d}  \avail_t (1-\var_t)
        \geq 2,
    \end{align}
    where $\var_t$ denotes the value of the variable $\var$ for the user at decision time $t$,
    and (b) the RL algorithm posterior was updated on the night of at least one of the days in $\sbraces{d-1, d, d+1}$. In all other cases, we set $G_{d,\var}=0$.
    \item \tbf{Interestingness score for a user trajectory:}
    We consider a user for interestingness only if the fraction of good days is greater than a certain threshold, i.e., 
    \begin{align}
    \label{eq:good_days}
    \frac1D \sum_{d=1}^{D} G_{d,1 } \geq (1-\gamma) \stext{for} \intscore[1] \qtext{or} \frac1D \sum_{d=1}^{D} G_{d, 2, \var} \geq (1-\gamma) \stext{for} \intscore[2, \var],
    \end{align}
    for a suitable $\gamma \in (0, 1)$. (Note that increasing the value of $\gamma$ lowers the cutoff for a user to become eligible for being considered for interestingness.)
    For such a user, we define the interestingness scores as follows:
    \begin{align}
        \intscore[1](\mc U) & \defeq  \frac{1}{\sum_{d=1}^{D} G_{d, 1}} \sum_{d=1}^{D} G_{d, 1} \indicator\parenth{\frac{\sum_{t\in W_d} \avail_t \hat \Delta_t(S_t) }{\sum_{t\in W_d}\avail_t} > 0} \\ 
        \intscore[2, \var](\mc U) & \defeq  \frac{1}{\sum_{d=1}^{D} G_{d, 2, \var}} \sum_{d=1}^{D} G_{d, 2, \var} \indicator\parenth{\frac{\sum_{t\in W_d} \avail_t \var_t \hat \Delta_t(S_t) }{\sum_{t\in W_d}\avail_t \var_t}
        > 
        \frac{\sum_{t\in W_d}\avail_t (1-\var_t)  \hat \Delta_t(S_t) }{\sum_{t\in W_d}\avail_t (1-\var_t)}},
    \end{align}
    where we multiply by indicators $G_{d, 1}$ and $G_{d, 2, \var}$ to include only ``good days'' in our score computations. Note that $\frac{\sum_{t\in W_d} \avail_t \var_t \hat \Delta_t(S_t) }{\sum_{t\in W_d}\avail_t \var_t}$ is the stable proxy (without counterfactual imputation) for the quantity $\hat{\Delta}_{t}(S_t(\var=1))$ in \cref{eq:intscore2}. \\
\end{enumerate}

\begin{remark}
\label{rem:gamma}
Note that we omit all users from our results who do not satisfy the good day requirement \cref{eq:good_days}. The number of omitted users depends on the value of $\gamma$; see \cref{gamma_hist} for the histogram of $\sum_{d}G_{d, 1}/D$ and $\sum_{d}G_{d, 2, \var}/D$ across the 91 users in HeartSteps. For the results in \cref{subsec:hs_int1,subsec:hs_int2}, we use $\gamma=0.4$, that allows $63$, $60$, $12$, and $43$ to be considered respectively for interestingness of type $1$, and of type $2$ for feature \variation, \location, and \engagement.
\end{remark}

\section{Another look at user 2's advantage forecasts}
\label{sec:user_2_details}
\cref{fig:interestingUser_revisit}(a) reproduces the advantage forecasts for user 2 from \cref{fig:interestingUser}(b). In addition, panels (b) and (c) of \cref{fig:interestingUser_revisit} show the analogs of the panel (a), where user 2's standardized advantage forecasts are color-coded based on the values of \location and \engagement respectively. Overall, we observe from the three panels in \cref{fig:interestingUser_revisit} that user 2 does not appear interesting of type 2 for $\var \in \{ \location, \engagement \}$ since the standardized advantage forecasts are not well separated when $\var = 0$ versus $\var = 1$ like that for $\var = \variation$ in panel (a) (or equivalently in \cref{fig:interestingUser}(b)). 
In particular, for this user, we have $\intscore[2, \location] = 0.38$, and $\intscore[2, \engagement] =0.38$, while $\intscore[2,\variation]=0$.  %

\begin{figure}[ht!]
\centering
\begin{tabular}{ccc}
    \includegraphics[width=0.31\textwidth]{figs/updatedBlueRed_Serif/blueRedOverAll_user-4-state-variation.pdf} 
     & \includegraphics[width=0.31\textwidth]{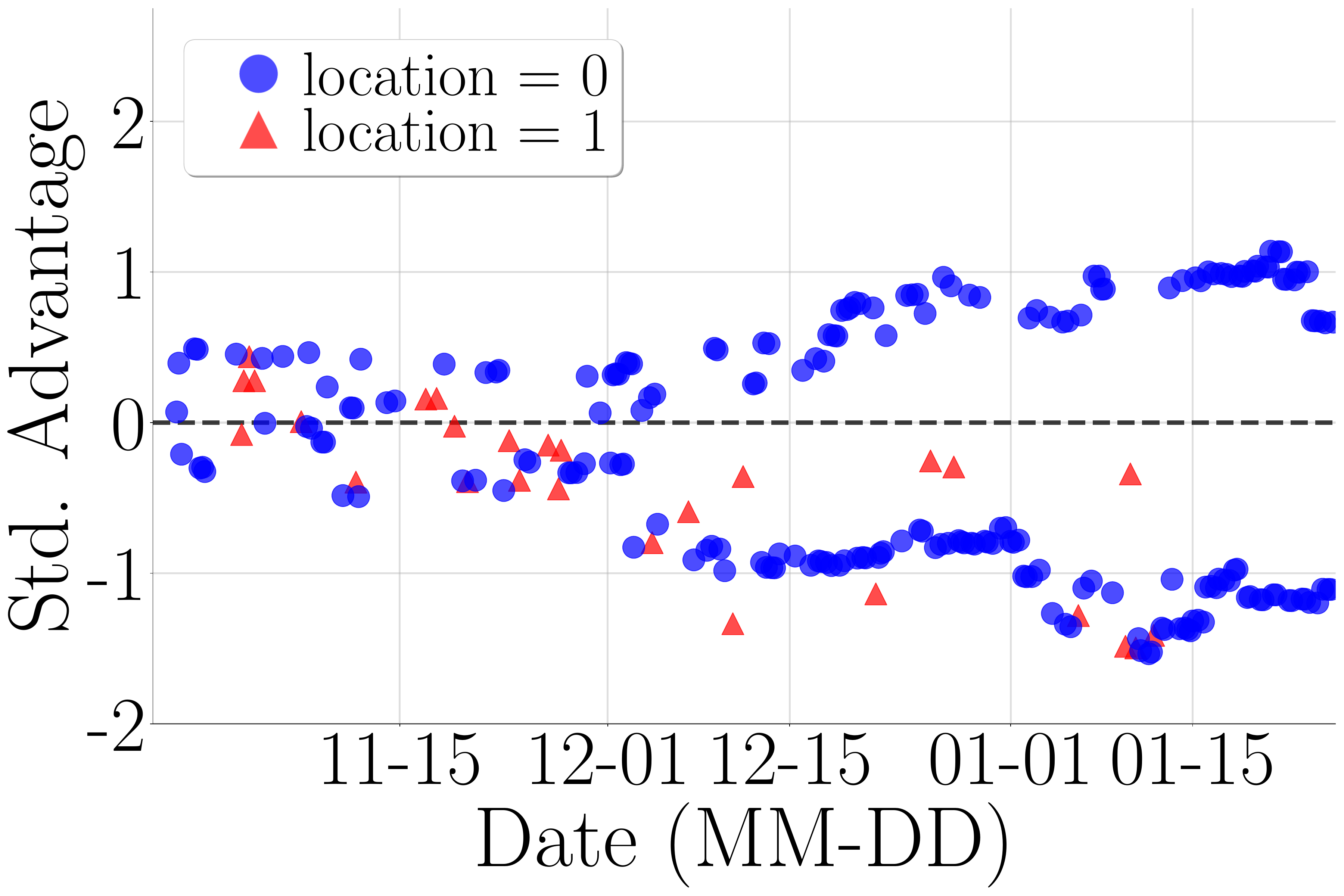} 
     & \includegraphics[width=0.31\textwidth]{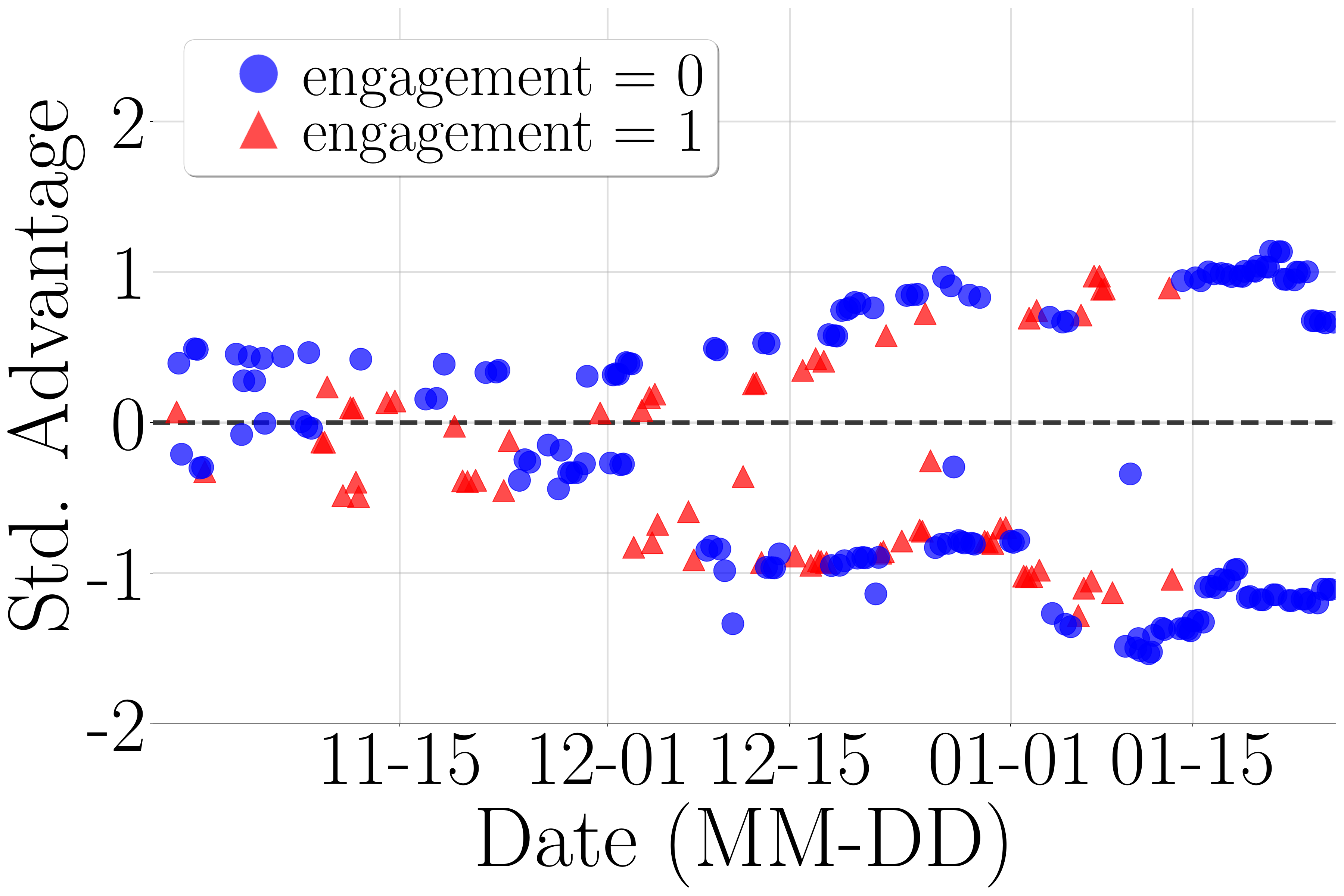}  \\
     (a) & (b) & (c)
\end{tabular}
\caption{\tbf{User 2's standardized advantage forecasts from \cref{fig:interestingUser}(b), color-coded by values of \var = \variation, \location, and \engagement in panels (a), (b), and (c) respectively.}  The value on the vertical axis represents the RL algorithm's forecast of the standardized advantage of sending an activity message for the user if the user was available for sending a message on the day marked on the horizontal axis. (Note each day has $5$ decision times.) The forecasts are marked as blue circles based on \var = 1 and red triangles if \var = 0 at the decision time. Panels (a) to (c) exhibit, respectively, $\intscore[2, \variation]=0$, $\intscore[2, \location]=0.38$, and $\intscore[2, \engagement]=0.38$. Note that we reproduced \cref{fig:interestingUser}(b) in panel (a) for the reader's convenience, and the three panels plot the same data and differ only in the color coding. 
\label{fig:interestingUser_revisit}}
 \end{figure}

\section{Deeper dive into interestingness of type $1$ for HeartSteps}
\label{sec:unpack_int1}

\label{sec:one-sided-explanation}
To further refine the conclusions from \cref{fig:numint1_null_hist}, we consider one-sided variants of the definition~\cref{eq:numint1} for the number of interesting users. For reader's convenience, we reproduce  \cref{fig:numint1_null_hist} in panel (a) \cref{fig:int1_oneSided_histograms}, besides the corresponding results with one-sided interesting user counts, defined by counting the users with $\intscore[1] \geq 0.9$ and $\intscore[1] \leq 0.1$ separately in panels (b) and (c).

\begin{figure}[ht!]
\centering
\resizebox{\textwidth}{!}{
\begin{tabular}{ccc}
    \includegraphics[width=0.33\textwidth]{eta_hm/histogram_Interesting_Zero_delta1=0.75_delta2=0.4_B=500.pdf} 
    & \includegraphics[width=0.33\textwidth]{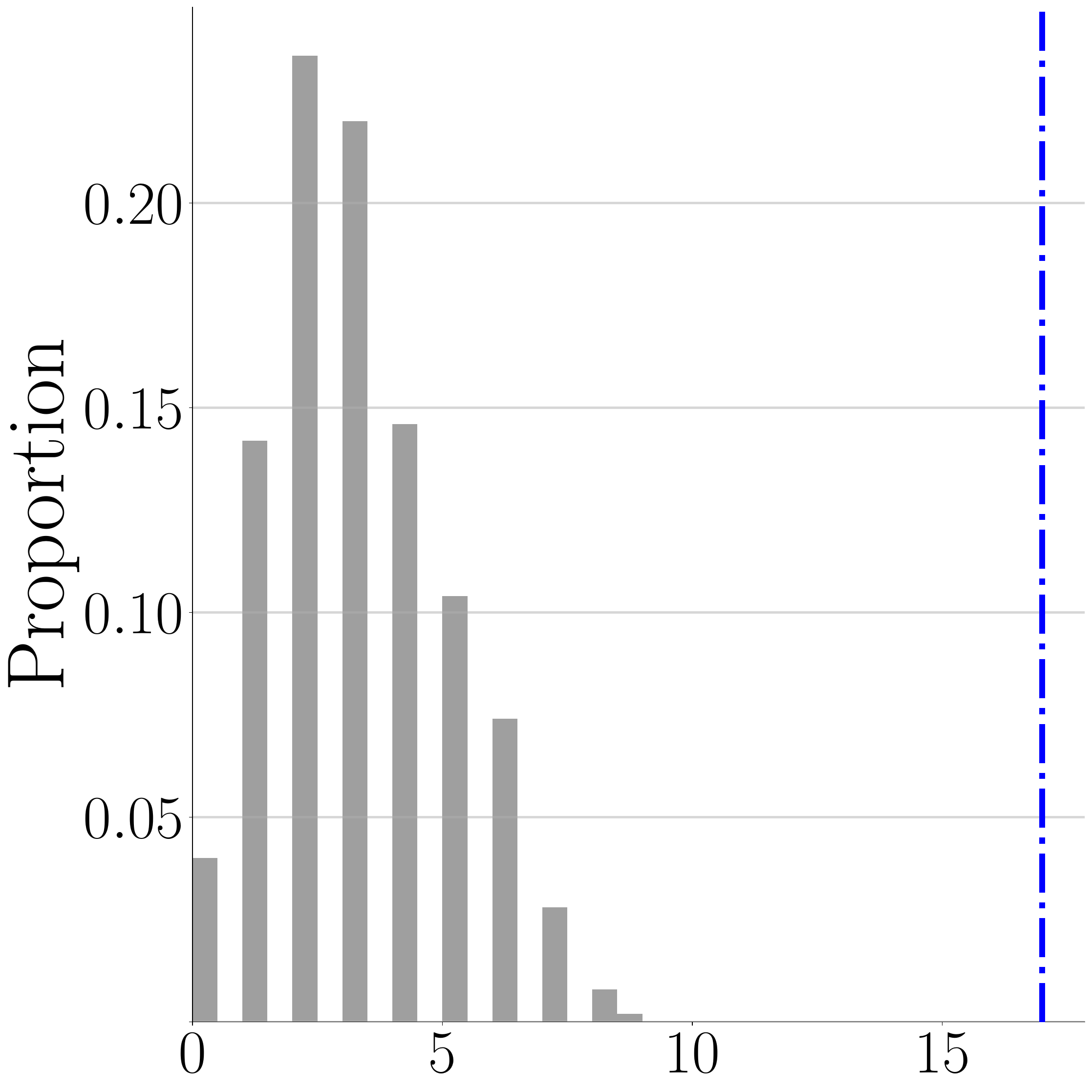} 
     & \includegraphics[width=0.33\textwidth]{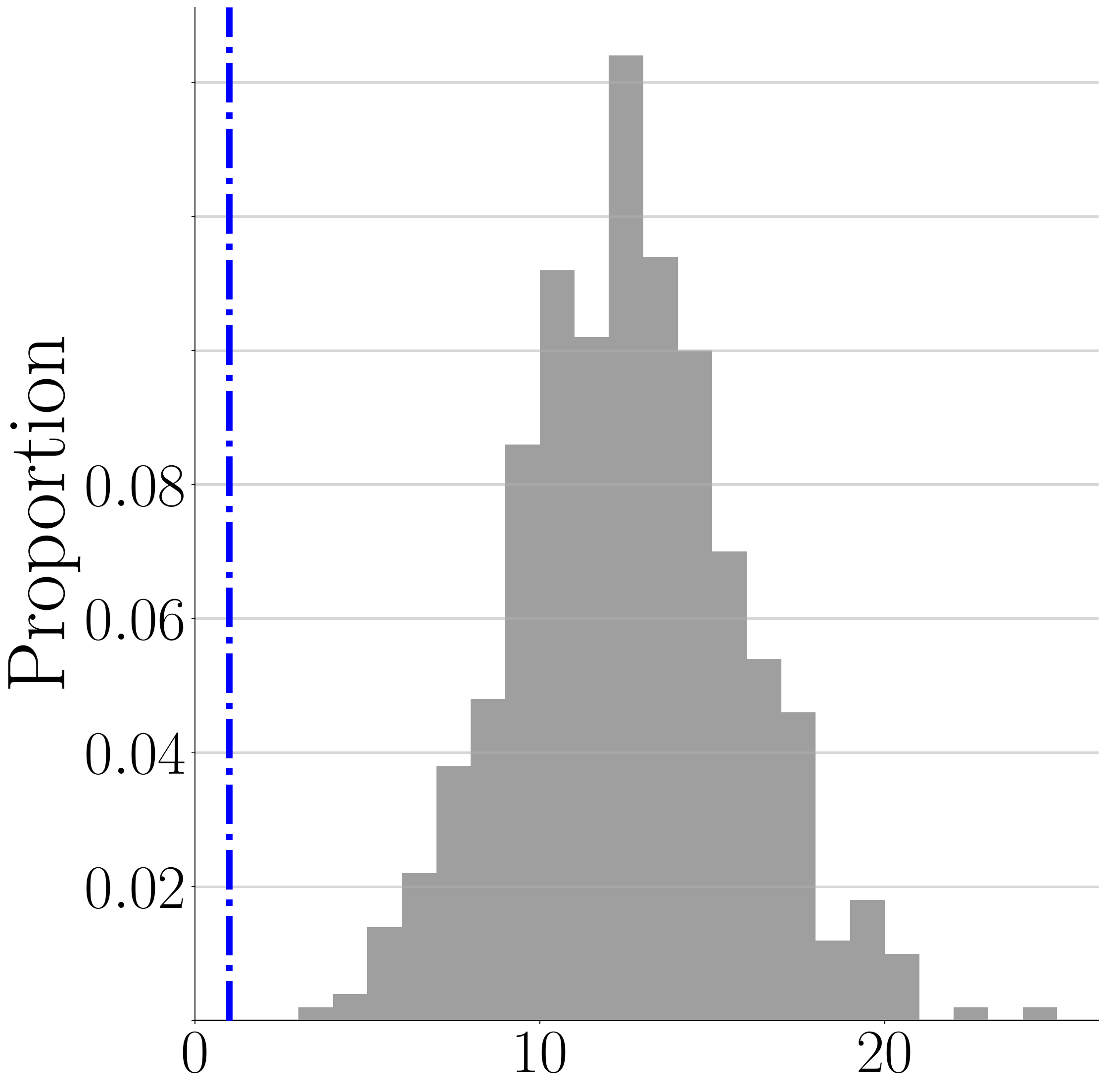}
       \\[-1mm] 
        \quad \quad $\numint[1]$
        & \quad \quad  $\numint[1]^+$
        &  \quad \quad $\numint[1]^-$ \\ %
        (a) & (b) & (c)
\end{tabular}
}
\caption{
\tbf{Histogram of  the number of interesting users of type $1$, $\numint[1]$~\cref{eq:numint1} and its one-sided variant across $500$ trials. 
}  Here panel~(a) with $\numint[1] \defeq \sum_{i=1}^n \indicator(|\intscore[1] -0.5|\geq 0.4)$ reproduces \cref{fig:numint1_null_hist} for the reader's convenience. In panels (b) and (c), the one-sided interesting users of type $1$ are defined as $\numint[1]^+ \defeq \sum_{i=1}^n \indicator(\intscore[1] \geq 0.9)$ and $\numint[1]^- \defeq \sum_{i=1}^n \indicator(\intscore[1] \leq 0.4)$ respectively. The trial data is the same as that in \cref{fig:numint1_null_hist}. That is, each is composed of 63 resampled trajectories.  The trajectories are  generated such that  the true advantage is zero.  The proportions on the vertical axis represent the fraction of the 500 trials, with the value of $\numint[1], \numint[1]^+,$ and $\numint[1]^-$ on the horizontal axis.  The vertical blue dashed line in each panel marks the corresponding interesting user count observed in the original data.
\label{fig:int1_oneSided_histograms}
}
\end{figure}

From \cref{fig:int1_oneSided_histograms}(b), we find that in the original data 17 users exhibit $\intscore[1] \geq 0.9$; we denote this user count by $\numint[1]^+$. However, the value of $\numint[1]^+$ is always significantly smaller than $17$ across the 500 trials with resampled trajectories. In \cref{tab:results}, we denote this analysis as Type $1^+$.

On the other hand, \cref{fig:int1_oneSided_histograms}(c) shows that one user exhibits $\intscore[1]\leq 0.1$ in the original data; we denote this count by $\numint[1]^-$. We also find that all 500 trials have $\numint[1]^->1$.   In \cref{tab:results}, we denote this analysis as Type $1^-$.

Overall, we conclude that the data presents evidence in favor that the RL algorithm is potentially personalizing by learning that many users benefit from sending an activity message. However, many users might exhibit $\intscore[1] \leq 0.1$ and it might appear that sending the message is less beneficial than not sending for these users, just due to algorithmic stochasticity. Consequently, the value of $\numint[1]$---the number of interesting users with $|\intscore[1]-0.5| \geq 0.4$, which is also equal to $\numint[1]^+ + \numint[1]^-$---can be as high as 18 (the observed value in the original data) due to algorithmic stochasticity.

\paragraph{Stability of conclusions with respect to the choice of $(\delta, \gamma)$}
Next, we investigate the stability of the claims made above for $\intscore[1]$ and the one-sided variants to the choice of hyper-parameters $\delta$ and $\gamma$, appearing in \cref{eq:numint1,eq:good_days}, respectively. Note that for a given definition of \intscore[], increasing $\gamma$ in \cref{eq:good_days} for a fixed $\delta$ in \cref{eq:numint1} allows more users to become eligible for being considered as interesting, both in original data and the resampled trials. Similarly, decreasing $\delta$ in \cref{eq:numint1} for a fixed $\gamma$ in \cref{eq:good_days} would typically lead to more number of interesting users, both in original data and the resampled trials. 

The results of this exploration for the choices $\delta \in \sbraces{0.35, 0.40, 0.45}$, and $\gamma \in \sbraces{0.65, 0.70, 0.75}$ are presented in \cref{heatmaps}. For a given panel, the value in the cell corresponding to the value of $\delta$ on the horizontal axis and $\gamma$ on the vertical axis is equal to the fraction of the 500 trials for which the number of interesting users $\numint[]$ computed using those hyperparameter choices was as at least as large that in was at least as large as that in the original data. Looking at \cref{heatmaps}(b,c), we find the conclusions drawn from \cref{fig:int1_oneSided_histograms}(b,c) with $(\delta, \gamma)=(0.4, 0.75)$ about $\numint[1]^+$ and $\numint[1]^-$ remains stable even if we slightly perturb the values of $\delta$ and $\gamma$. In particular, across the $3\times 3$ choices for $\delta$ and $\gamma$, the value of $\numint[1]^+$ would \emph{not} appear as high as the observed value in the original data just by chance. On the other hand, the value of $\numint[1]^-$ might appear higher than the observed value in the original data simply due to algorithmic stochasticity. Given the competing nature of these two quantities and the fact that $\numint[1] = \numint[1]^+ + \numint[1]^-$, the resulting fraction of trials with a count at least as high as $\numint[1]$ in the original data is quite sensitive to the particular choice of $(\delta, \gamma)$ as \cref{fig:int1_oneSided_heatmaps}(a) illustrates.

\begin{figure}[ht!]
\centering
\resizebox{\textwidth}{!}{
\begin{tabular}{ccc}
    \includegraphics[width=0.33\textwidth]{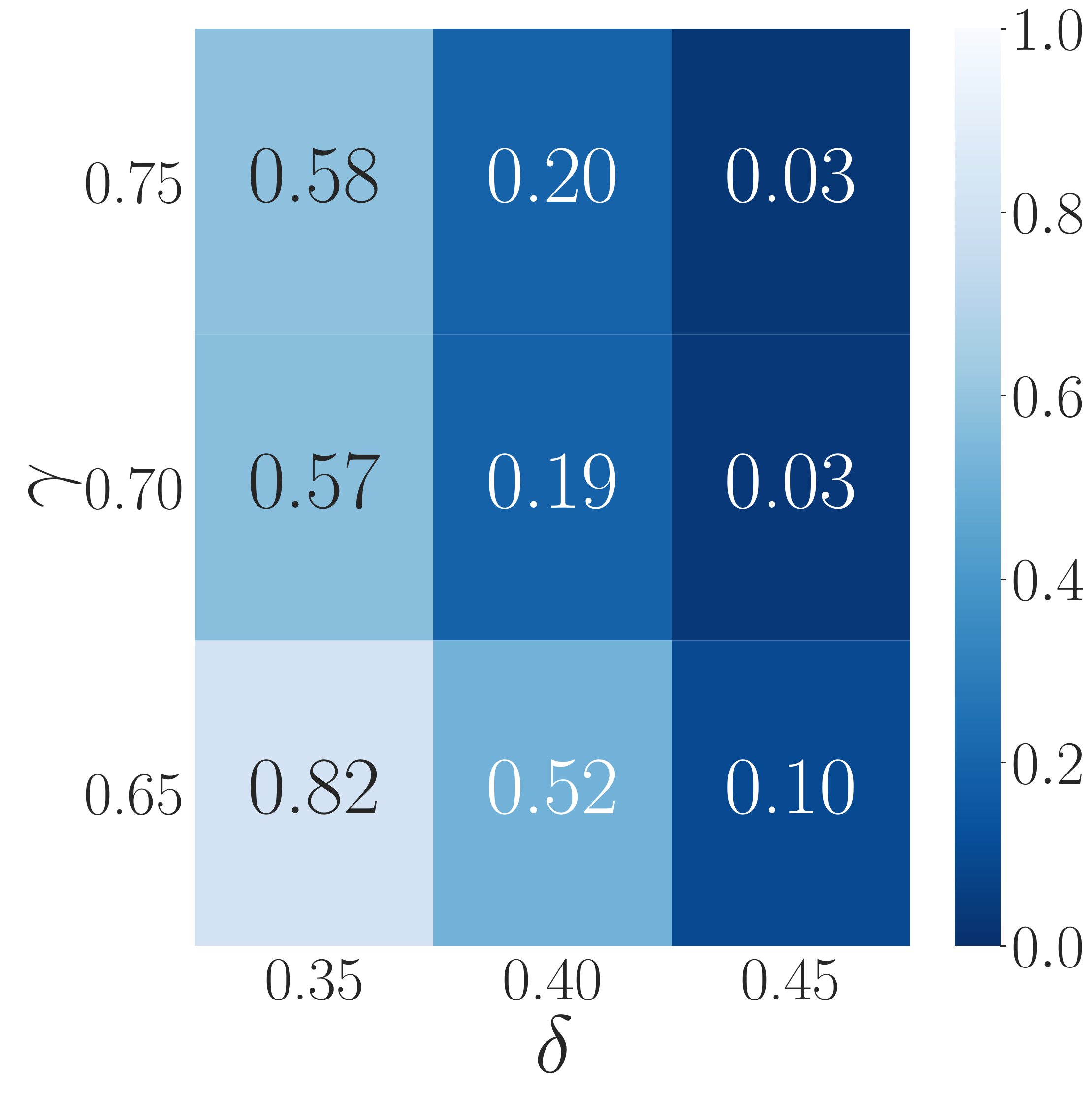}  & \includegraphics[width=0.33\textwidth]{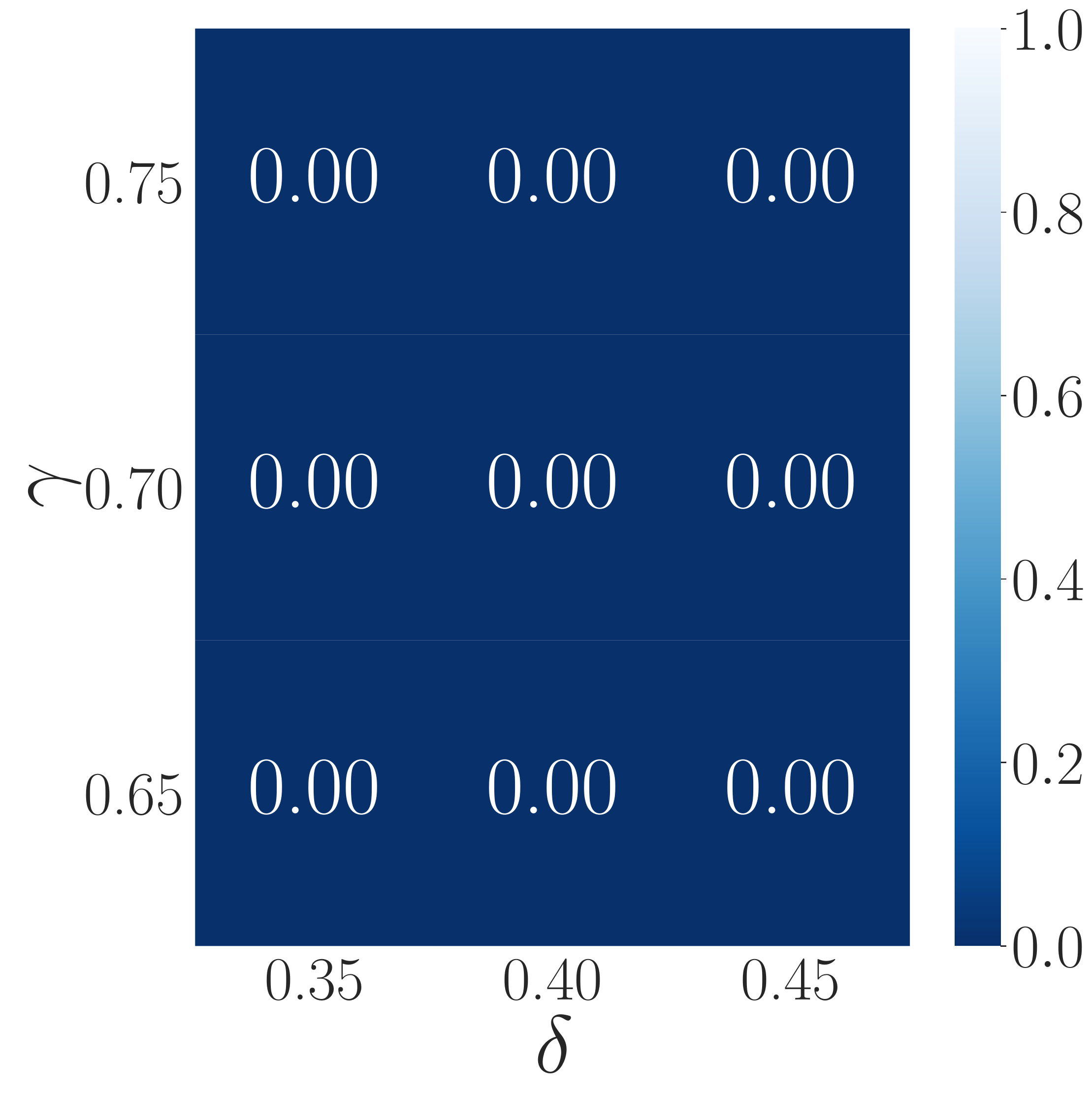}  
     & \includegraphics[width=0.33\textwidth]{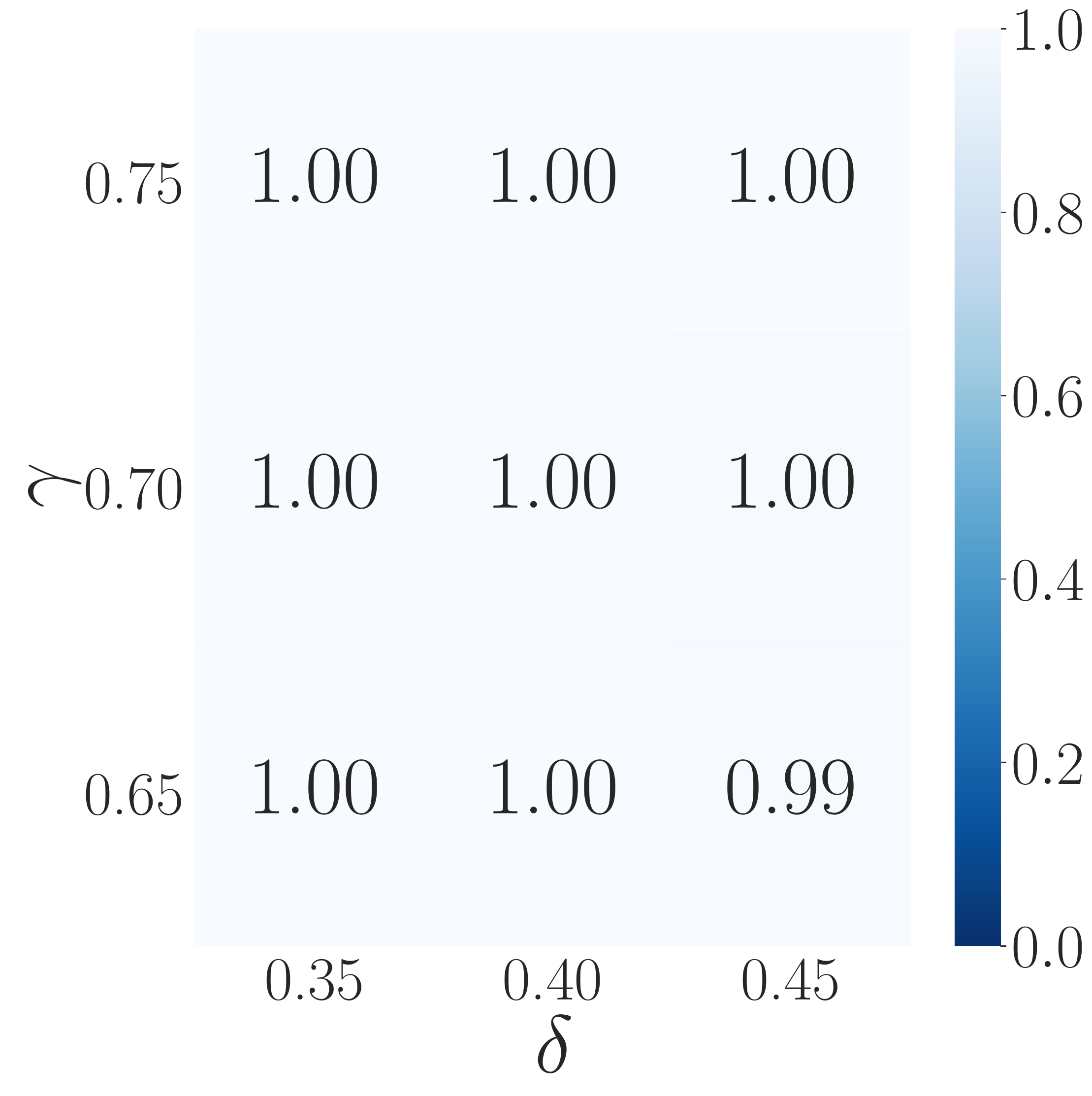}\\
     (a) $\numint[1]$
        & (b) $\numint[1]^+$
        &  (c) $\numint[1]^-$
\end{tabular}
}
\caption{\tbf{Stability of conclusions from \cref{fig:numint1_null_hist,fig:int1_oneSided_histograms} for interestingness of type $1$ with respect to the choice of hyperparameters $(\delta, \gamma)$.} Panels (a) to (c) respectively plot the results for the number of interesting users of type 1, $\numint[1]\defeq \sum_{i=1}^n \indicator(|\intscore[1] -0.5|\geq \delta)$, and its one-sided variants, namely $\numint[1]^+\defeq \sum_{i=1}^n \indicator(\intscore[1] \geq 0.5+\delta)$, and  $\numint[1]^-\defeq \sum_{i=1}^n \indicator(\intscore[1] \leq 0.5-\delta)$. For a given panel, the value in the cell corresponding to $\delta$ on the horizontal axis and $\gamma$ on the vertical  axis is equal to the fraction of the 500 trials for which the number of interesting users was at least as large as that in 
the original data.  Recall that the full histogram of $\numint[1],\numint[1]^+,$ and $\numint[1]^-$ across these 500 trials  in \cref{fig:numint1_null_hist,fig:int1_oneSided_histograms} correspond to the choice of $\delta=0.4$ and $\gamma=0.75$. \label{heatmaps_one}
}
\label{fig:int1_oneSided_heatmaps}
\end{figure}

\section{Stability of HeartSteps results for interestingness of type 2}
\label{sec:additional_results_heartsteps}
We perform stability analysis for $\numint[2,\var]$ with respect to the choice of $(\delta, \gamma)$ similarly to that done for $\numint[1]$ above in \cref{fig:int1_oneSided_heatmaps} and provide the results in \cref{heatmaps}.

Panels (a), (b), and (c) display the results, respectively, for $\var= \variation, \location$, and \engagement. In a given panel, the value in the cell corresponding to the value of $\delta$ on the horizontal axis and $\gamma$ on the vertical axis is equal to the fraction of the 500 trials for which the number of interesting users $\numint[2, \var]$ computed using those hyperparameter choices was as at least as large that in the original data.  Across the $3\times 3$ choices for $\delta$ and $\gamma$, we notice that for interestingness of type $2$ for the features \variation, \location, and \engagement, the fraction remains stable around 0, 0, and 1, same as the fraction in \cref{fig:numint2_null_hist} for $(\delta, \gamma)=(0.4, 0.75)$.

\begin{figure}[ht!]
\centering
 \resizebox{\textwidth}{!}{
\begin{tabular}{ccc}
     \includegraphics[width=0.33\textwidth]{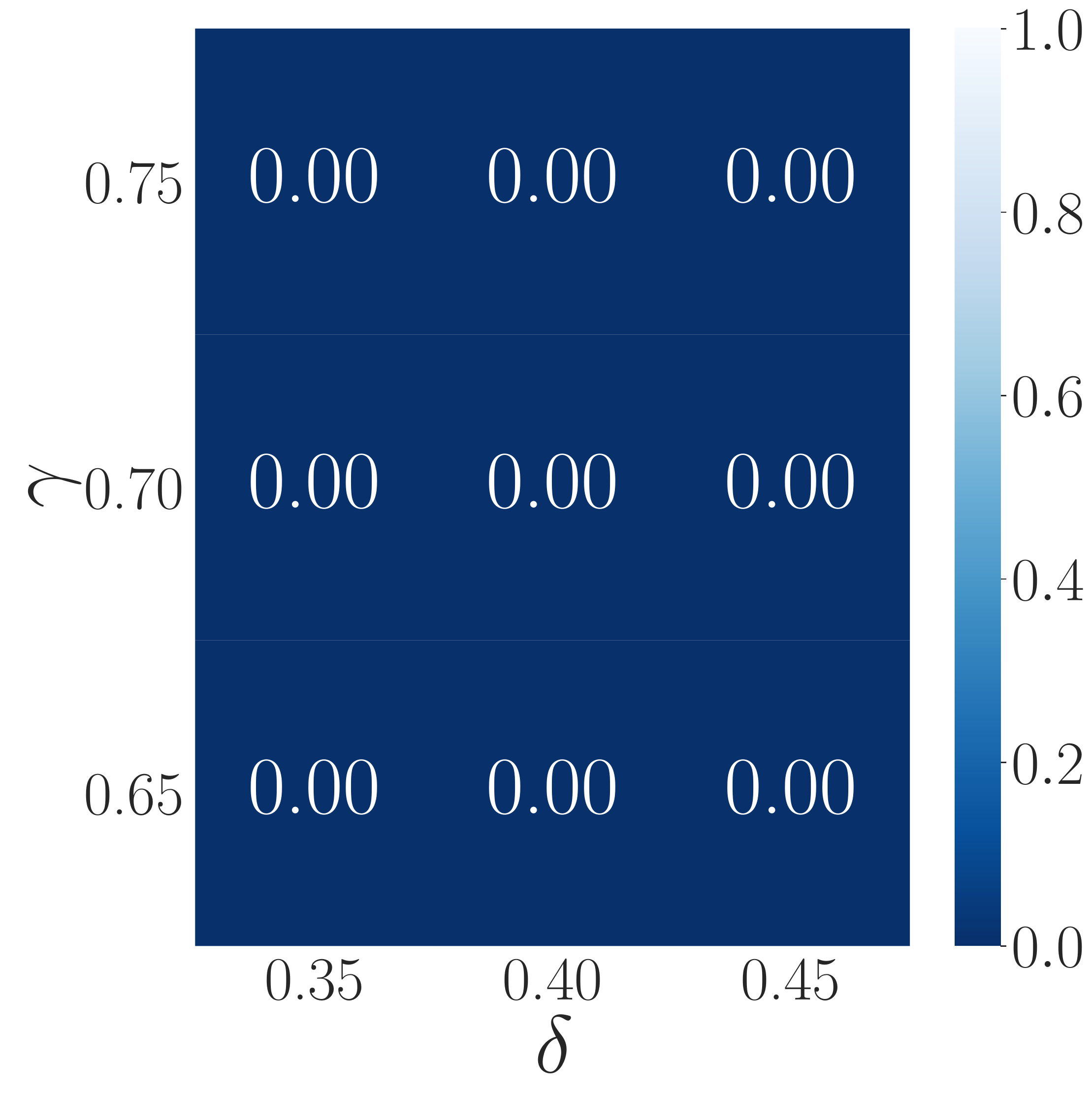} 
     &
     \includegraphics[width=0.33\textwidth]{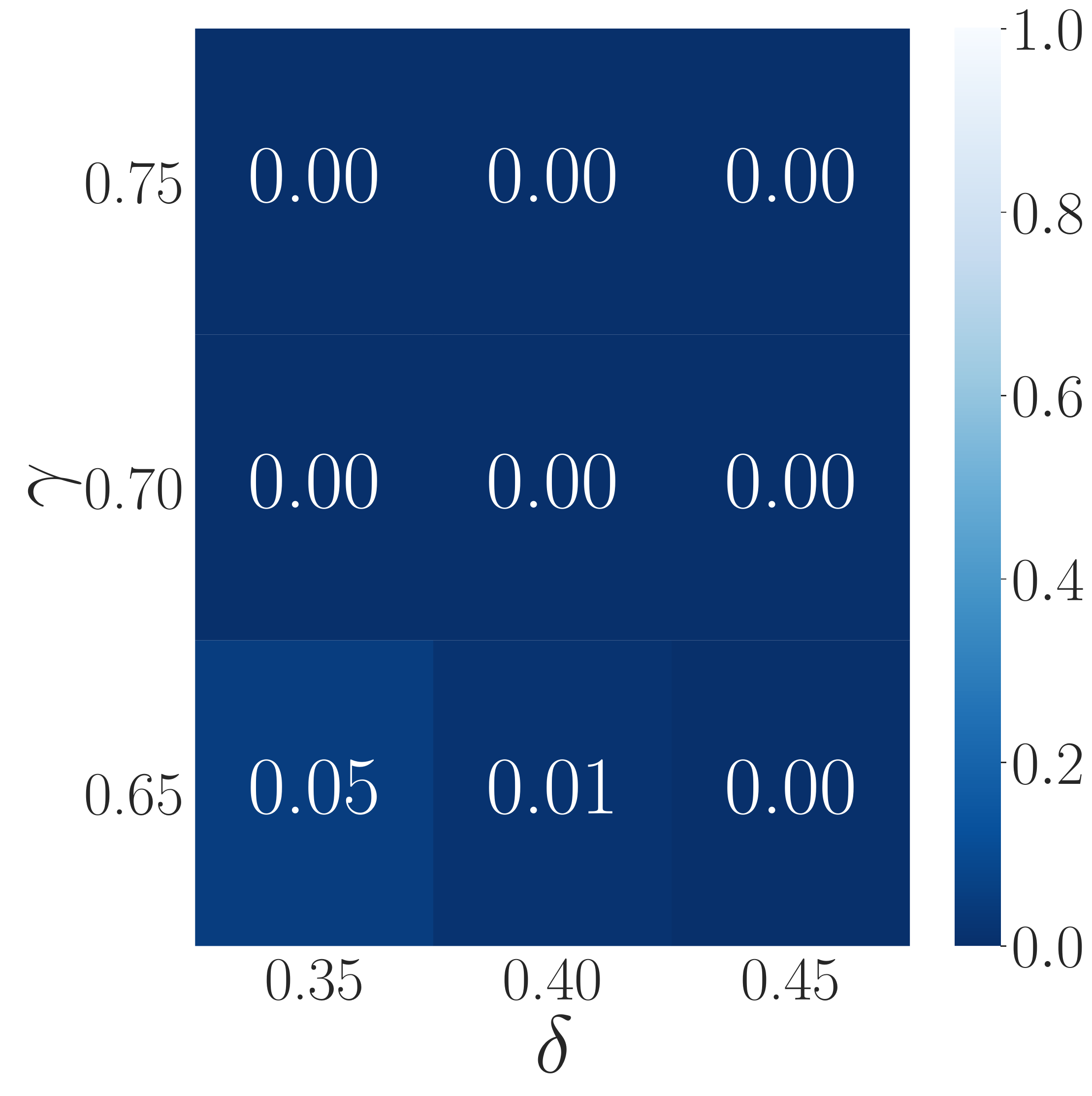}
     &
     \includegraphics[width=0.33\textwidth]{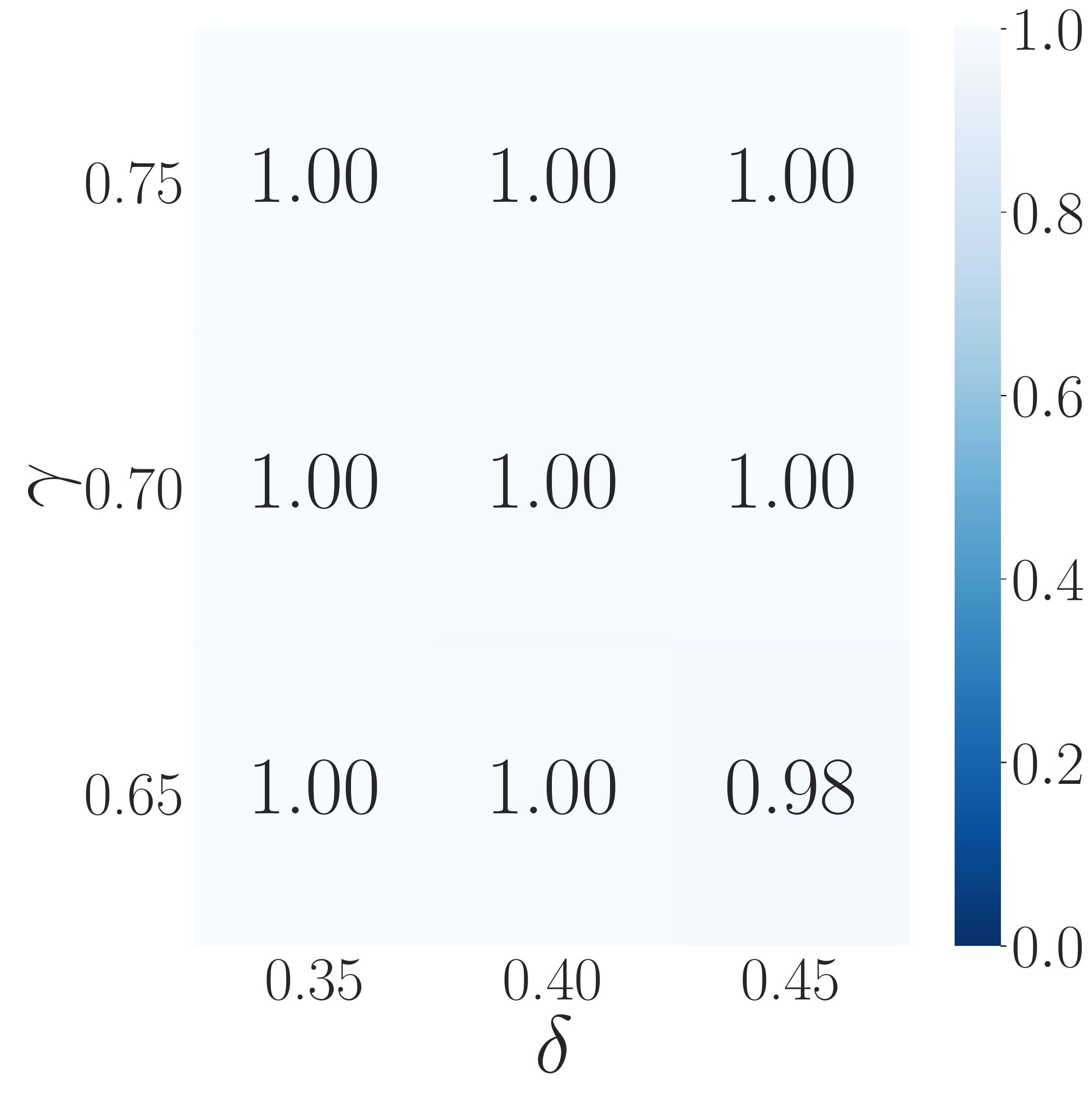}
     \\
     (a) \variation  & (b) \location  & (c) \engagement
\end{tabular}
}
\caption{\tbf{Stability of conclusions from \cref{fig:numint2_null_hist} for interestingness of type $2$ with respect to the choice of hyperparameters $(\delta, \gamma)$.} Panels (a) to (c) respectively plot the results for $\numint[2,\var]$ for interestingness of type $2$ for feature $\var\in\sbraces{\variation, \location, \engagement}$.
For a given panel, the value in the cell corresponding to $\delta$ on the horizontal axis and $\gamma$ on the vertical 
axis is equal to the fraction of the 500 trials for which the number of interesting users $\numint[2, \var]$ was at least as large as that in
the original data.  Recall that the full histogram of $\numint[2,\var]$ across these 500 trials 
in \cref{fig:numint2_null_hist} corresponds to the choice of $\delta=0.4$ and $\gamma=0.75$. \label{heatmaps}}
\end{figure}

\section{Interesting users of type 2 for \location and \engagement}
We now demonstrate the analysis (like in  \cref{fig:interestingUser}(b) and \cref{fig:user_int2_variation}) for two different users, who exhibit potential interestingness of type 2 for \location and \engagement.

\paragraph{A potentially interesting user of type 2 for \location}
\cref{fig:interestingUserLocation} displays the advantage forecasts for a user, who we call user 3, to distinguish them from the two users associated with \cref{fig:interestingUser}. The three panels in \cref{fig:interestingUserLocation} plot user 3's advantages color-coded by the value of the three features for that user. We find that this user admits  $\intscore[2, \variation]=0.68$, $\intscore[2, \location]=0$, and $\intscore[2, \engagement]=0.43$---so that this user would be deemed potentially interesting of type 2 for \location (and not other features) as per our definition~\cref{eq:numint2}.

\begin{figure}[ht!]
\centering
\begin{tabular}{ccc}
    \includegraphics[width=0.31\textwidth]{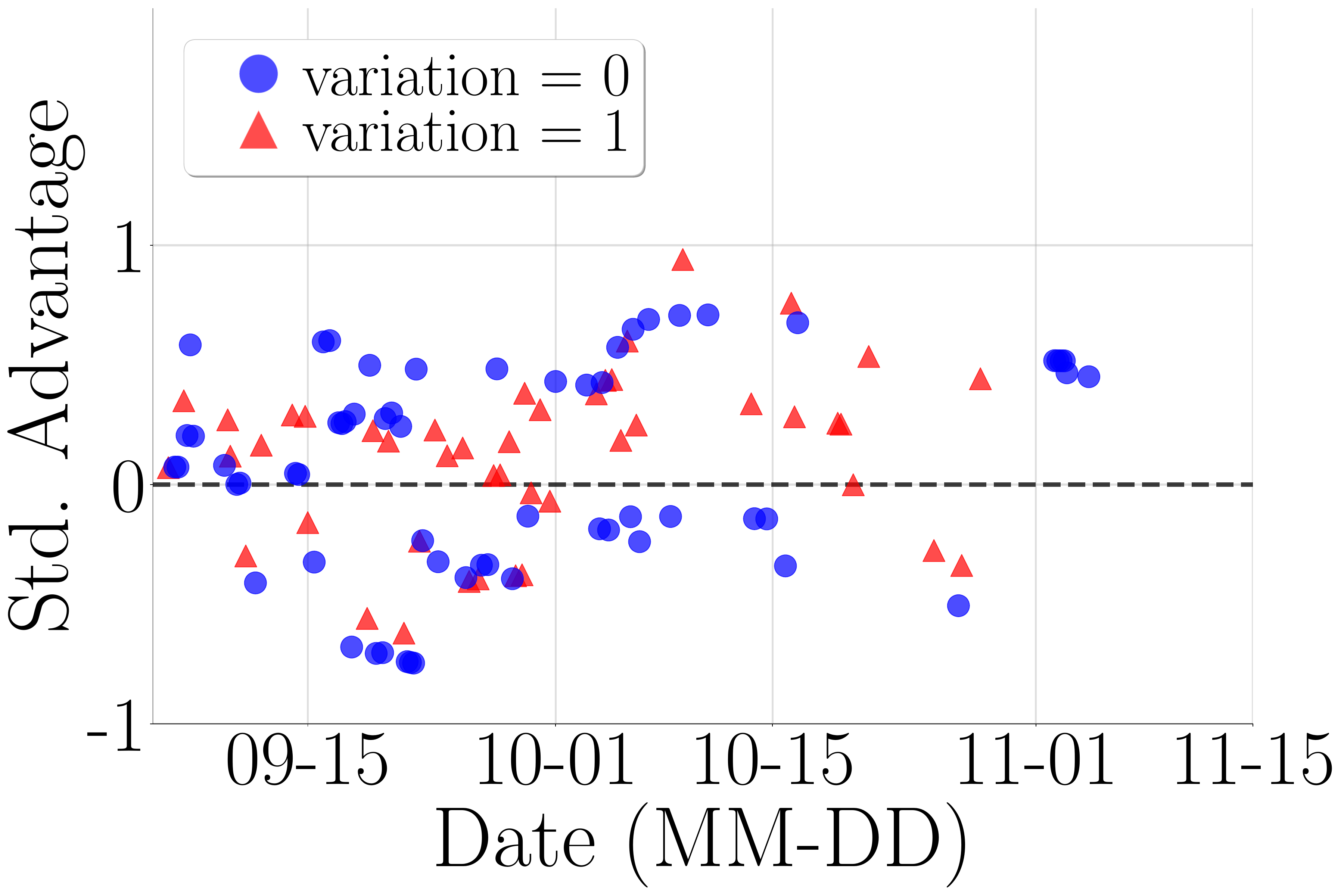} 
     & \includegraphics[width=0.31\textwidth]{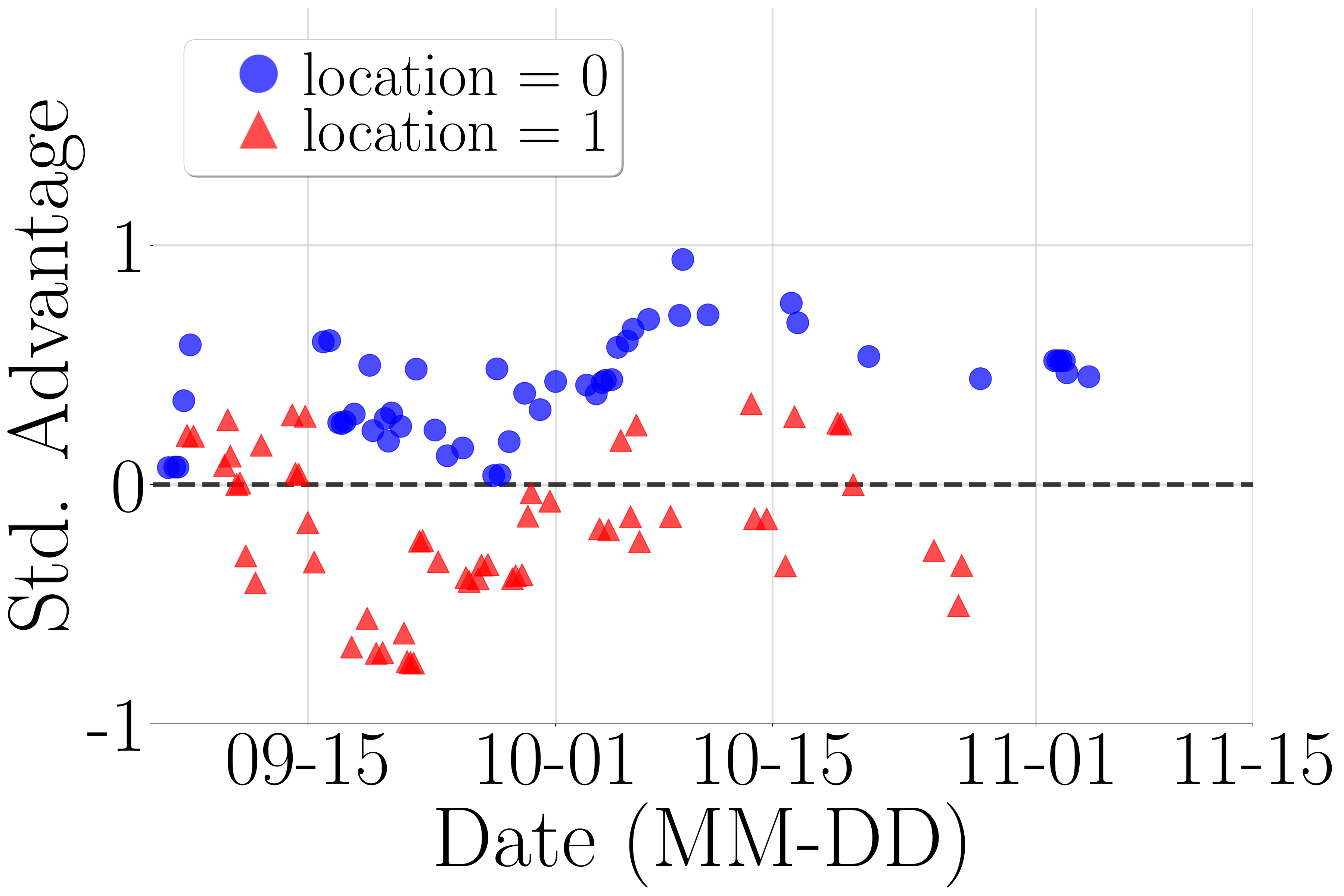} 
     & \includegraphics[width=0.31\textwidth]{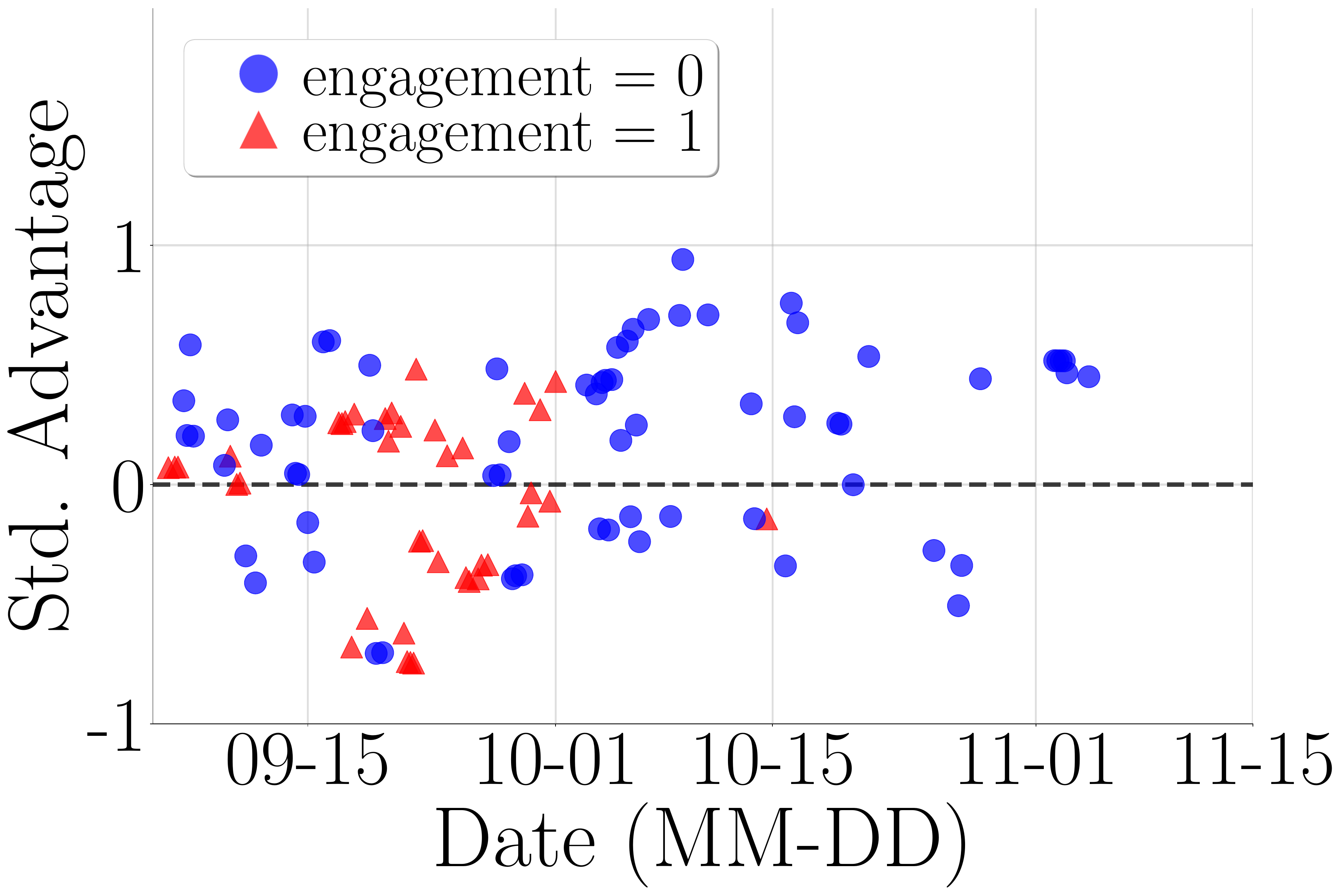}   \\
     (a) & (b) & (c)
\end{tabular}
\caption{
\tbf{Standardized advantage forecasts of user 3, an interesting user of type $2$ for \location, color-coded by \var = \variation, \location, and \engagement in panels (a), (b), and (c) respectively.} The value on the vertical axis represents the RL algorithm's forecast of the standardized advantage of sending an activity message for the user if the user was available for sending a message on the day marked on the horizontal axis. (Note each day has $5$ decision times.) The forecasts are marked as blue circles based on \var = 1 and red triangles if \var = 0 at the decision time. Panels (a) to (c) exhibit, respectively, $\intscore[2, \variation]=0.68$, $\intscore[2, \location]=0$, and $\intscore[2, \engagement]=0.43$. Note that the three panels plot the same data and differ only in the color coding. \label{fig:interestingUserLocation}}
 \end{figure}
 
Next, we evaluate how likely the user graph in \cref{fig:interestingUserLocation}(b) would appear just by chance.  Panels (a) and (b) of  \cref{fig:user_int2_location} visualize two resampled trajectories of user 3 (chosen uniformly at random from user 3's 500 resampled trajectories) generated under the generative model that there is no differential advantage of sending a message based on the value of \location. The color coding is as in \cref{fig:interestingUserLocation}(b), namely, the forecasts are marked in red triangles if \location = 1 and blue circles if \location = 0. In panel (c) of \cref{fig:user_int2_location}, we plot the histogram for the \intscore[2, \location] for this user across all 500 resampled trajectories and denote the observed value in the original data as a vertical dotted line.  %

\cref{fig:user_int2_location}(a,b) show that the resampled trajectories do not appear interesting of type $2$ for \location as in \cref{fig:interestingUserLocation}(b); the two trajectories, respectively, have $\intscore[2,\variation]=$ 0.50, and 0.42. Moreover, panel (c) shows that the interestingness score of $0$, which was observed for user 3 in the original data (\cref{fig:interestingUserLocation}(b)), never appears across any of the resampled trajectories. Thus we can conclude that the data presents evidence that the RL algorithm potentially personalized for user 3 by learning to treat the user based on \location differentially, and this personalization would not likely arise simply due to algorithmic stochasticity. 

\begin{figure}[ht!]
\centering
\begin{tabular}{ccc}
     \includegraphics[width=0.31\textwidth]{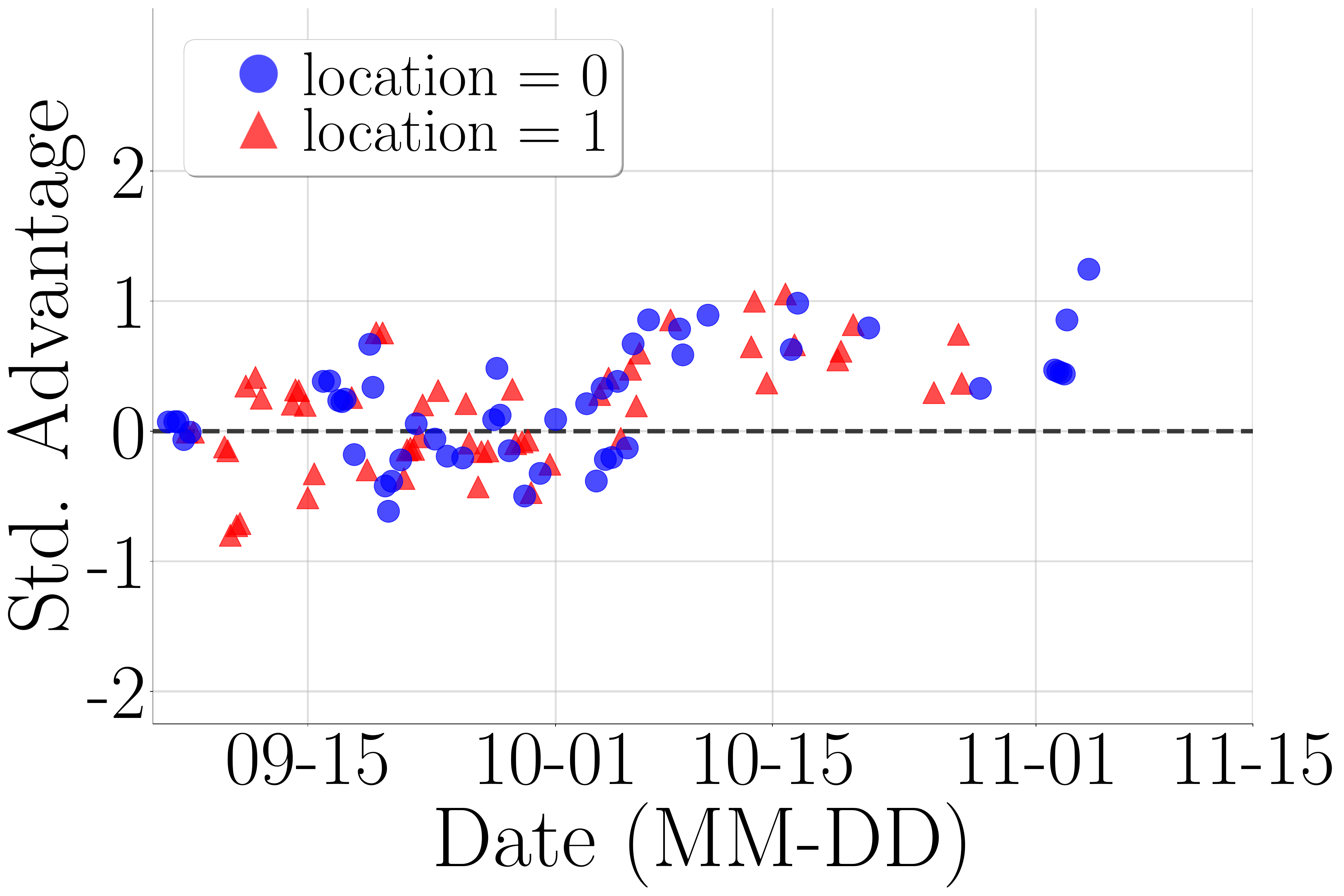} 
     & \includegraphics[width=0.31\textwidth]{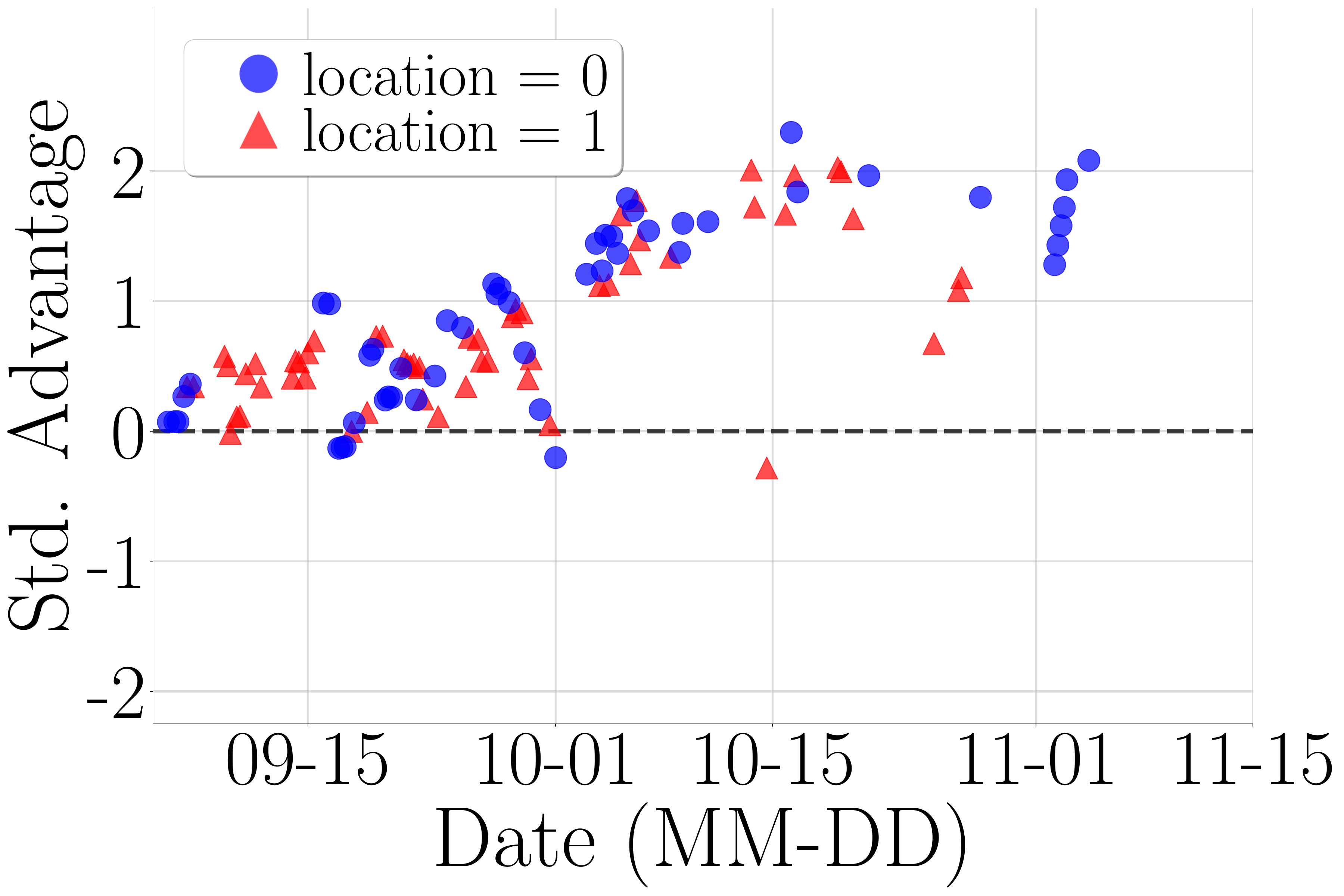} 
     & \includegraphics[width=0.31\textwidth]{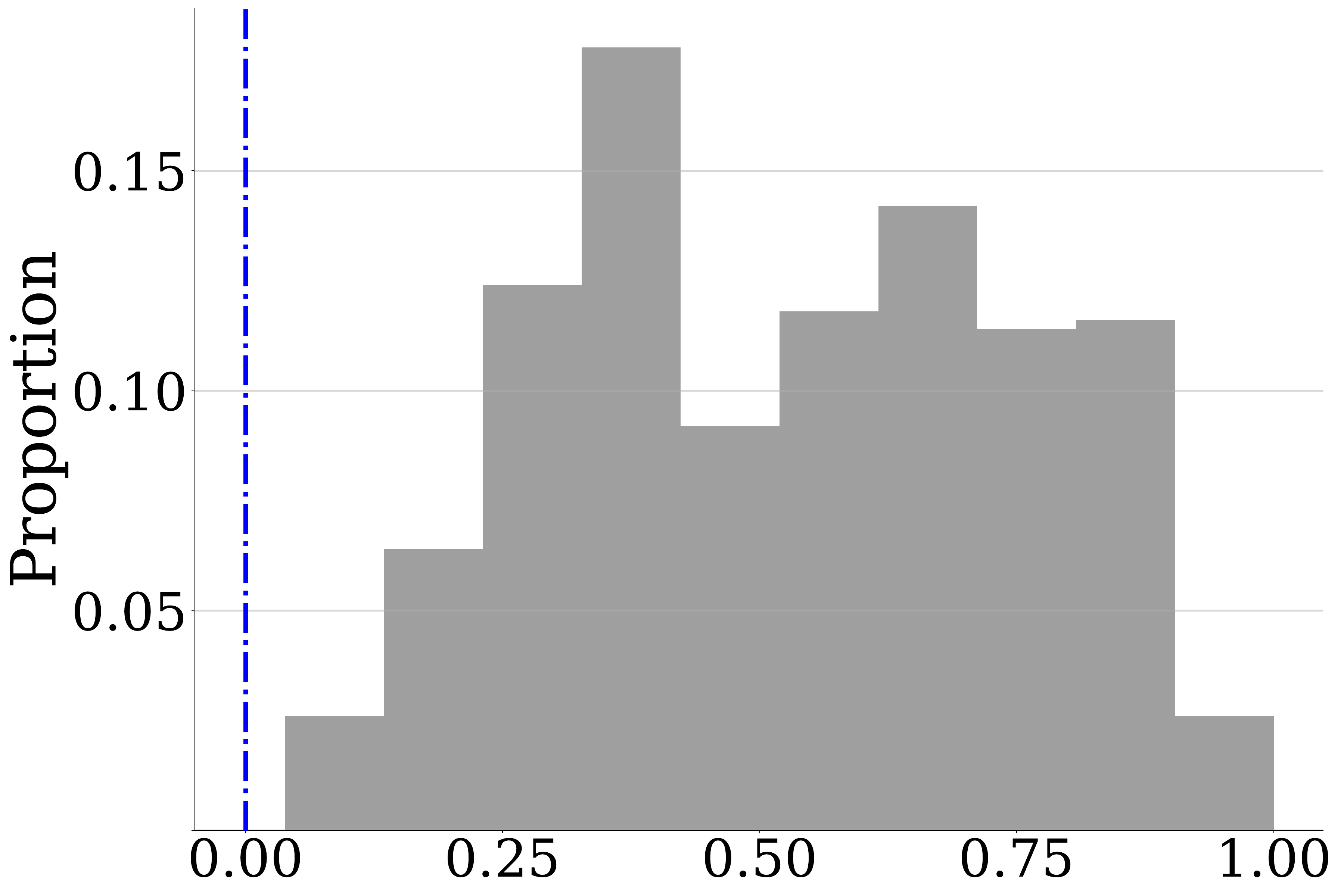}  \\
     \qquad Resampled trajectory 1 &
     \qquad Resampled trajectory 2 &
     \quad \qquad $\intscore[2,\location]$ \\
     (a)  & (b) & (c)
\end{tabular}
\caption{\tbf{Resampling results for user 3 considered in \cref{fig:interestingUserLocation}, whose original trajectory exhibits interestingness of type $2$ for $\location$.} Panels (a) and (b) plot two randomly chosen (out of 500) resampled trajectories generated with zero advantage; the two trajectories, respectively, have $\intscore[2,\location]=$ 0.50, and 0.42. In panel (c), the vertical axis represents the fraction of the 500 resampled trajectories for this user with the value of $\intscore[2,\location]$ on the horizontal axis; and the vertical blue dashed line marks the observed $\intscore[2,\location]$ (value 0) for this user. \label{fig:user_int2_location}}
\end{figure}

\paragraph{A potentially interesting user of type 2 for \engagement}

\cref{fig:interestingUserEngagement} displays the advantage forecasts for a user, who we call user 4, to distinguish them from the three users associated with \cref{fig:interestingUser,fig:interestingUserLocation}. The three panels in \cref{fig:interestingUserEngagement} plot user 4's advantages color-coded by the three features; which admit $\intscore[2, \variation]=0.65$, $\intscore[2, \location]=0.9$, and $\intscore[2, \engagement]=0.037$, respectively. Thus based on our definition~\cref{eq:numint2}, this user is potentially interesting of type 2 for \engagement, but not for \variation. The user does not qualify the criterion ($\gamma = 0.75$ in \cref{eq:good_days}) for being considered as a potentially interesting user for \location due to a lack of diversity in the values taken by its \location feature.

 \begin{figure}[ht!]
\centering
\begin{tabular}{ccc}
    \includegraphics[width=0.31\textwidth]{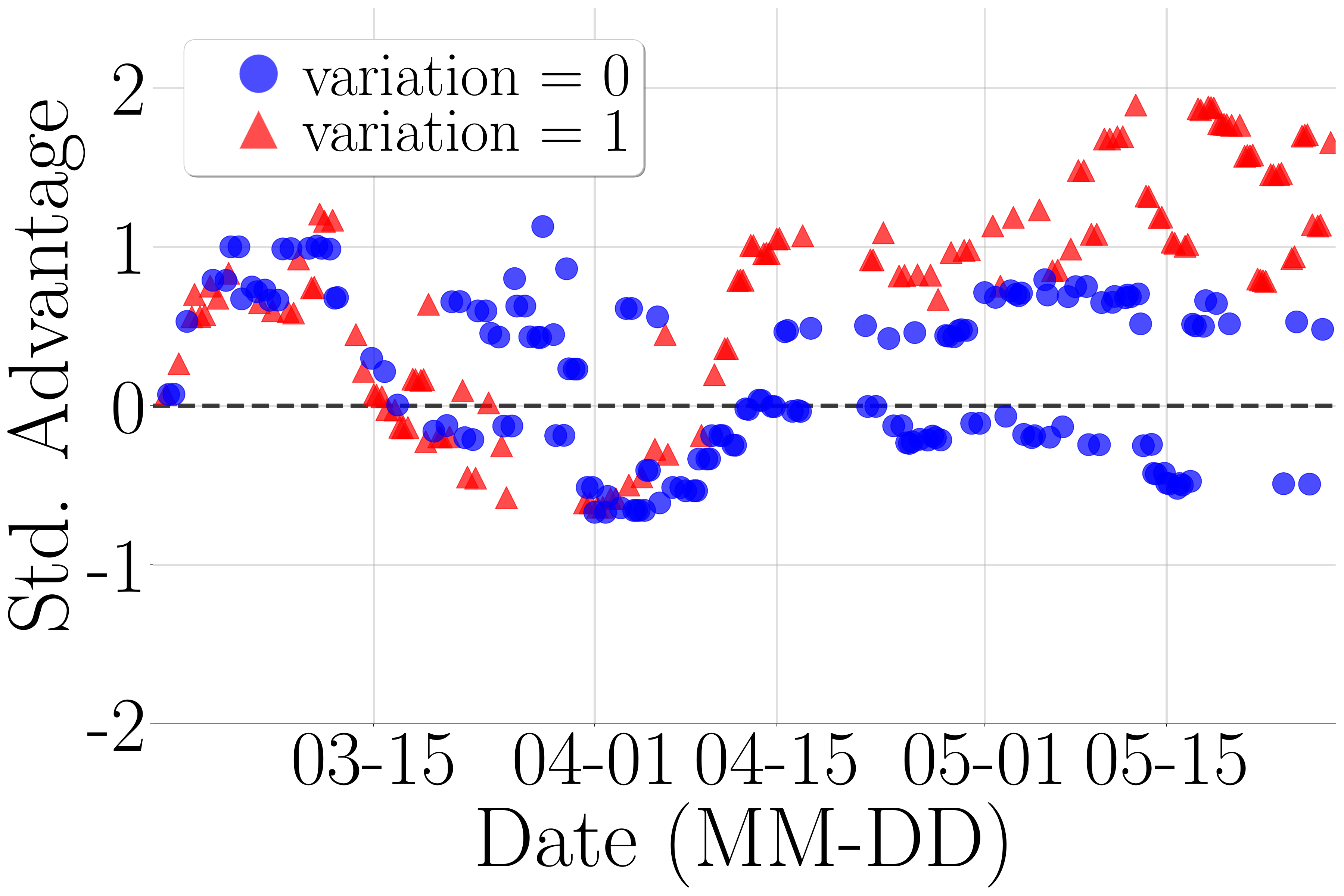} 
     & \includegraphics[width=0.31\textwidth]{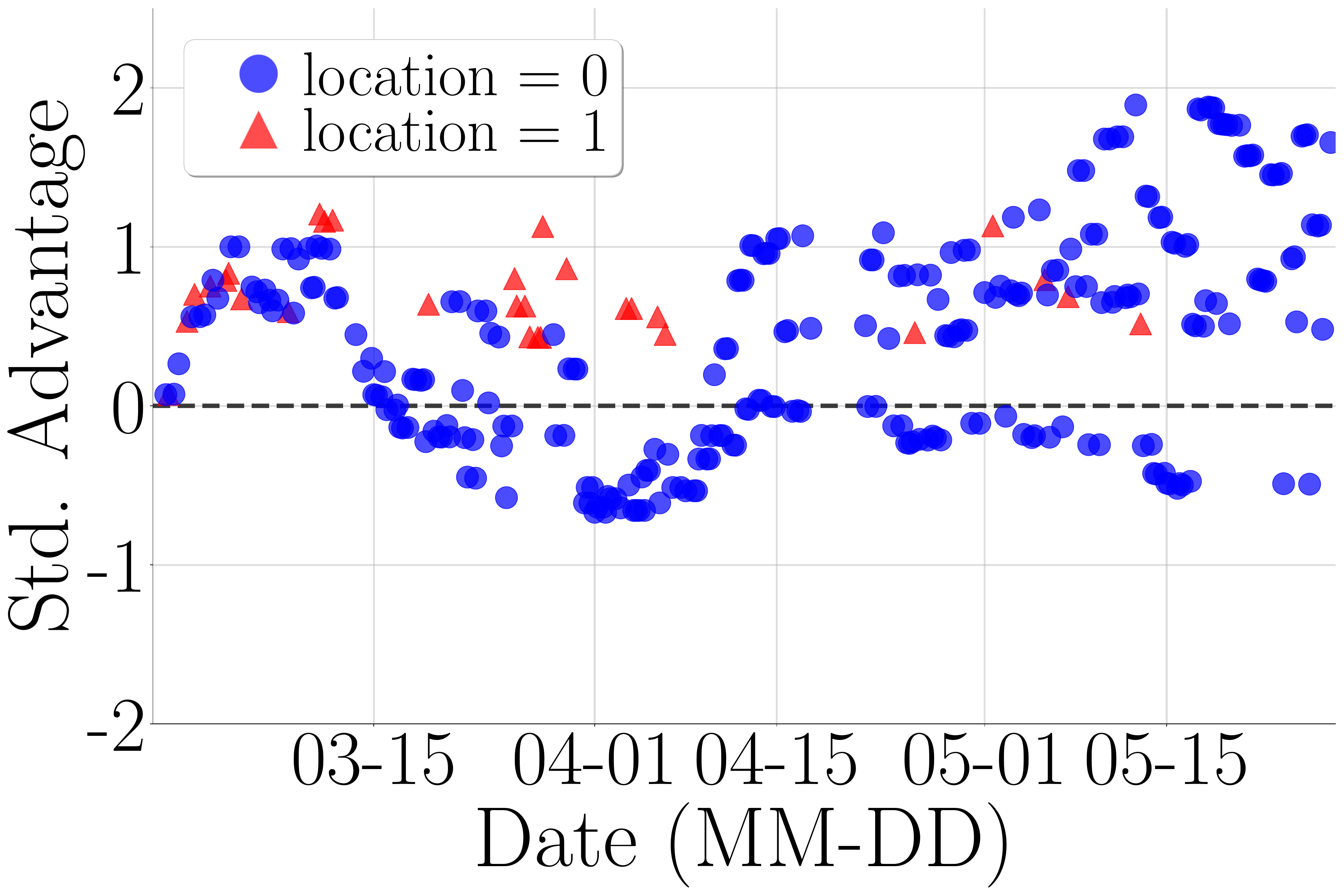} 
     & \includegraphics[width=0.31\textwidth]{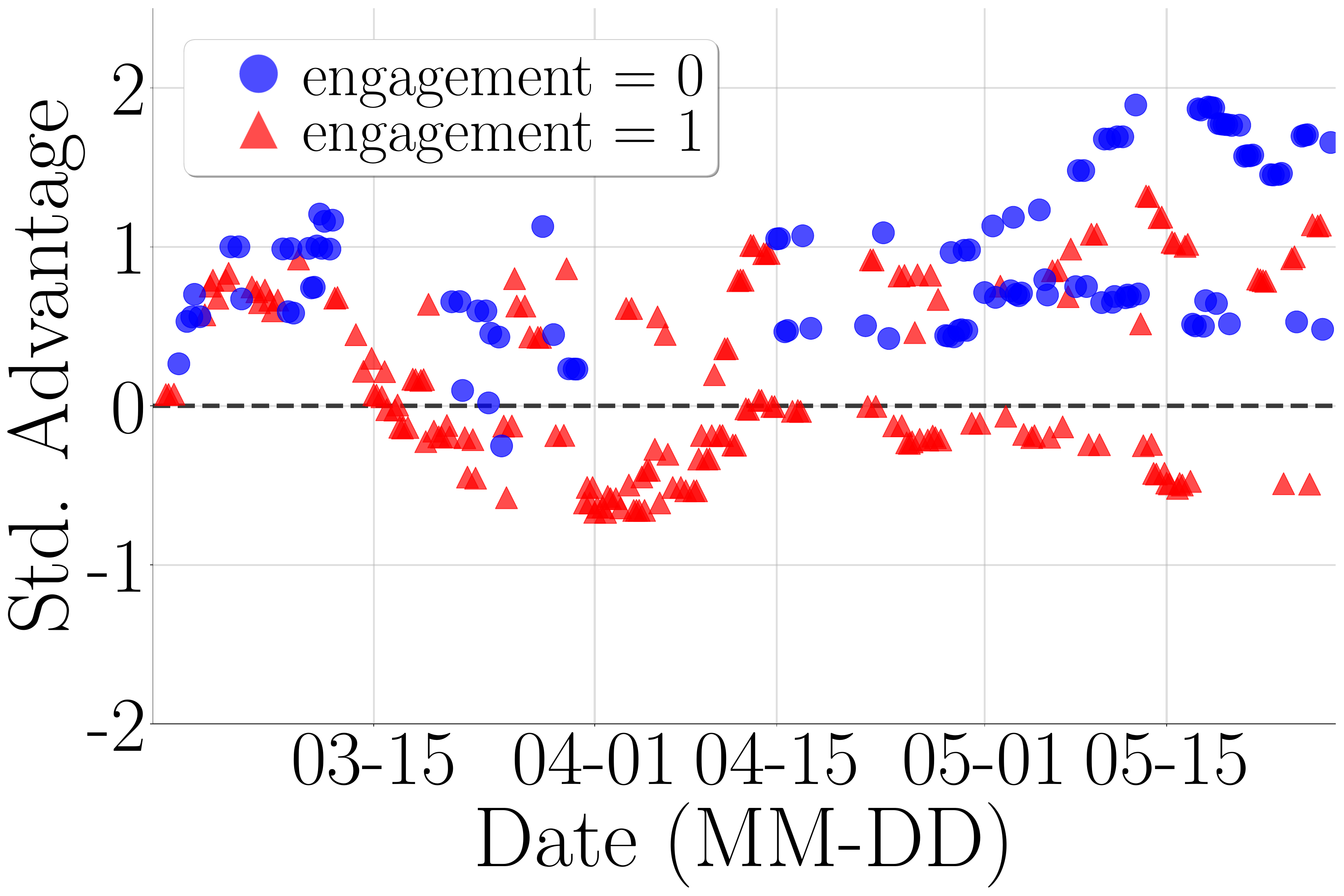}   \\
     (a) & (b) & (c)
\end{tabular}
\caption{\tbf{Standardized advantage forecasts of user 4, an interesting user of type $2$ for \engagement, color-coded by \var = \variation, \location, and \engagement in panels (a), (b), and (c) respectively.} The value on the vertical axis represents the RL algorithm's forecast of the standardized advantage of sending an activity message for the user if the user was available for sending a message on the day marked on the horizontal axis. (Note each day has $5$ decision times.) The forecasts are marked as blue circles based on \var = 1 and red triangles if \var = 0 at the decision time. Panels (a) to (c) exhibit, respectively, $\intscore[2, \variation]=0.65$, $\intscore[2, \location]=0.9$, and $\intscore[2, \engagement]=0.037$. Note that the three panels plot the same data and differ only in the color coding. 
\label{fig:interestingUserEngagement}}
 \end{figure}

Next, we evaluate how likely the user graph in \cref{fig:interestingUserEngagement}(c) would appear just by chance.  Panels (a) and (b) of  \cref{fig:user_int2_engagement} visualize two resampled trajectories of user 4 (chosen uniformly at random from user 4's 500 resampled trajectories) generated under the generative model that there is no differential advantage of sending a message based on the value of \engagement. The color coding is as in \cref{fig:interestingUserEngagement}(c), namely, the forecasts are marked in red triangles if \engagement = 1 and blue circles if \engagement = 0. In panel (c) of \cref{fig:user_int2_engagement}, we plot the histogram for the \intscore[2, \engagement] for this user across all 500 resampled trajectories and denote the observed value in the original data as a vertical dotted line.  

\cref{fig:user_int2_engagement}(a,b) show that the resampled trajectories do not appear interesting of type $2$ for \engagement as in \cref{fig:interestingUserEngagement}(c); the two trajectories, respectively, have $\intscore[2,\engagement]=$ 0.94, and 0.41. However, panel (c) shows that the interestingness score of $0.037$, which was observed for user 4 in the original data (\cref{fig:interestingUserEngagement}(c)), appears for around 20\% of the resampled trajectories. Thus we can conclude that the data presents evidence that user 4's interestingness score for \engagement might appear extreme simply due to algorithmic stochasticity.

\begin{figure}[ht!]
\centering
\begin{tabular}{ccc}
     \includegraphics[width=0.31\textwidth]{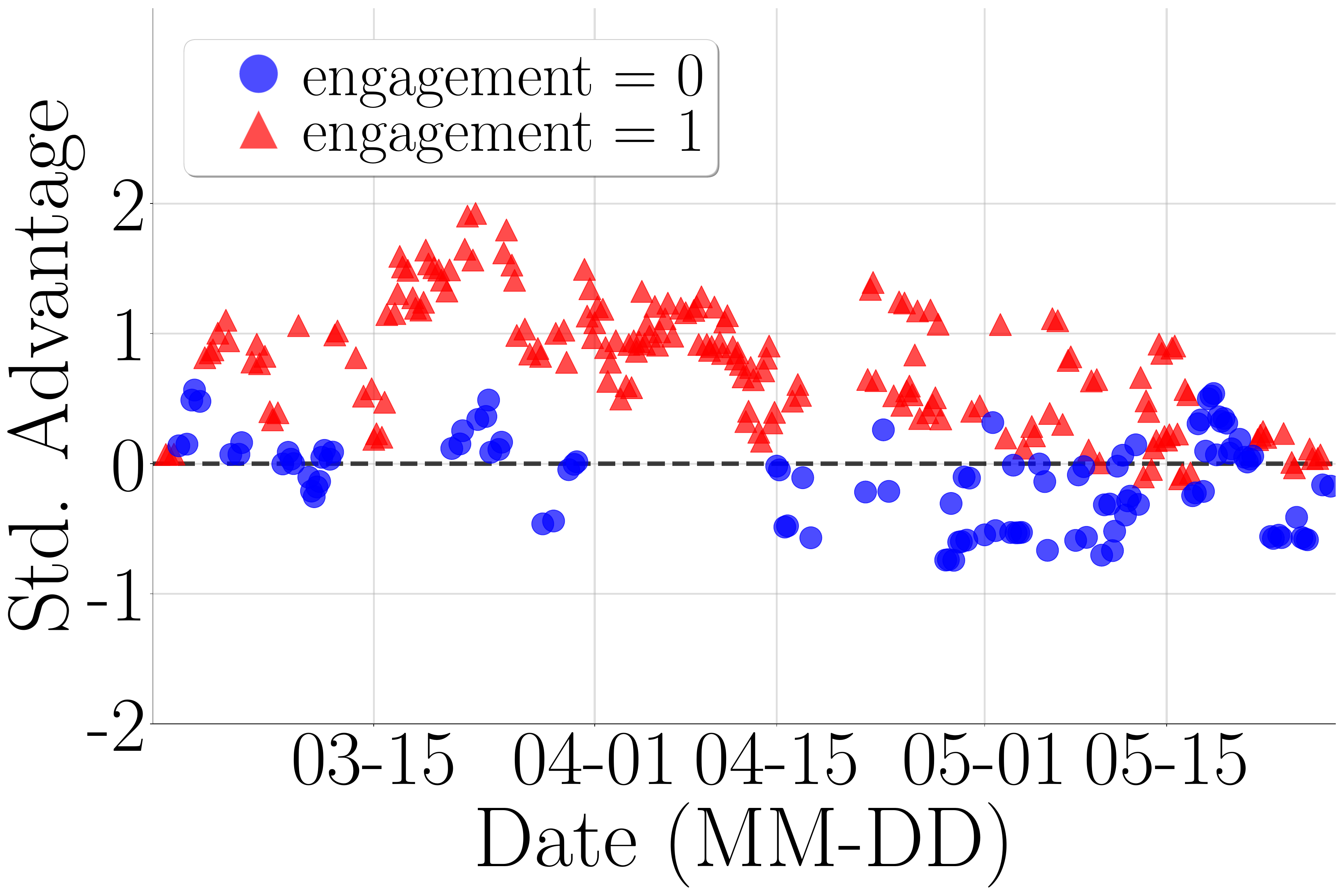} 
     & \includegraphics[width=0.31\textwidth]{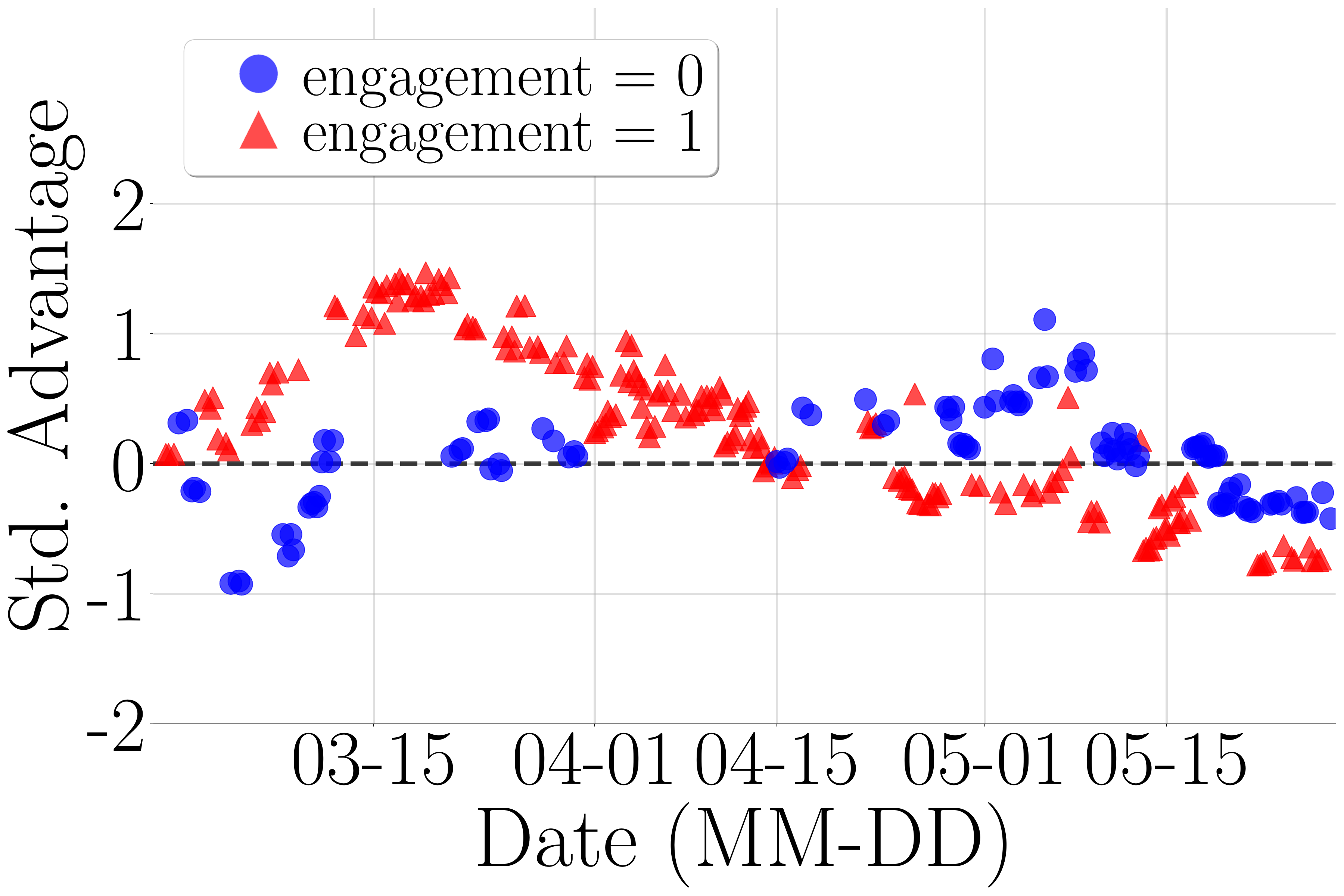} 
     & \includegraphics[width=0.31\textwidth]{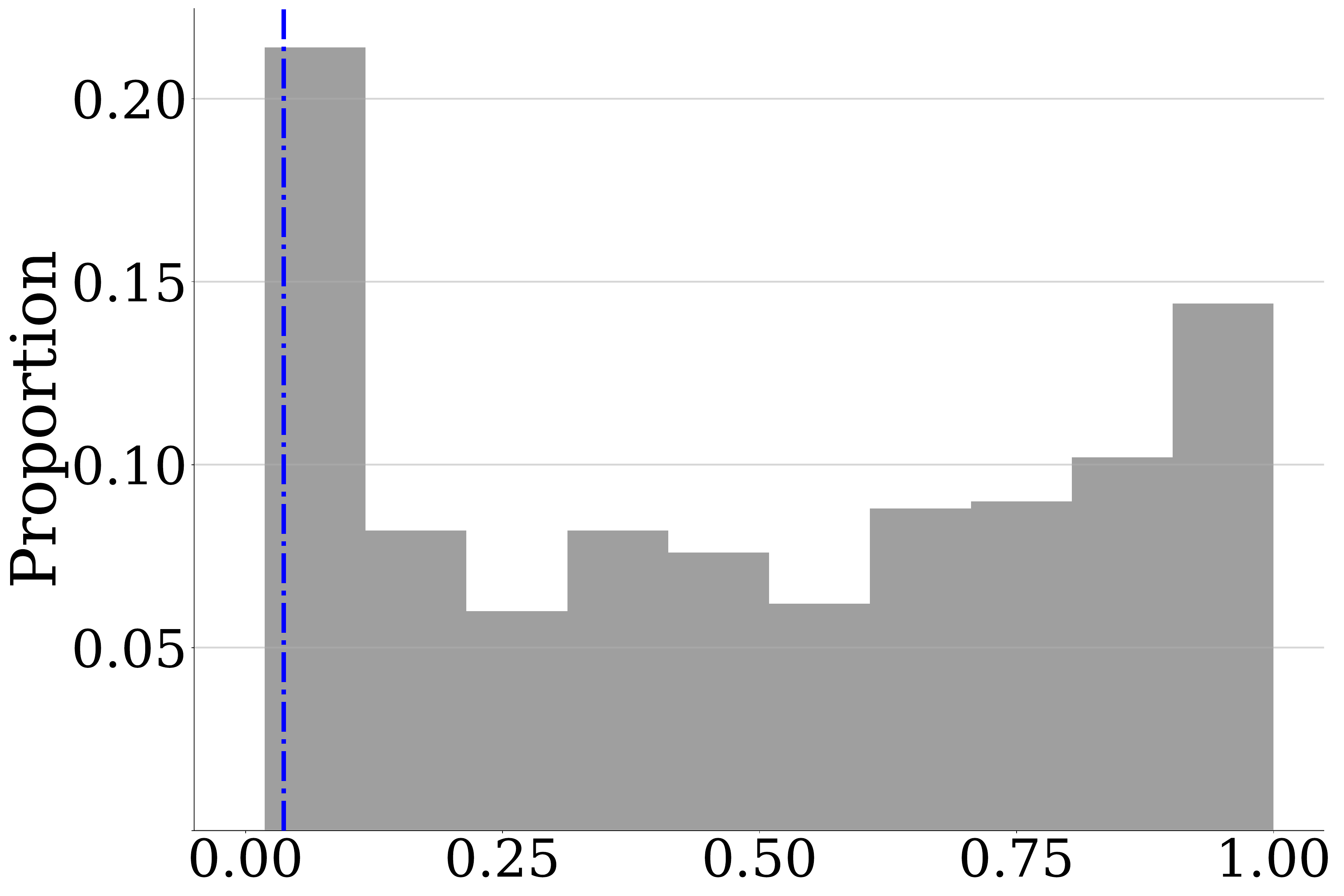}  \\
     \qquad Resampled trajectory 1 &
     \qquad Resampled trajectory 2 &
     \quad \qquad $\intscore[2,\engagement]$ \\
     (a)  & (b) & (c)
\end{tabular}
\caption{\tbf{Resampling results for user 4 considered in \cref{fig:interestingUserEngagement}, whose original trajectory exhibits interestingness of type $2$ for $\engagement$.} Panels (a) and (b) plot two randomly chosen (out of 500) resampled trajectories generated with zero advantage; the two trajectories, respectively, have $\intscore[2,\engagement]=$ 0.94, and 0.41. In panel (c), the vertical axis represents the fraction of the 500 resampled trajectories for this user with the value of $\intscore[2,\engagement]$ on the horizontal axis; and the vertical blue dashed line marks the observed $\intscore[2,\engagement]$ (value 0.037) for this user. \label{fig:user_int2_engagement}}
\end{figure}

\end{document}